\documentclass[twoside]{article} 

\usepackage{jmlr2e} 
\usepackage{epsfig} 
\usepackage{subfigure} 

\usepackage{natbib}

\usepackage{algorithm}
\usepackage{algorithmic}

\makeatletter
\newcommand{\ALOOP}[1]{\ALC@it\algorithmicloop\ #1%
  \begin{ALC@loop}}
\newcommand{\ENDALOOP}{\end{ALC@loop}\ALC@it\algorithmicendloop}
\newcommand{\INDSTATE}[1][1]{\STATE\hspace{3mm}}

\usepackage{hyperref}

\usepackage{amsmath}
\usepackage{pgfplots}
\usepackage{color}
\usepackage{flushend}
\usepackage{multirow}
\usepackage{rotating}
\usepackage{appendix}
\usepackage{tikz,pgfplots}

\usepackage{standalone} 

\newif\ifcompiletikz
\compiletikztrue

\input{mysymbol.sty}

\newcommand{\QED}{\hfill\ensuremath{\blacksquare}}

\newtheorem{assumption}{\hspace{0pt}\bf Assumption}

\begin{document}


\jmlrheading{1}{2000}{1-48}{4/00}{10/00}{Aryan Mokhtari, Alec Koppel, and Alejandro Ribeiro}


\ShortHeadings{A Class of Parallel Doubly Stochastic Algorithms for Large-Scale Learning}{Mokhtari, Koppel, and Ribeiro}
\firstpageno{1}

\title{A Class of Parallel Doubly Stochastic Algorithms\\ for Large-Scale Learning}

\author{\name Aryan Mokhtari  \email aryanm@seas.upenn.edu 
 \AND
       \name Alec Koppel \email akoppel@seas.upenn.edu  
       \AND
       \name Alejandro Ribeiro \email aribeiro@seas.upenn.edu  \\
       \addr Department of Electrical and Systems Engineering\\
       University of Pennsylvania\\
       Philadelphia, PA 19104, USA}

\editor{}

\maketitle


\thispagestyle{empty}
\pagestyle{empty}

\begin{abstract}
We consider learning problems over training sets in which both, the number of training examples and the dimension of the feature vectors, are large. To solve these problems we propose the random parallel stochastic algorithm (RAPSA). We call the algorithm random parallel because it utilizes multiple parallel processors to operate on a randomly chosen subset of blocks of the feature vector. We call the algorithm stochastic because processors choose training subsets uniformly at random. Algorithms that are parallel in either of these dimensions exist, but RAPSA is the first attempt at a methodology that is parallel in both the selection of blocks and the selection of elements of the training set. In RAPSA, processors utilize the randomly chosen functions to compute the stochastic gradient component associated with a randomly chosen block. The technical contribution of this paper is to show that this minimally coordinated algorithm converges to the optimal classifier when the training objective is convex. Moreover, we present an accelerated version of RAPSA (ARAPSA) that incorporates the objective function curvature information by premultiplying the descent direction by a Hessian approximation matrix. We further extend the results for asynchronous settings and show that if the processors perform their updates without any coordination the algorithms are still convergent to the optimal argument. RAPSA and its extensions are then numerically evaluated on a linear estimation problem and a binary image classification task using the MNIST handwritten digit dataset.
\end{abstract}


%
\section{Introduction}\label{sec_Introduction}

Learning is often formulated as an optimization problem that finds a vector of parameters $\bbx^*\in\reals^p$ that minimizes the average of a loss function across the elements of a training set. For a precise definition consider a training set with $N$ elements and let $f_{n}:\reals^p\to\reals$ be a convex loss function associated with the $n$-th element of the training set. The optimal parameter vector $\bbx^*\in\reals^p$ is defined as the minimizer of the average cost $F(\bbx) := (1/N)\sum_{n=1}^N f_{n}(\bbx)$,
\begin{equation}\label{eq:empirical_min}
   \bbx^* := \argmin_{\bbx} F(\bbx) 
          := \argmin_{\bbx}\frac{1}{N}\sum_{n=1}^N f_{n}(\bbx).
\end{equation}
Problems such as support vector machine classification, logistic and linear regression, and matrix completion can be put in the form of problem \eqref{eq:empirical_min}. In this paper, we are interested in large scale problems where both the number of features $p$ and the number of elements $N$ in the training set are very large -- which arise, e.g., in text \citep{Sampson:1990:NLA:104905.104911}, image \citep{mairal2010online}, and genomic \citep{tacsan2014selecting} processing.

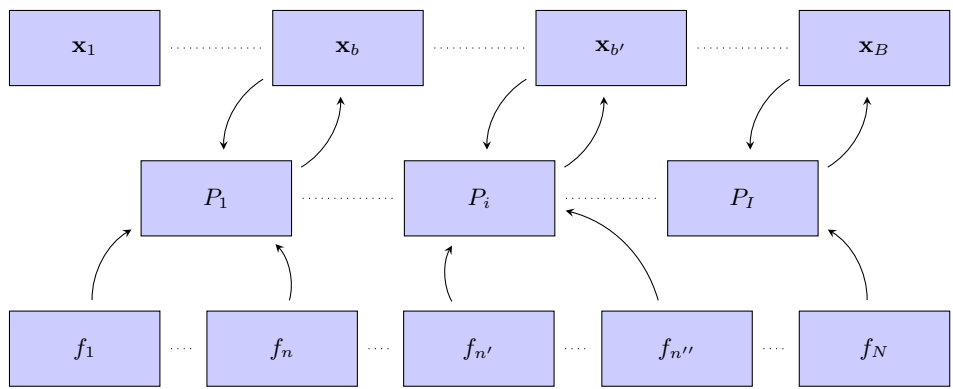
\begin{figure*}
\centering\def \thisplotscale {1}
\def \unit {\thisplotscale cm}

\tikzstyle{block} = [draw,
                     fill=blue!20,
                     minimum width  = 2*\unit,
                     minimum height = 1*\unit,]

{\small \begin{tikzpicture}[scale=\thisplotscale, shorten >=4,  shorten <=4]

    \node[block] at ( 1.75, 0) (processor 1) {$P_1$};
    \node[block] at ( 5.25, 0) (processor i) {$P_i$};
    \node[block] at ( 8.75, 0) (processor I) {$P_I$};            

    \node[block] at ( 0.0, 2) (variable 1)   {$\bbx_1$};
    \node[block] at ( 3.5, 2) (variable b)   {$\bbx_b$};
    \node[block] at ( 7.0, 2) (variable bp)  {$\bbx_{b'}$};
    \node[block] at (10.5, 2) (variable B)   {$\bbx_B$};            

    \node[block] at ( 0.000, -2) (function 1)   {$f_1$};
    \node[block] at ( 2.625, -2) (function n)   {$f_n$};
    \node[block] at ( 5.240, -2) (function np)   {$f_{n'}$};
    \node[block] at ( 7.875, -2) (function npp)   {$f_{n''}$};
    \node[block] at (10.500, -2) (function N)   {$f_N$};            

    \path (processor 1)   edge [dotted] (processor i);
    \path (processor i)   edge [dotted] (processor I);

    \path (variable 1)   edge [dotted] (variable b);
    \path (variable b)   edge [dotted] (variable bp);
    \path (variable bp)  edge [dotted] (variable B);
    
    \path (function 1)   edge [dotted] (function n);
    \path (function n)   edge [dotted] (function np);
    \path (function np)  edge [dotted] (function npp);
    \path (function npp) edge [dotted] (function N);

    \path[-stealth] (variable b)   edge [bend right] (processor 1);
    \path[-stealth] (processor 1)  edge [bend right] (variable b);
    \path[-stealth] (function 1)   edge [bend left]  (processor 1);
    \path[-stealth] (function n)   edge [bend right] (processor 1);    

    \path[-stealth] (variable bp)  edge [bend right] (processor i);
    \path[-stealth] (processor i)  edge [bend right] (variable bp);
    \path[-stealth] (function np)  edge [bend left]  (processor i);
    \path[-stealth] (function npp) edge [bend right] (processor i);    

    \path[-stealth] (variable B)   edge [bend right] (processor I);
    \path[-stealth] (processor I)  edge [bend right] (variable B);
    \path[-stealth] (function N)   edge [bend right]  (processor I);

\end{tikzpicture}} 
 \caption{Random parallel stochastic algorithm (RAPSA). At each iteration, processor $P_i$ picks a random block from the set $\{\bbx_1,\dots,\bbx_B\}$ and a random set of functions from the training set $\{f_1,\dots,f_N\}$. The functions drawn are used to evaluate a stochastic gradient component associated with the chosen block. RAPSA is shown here to converge to the optimal argument $\bbx^*$ of \eqref{eq:empirical_min}.}
\label{fig_diagram}
\end{figure*}

When $N$ and $p$ are large, the parallel processing architecture in Figure \ref{fig_diagram} becomes of interest. In this architecture, the parameter vector $\bbx$ is divided into $B$ blocks each of which contains $p_b\ll p$ features and a set of $I\ll B$ processors work in parallel on randomly chosen parameter blocks while using a stochastic subset of elements of the training set. In the schematic shown, Processor 1 fetches functions $f_1$ and $f_n$ to operate on block $\bbx_b$ and Processor $i$ fetches functions $f_{n'}$ and $f_{n''}$ to operate on block $\bbx_{b'}$. Other processors select other elements of the training set and other blocks with the majority of blocks remaining unchanged and the majority of functions remaining unused. The blocks chosen for update and the functions fetched for determination of block updates are selected independently at random in subsequent slots.

%

%
Problems that operate on blocks of the parameter vectors {\it or} subsets of the training set, but not on both, blocks {\it and} subsets, exist. Block coordinate descent (BCD) is the generic name for methods in which the variable space is divided in blocks that are processed separately. Early versions operate by cyclically updating all coordinates at each step \citep{Luo1992,Tseng01convergenceof,xu2014globally}, while more recent parallelized versions of coordinate descent have been developed to accelerate convergence of BCD \citep{richtarik2012parallel,lu2013complexity,nesterov2012efficiency,beck2013convergence}. Closer to the architecture in Figure \ref{fig_diagram}, methods in which subsets of blocks are selected at random have also been proposed \citep{liu2013asynchronous,6612036,nesterov2012efficiency,lu2015complexity}. BCD, serial, parallel, or random, can handle cases where the parameter dimension $p$ is large but requires access to all $N$ training samples at each iteration. 

Parallel implementations of block coordinate methods have been developed initially in this setting for composite optimization problems \citep{richtarik2012parallel}. A collection of parallel processors update randomly selected blocks concurrently at each step. Several variants that select blocks in order to maximize the descent at each step are proposed in \citep{scherrer2012feature,facchinei2015parallel,shalev2013accelerated}. The aforementioned works require that parallel processors operate on a common time index. In contrast, asynchronous parallel methods, originally proposed in \cite{bertsekas1989parallel}, have been developed to solve optimization problems where processors are \emph{not} required to operate with a common global clock. This work focused on solving a fixed point problem over a separable convex set, but the analysis is more restrictive than standard convexity assumptions. For a standard strongly convex optimization problem, in contrast, \cite{liu2013asynchronous} establish linear convergence to the optimum. All of these works are developed for optimization problems with deterministic objectives.

To handle the case where the number of training examples $N$ is very large, methods have been developed to only process a subset of sample points at a time. These methods are known by the generic name of stochastic approximation and rely on the use of stochastic gradients. In plain stochastic gradient descent (SGD), the gradient of the aggregate function is estimated by the gradient of a randomly chosen function $f_n$ \citep{robbins1951}. Since convergence of SGD is slow more often that not, various recent developments have been aimed at accelerating its convergence. These attempts include methodologies to reduce the variance of stochastic gradients \citep{schmidt2013minimizing,johnson2013accelerating,defazio2014saga} and the use of ideas from quasi-Newton optimization to handle difficult curvature profiles \citep{schraudolph2007stochastic,bordes2009sgd,mokhtari2014res,mokhtari2014global}. 
More pertinent to the work considered here are the use of cyclic block SGD updates \citep{xuyin2014} and the exploitation of sparsity properties of feature vectors to allow for parallel updates \citep{Niu11hogwild:a}. These methods are suitable when the number of elements in the training set $N$ is large but don't allow for parallel feature processing unless parallelism is inherent to the problem's structure.

The random parallel stochastic algorithm (RAPSA) proposed in this paper represents the first effort at implementing the architecture in Fig. \ref{fig_diagram} that randomizes over both parameters and sample functions, and may be implemented in parallel. In RAPSA, the functions fetched by a processor are used to compute the stochastic gradient component associated with a randomly chosen block (Section \ref{sec:rapsa}). The processors do not coordinate in either choice except to avoid selection of the same block. Our main technical contribution is to show that RAPSA iterates converge to the optimal classifier $\bbx^*$ when using a sequence of decreasing stepsizes and to a neighborhood of the optimal classifier when using constant stepsizes (Section \ref{sec:convergence_analysis}). In the latter case, we further show that the rate of convergence to this optimality neighborhood is linear in expectation. These results are interesting because only a subset of features are updated per iteration and the functions used to update different blocks are, in general, different. We propose two extensions of RAPSA. Firstly, motivated by the improved performance results of quasi-Newton methods relative to gradient methods in online optimization, we propose an extension of RAPSA which incorporates approximate second-order information of the objective, called Accelerated RAPSA. We also consider an extension of RAPSA in which parallel processors are not required to operate on a common time index, which we call Asynchronous RAPSA. We further show how these extensions yield an accelerated doubly stochastic algorithm for an asynchronous system. We establish that the performance guarantees of RAPSA carry through to asynchronous computing architectures. We then numerically evaluate the proposed methods on a large-scale linear regression problem as well as the MNIST digit recognition problem (Section \ref{sec:simulations}).

%

\section{Random Parallel Stochastic Algorithm (RAPSA)}\label{sec:rapsa}

%

We consider a more general formulation of \eqref{eq:empirical_min} in which the number $N$ of functions $f_n$ is not necessarily finite. Introduce then a random variable $\bbtheta \in \bbTheta \subset \reals^q$ that determines the choice of the random smooth convex function $f(\cdot,\bbtheta) : \reals^p \to \reals$. We consider the problem of minimizing the expectation of the random functions $F(\bbx):=\mbE_{\bbtheta}[f(\bbx,\bbtheta)]$, 
\begin{equation} \label{eq:block_stoch_opt}
   \bbx^* := \argmin_{\bbx\in \reals^p} F(\bbx)
          := \argmin_{\bbx\in \reals^p} \mbE_{\bbtheta}\left[{f(\bbx , \bbtheta)}\right].
\end{equation}
Problem \eqref{eq:empirical_min} is a particular case of \eqref{eq:block_stoch_opt} in which each of the functions $f_n$ is drawn with probability $1/N$. Observe that when $\bbtheta=(\bbz, \bby)$ with feature vector $\bbz\in\reals^p$ and target variable $\bby\in\reals^q$ or $y\in\{0,1\}$, the formulation in \eqref{eq:block_stoch_opt} encapsulates generic supervised learning problems such as regression or classification, respectively. We refer to $f(\cdot,\bbtheta)$ as instantaneous functions and to $F(\bbx)$ as the average function. 

RAPSA utilizes $I$ processors to update a random subset of blocks of the variable $\bbx$, with each of the blocks relying on a subset of randomly and independently chosen elements of the training set; see Figure \ref{fig_diagram}. Formally, decompose the variable $\bbx$ into $B$ blocks to write $\bbx=[\bbx_1;\dots;\bbx_B]$, where block $b$ has length $p_b$ so that we have $\bbx_b \in \reals^{p_b}$. At iteration $t$, processor $i$ selects a random index $b_i^t$ for updating and a random subset $\bbTheta_i^t$ of $L$ instantaneous functions. It then uses these instantaneous functions to determine stochastic gradient components for the subset of variables $\bbx_b=\bbx_{b_i^t}$ as an average of the components of the gradients of the functions $f( \bbx^{t}, \bbtheta)$ for $\bbtheta\in\bbTheta_i^t$,
\begin{equation}\label{block_sto_grad}
    \nabla_{\bbx_b} f( \bbx^{t}, \bbTheta_i^t)
      = \frac{1}{L}\sum_{\bbtheta\in \bbTheta_i^t} \nabla_{\bbx_b} f( \bbx^{t}, \bbtheta), 
    \qquad   b = b_i^t.
\end{equation}
Note that $L$ can be interpreted as the size of mini-batch for gradient approximation. 
The stochastic gradient block in \eqref{block_sto_grad} is then modulated by a possibly time varying stepsize $\gamma^t$ and used by processor $i$ to update the block $\bbx_b=\bbx_{b_i^t}$ 
\begin{align} \label{eq:block_stochastic_gradient_1}
   \bbx^{t+1}_{b}  =  \bbx^{t}_b  - \gamma^t  \nabla_{\bbx_b} f( \bbx^{t}, \bbTheta_i^t) 
   \qquad   \qquad   b = b_i^t .
\end{align}
RAPSA is defined by the joint implementation of \eqref{block_sto_grad} and \eqref{eq:block_stochastic_gradient_1} across all $I$ processors, and is summarized in Algorithm \ref{algo_RAPSA}. We would like to emphasize that the number of updated blocks which is equivalent to the number of processors $I$ is not necessary equal to the total number of blocks $B$. In other words, we may update only a subset of coordinates $I/B<1$ at each iteration. We define $r:=I/B$ as the ratio of the updated blocks to the total number of blocks which is smaller than $1$. 

The selection of blocks is coordinated so that no processors operate in the same block. The selection of elements of the training set is uncoordinated across processors. The fact that at any point in time a random subset of blocks is being updated utilizing a random subset of elements of the training set means that RAPSA requires almost no coordination between processors. The contribution of this paper is to show that this very lean algorithm converges to the optimal argument $\bbx^*$ as we show in Section \ref{sec:convergence_analysis}.

%
\begin{algorithm}[t]
\caption{Random Parallel Stochastic  Algorithm (RAPSA)}\label{algo_RAPSA} 
\begin{algorithmic}[1] 
{\FOR {$t=0,1,2,\dots$}
{ \ALOOP {{\bf{in parallel}}, processors $i=1,\dots,I$ execute:}
\STATE Select block $b_i^t\in\{1,\dots,B\}$ uniformly at random from set of blocks 
\STATE Choose training subset $\bbTheta_i^t$ for block $\bbx_b$, 
\STATE Compute stochastic gradient :
$\displaystyle{
\nabla_{\bbx_b} f( \bbx^{t}, \bbTheta_i^t)=\frac{1}{L}\sum_{\bbtheta\in \bbTheta_i^t}\nabla_{\bbx_b} f( \bbx^{t}, \bbtheta), \quad b=b_i^t
}$ [cf. \eqref{block_sto_grad}]
\STATE Update the coordinates $b_i^t$ of the decision variable 
$\displaystyle{
\bbx^{t+1}_{b}  \ = \ \bbx^{t}_{b}  - \gamma^t\ \nabla_{\bbx_b} f( \bbx^{t}, \bbTheta_i^t)
}$
 \ENDALOOP {; Transmit updated blocks $i\in\ccalI^t\subset \{1,\dots,B\}$ to shared memory}}
\ENDFOR}
\end{algorithmic}
\end{algorithm}

%
\section{Accelerated Random Parallel Stochastic Algorithm (ARAPSA)}\label{sec:arapsa}

As we mentioned in Section \ref{sec:rapsa}, RAPSA operates on first-order information which may lead to slow convergence in ill-conditioned problems. We introduce Accelerated RAPSA (ARAPSA) as a parallel doubly stochastic algorithm that incorporates second-order information of the objective by separately approximating the function curvature for each block. We do this by implementing the oLBFGS algorithm for different blocks of the variable $\bbx$. For related approaches, see, for instance, \cite{Broyden,Byrd,Dennis,Li}. Define $\hbB_b^t$ as an approximation for the Hessian inverse of the objective function that corresponds to the block $b$ with the corresponding variable $\bbx_b$. If we consider $b_i^t$ as the block that processor $i$ chooses at step $t$, then the update of ARAPSA is defined as multiplication of the descent direction of RAPSA by $\hbB_b^t$, i.e., 
\begin{equation}\label{ARAPS_update}
\bbx^{t+1}_{b}  \ = \ \bbx^{t}_{b}  - \gamma^t\  \hbB_b^t\ \nabla_{\bbx_b} f( \bbx^{t}, \bbTheta_i^t)  \qquad   b = b_i^t.
\end{equation}
Subsequently, we define the $\hbd_{b}^t:= \hbB_b^t\ \nabla_{\bbx_b} f( \bbx^{t}, \bbTheta_i^t)$. We next detail how to properly specify the block approximate Hessian $\hbB_b^t$  so that it behaves in a manner comparable to the true Hessian. To do so, define for each block coordinate $\bbx_b$ at step $t$ the variable variation $\bbv_b^t$ and the stochastic gradient variation $\hbr_b^t$ as
\begin{equation}\label{var_grad_var}
\bbv_b^t \:=\ \bbx^{t+1}_{b}  - \bbx^{t}_{b} , \qquad 
\hbr_b^t \:=\ \nabla_{\bbx_b} f( \bbx^{t+1}, \bbTheta_i^t)  - \nabla_{\bbx_b} f( \bbx^{t}, \bbTheta_i^t).
\end{equation}
Observe that the stochastic gradient variation $\hbr_b^t$ is defined as the difference of stochastic gradients at times $t+1$ and $t$ corresponding to the block $\bbx_b$ for a common set of realizations $\bbTheta_i^t$. The term $ \nabla_{\bbx_b} f( \bbx^{t}, \bbTheta_i^t)$ is the same as the stochastic gradient used at time $t$ in \eqref{ARAPS_update}, while $\nabla_{\bbx_b} f( \bbx^{t+1}, \bbTheta_i^t)$ is computed only to determine the stochastic gradient variation $\hbr_b^t$. An alternative and perhaps more natural definition for the stochastic gradient variation is $\nabla_{\bbx_b} f( \bbx^{t+1}, \bbTheta_i^{t+1})  - \nabla_{\bbx_b} f( \bbx^{t}, \bbTheta_i^t)$. However, as pointed out in \cite{schraudolph2007stochastic}, this formulation is insufficient for establishing the convergence of stochastic quasi-Newton methods. We proceed to developing a block-coordinate quasi-Newton method by first noting an important property of the true Hessian, and design our approximate scheme to satisfy this property. 
 The secant condition may be interpreted as stating that the stochastic gradient of a quadratic approximation of the objective function evaluated at the next iteration agrees with the stochastic gradient at the current iteration. We select a Hessian inverse approximation matrix associated with block $\bbx_b$ such that it satisfies the secant condition $\hbB_b^{t+1}\hbr_b^{t}=\bbv_b^{t}$, and thus behaves in a comparable manner to the true block Hessian.

%
\begin{algorithm}[t]
\caption{Computation of the ARAPSA step $\hbd_b^t =\hbB_b^{t} \nabla_{\bbx_b} f( \bbx^{t}, \bbTheta_i^t) $ for block $\bbx_b$.}
\label{algo_lbfgs} 
\begin{algorithmic}[1]
   \STATE \textbf{function}  
          $\hbd_b^t=\bbq^\tau$ 
          = ARAPSA Step$\left(\hbB_b^{t,0}, \
                              \bbp^0= \nabla_{\bbx_b} f( \bbx^{t}, \bbTheta_i^t), \
                                \{\bbv_b^{u},\hbr_b^{u}\}_{u=t-\tau}^{t-1}\right)$ 
   \FOR [Loop to compute constants $\alpha^u$ and sequence $\bbp^u$] 
        {$u= 0, 1, \ldots, \tau-1$ } 
      \STATE Compute and store scalar 
             $\alpha^{u}= \hat{\rho}_b^{t-u-1}(\bbv_b^{t-u-1})^{T}\bbp^{u}$
      \STATE Update sequence vector 
             $\bbp^{u+1}=\bbp^u-\alpha^{u}\hbr_b^{t-u-1}$.
   \ENDFOR
   \STATE Multiply $\bbp^\tau$ by initial matrix:  
          $\bbq^{0}= \hbB_b^{t,0} \bbp^{\tau}  $
   \FOR [Loop to compute constants $\beta_u$ and sequence $\bbq_u$] 
        {$u= 0, 1, \ldots, \tau-1$}
      \STATE Compute scalar 
             $\beta^u= \hat{\rho}_b^{t-\tau+u}(\hbr_b^{t-\tau+u})^{T}\bbq^{u}$ 
      \STATE Update sequence vector
             $\bbq^{u+1}=\bbq^{u}+(\alpha^{\tau-u-1}-\beta^{u})\bbv_b^{t-\tau+u}$        
   \ENDFOR \ \{return $\hbd_b^t = \bbq^\tau$\} 
\end{algorithmic}\end{algorithm}
 
 The oLBFGS Hessian inverse update rule maintains the secant condition at each iteration by using information of the last $\tau\geq1$ pairs of variable and stochastic gradient variations $\{\bbv_b^u,\hbr_b^{u}\}_{u=t-\tau}^{t-1}$. To state the update rule of oLBFGS for revising the Hessian inverse approximation matrices of the blocks, define a matrix as $\hbB_b^{t,0}:= \eta_b^t\bbI$ for each block $b$ and $t$, where the constant $\eta_b^t$ for $t>0$ is given by
\begin{equation}\label{initial_matrix_update}
\eta_b^t:= \frac{(\bbv_b^{t-1})^{T}\hbr_b^{t-1}}{\|\hbr_b^{t-1}\|^2},
\end{equation}
while the initial value is $\eta_b^t=1$.
The matrix $\hbB_b^{t,0}$ is the initial approximate for the Hessian inverse associated with block $\bbx_b$. The approximate matrix $\hbB_b^{t}$ is computed by updating the initial matrix $\hbB_b^{t,0}$ using the last $\tau $ pairs of curvature information $\{\bbv_b^u,\hbr_b^{u}\}_{u=t-\tau}^{t-1}$. We define the approximate Hessian inverse $\hbB_b^{t}=\hbB_b^{t,\tau}$ corresponding to block $\bbx_b$ at step $t$ as the outcome of $\tau$ recursive applications of the update
\begin{equation}\label{SLBFGS_update}
   \hbB_b^{t,u+1}
      =   (\hbZ_b^{t-\tau+u})^T\ \!
                \hbB_b^{t,u}\ \! (\hbZ_b^{t-\tau+u})   
            + \hat{ \rho}_b^{t-\tau+u} \ \!(\bbv_b^{t-\tau+u})\ \!(\bbv_b^{t-\tau+u})^{T},
\end{equation}
where the matrices $\hbZ_b^{t-\tau+u}$ and the constants $\hat{\rho}_b^{t-\tau+u} $ in \eqref{SLBFGS_update} for $u=0,\dots, \tau-1$ are defined as 
\begin{equation}\label{Z_rho_definitions}
   \hat{\rho}_b^{t-\tau+u}\ =\ \frac{1}{(\bbv_b^{t-\tau+u})^{T}\hbr_b^{t-\tau+u}}
   \quad \text{and}\quad
   \hbZ_b^{t-\tau+u}\ =\ \bbI - \hat{\rho}_b^{t-\tau+u} \hbr_b^{t-\tau+u} (\bbv_b^{t-\tau+u})^{T}.
\end{equation}

The block-wise oLBFGS update defined by \eqref{var_grad_var} - \eqref{Z_rho_definitions} is summarized in Algorithm \ref{algo_lbfgs}. The computation cost of $\hbB_b^{t}$ in \eqref{SLBFGS_update} is in the order of $O(p_b^2)$, however, for the update in \eqref{ARAPS_update} the descent direction $\hbd_b^t := \hbB_b^{t} \nabla_{\bbx_b} f( \bbx^{t}, \bbTheta_i^t) $ is required. \cite{DingNocedal} introduce an efficient implementation of product $\hbB_b^{t} \nabla_{\bbx_b} f( \bbx^{t}, \bbTheta_i^t)$ that requires computation complexity of order $O(\tau p_b)$. 
We use the same idea for computing the descent direction of ARAPSA for each block -- more details are provided below. Therefore, the computation complexity of updating each block for ARAPSA is in the order of  $O(\tau p_b)$, while RAPSA requires $O( p_b)$ operations. 
On the other hand, ARAPSA accelerates the convergence of RAPSA by incorporating the second order information of the objective function for the block updates, as may be observed in the numerical analyses provided in Section \ref{sec:simulations}.

For reference, ARAPSA is also summarized in algorithmic form in Algorithm \ref{algo_ARAPS}. Steps 2 and 3 are devoted to assigning random blocks to the processors. In Step 2 a subset of available blocks $\ccalI^t$ is chosen. These blocks are assigned to different processors in Step 3. In Step 5 processors compute the partial stochastic gradient corresponding to their assigned blocks $\nabla_{\bbx_b}f(\bbx^t,\bbTheta_i^t)$ using the acquired samples in Step 4. Steps 6 and 7 are devoted to the computation of the ARAPSA descent direction $\hbd_{i}^t$. In Step 6 the approximate Hessian inverse $\hbB_{b}^{t,0}$ for block $\bbx_b$ is initialized as $\hbB_b^{t,0}= {\eta_b^{t}}\bbI$ which is a scaled identity matrix using the expression for ${\eta_b^{t}}$ in \eqref{initial_matrix_update} for $t>0$. The initial value of $\eta_b^t$ is $\eta_b^0=1$. In Step 7 we use Algorithm \ref{algo_lbfgs} for efficient computation of the descent direction $\hbd_b^{t}=\hbB_b^t\ \nabla_{\bbx_b} f( \bbx^{t}, \bbTheta_i^t)$. The descent direction $\hbd_b^{t}$ is used to update the block $\bbx_b^t$ with stepsize $\gamma^t$ in Step 8. Step 9 determines the value of the partial stochastic gradient $ \nabla_{\bbx_b} f( \bbx^{t+1}, \bbTheta_i^t)$ which is required for the computation of stochastic gradient variation $\hbr_b^{t}$. In Step 10 the variable variation $\bbv_b^{t}$ and stochastic gradient variation $\hbr_b^{t}$ associated with block $\bbx_b$ are computed to be used in the next iteration.

%
\begin{algorithm}[t]
\caption{Accelerated Random Parallel Stochastic  Algorithm (ARAPSA)}\label{algo_ARAPS} 
\begin{algorithmic}[1] 
{\FOR {$t=0,1,2,\dots$}
{ \ALOOP {{\bf{in parallel}}, processors $i=1,\dots,I$ execute:}
\STATE Select block $b_i^t$ uniformly at random from set of blocks $\{1,\dots,B\}$
\STATE Choose a set of realizations $\bbTheta_i^t$ for the block $\bbx_b$
\STATE Compute stochastic gradient :
$\displaystyle{
\nabla_{\bbx_b} f( \bbx^{t}, \bbTheta_i^t)=\frac{1}{L}\sum_{\bbtheta\in \bbTheta_i^t}\nabla_{\bbx_b} f( \bbx^{t}, \bbtheta)
}$ [cf. \eqref{block_sto_grad}]
\STATE Compute the initial Hessian inverse approximation: $\displaystyle{\hbB_b^{t,0}=\eta_b^t \bbI}$
\STATE Compute descent direction:  
           $\displaystyle{\hbd_{b}^t = 
            \text{ARAPSA Step}\left(\hbB_b^{t,0}, \   
                              \nabla_{\bbx_b} f( \bbx^{t}, \bbTheta_i^t) , \
                               \{\bbv_b^{u},\hbr_b^{u}\}_{u=t-\tau}^{t-1}\right)}$  
\STATE Update the coordinates of the decision variable 
$\displaystyle{
\bbx^{t+1}_{b}  \ = \ \bbx^{t}_{b}  - \gamma^t\ \hbd_{b}^t
}$
   \STATE Compute {\it updated} stochastic gradient:       
          $\displaystyle{
             \nabla_{\bbx_b} f( \bbx^{t+1}, \bbTheta_i^t)=\frac{1}{L}\sum_{\bbtheta\in \bbTheta_i^t}\nabla_{\bbx_b} f( \bbx^{t+1}, \bbtheta)
          }$ [cf. \eqref{block_sto_grad}]
   \STATE Update variations    
          $\displaystyle{\bbv_b^t= \bbx_b^{t+1}-\bbx_b^t}$  and    
          $\displaystyle{\hbr_{i}^t=\nabla_{\bbx_b} f( \bbx^{t+1}, \bbTheta_i^t)-\nabla_{\bbx_b} f( \bbx^{t}, \bbTheta_i^t)}$ 
          [ cf.\eqref{var_grad_var}]
           \ENDALOOP {; Transmit updated blocks $i\in\ccalI^t\subset \{1,\dots,B\}$ to shared memory}}
\ENDFOR}
\end{algorithmic}
\end{algorithm}

%

\section{Asynchronous Architectures}\label{sec:asyn}

%
\begin{algorithm}[t]
\caption{Asynchronous RAPSA at processor $i$}\label{algo_asyn} 
\begin{algorithmic}[1] 
{\WHILE {$t< T$}
\STATE {\bf{Processor $i\in \{1,\dots,I\}$ at time index $t$ executes the following steps}}:
\INDSTATE Select block $b_i^t$ uniformly at random from set of blocks $\{1,\dots,B\}$
\INDSTATE Choose a set of realizations $\bbTheta_i^t$ for the block $\bbx_b$, $b=b_i^t$
\INDSTATE Compute stochastic gradient :
$\displaystyle{
\nabla_{\bbx_b} f( \bbx^{t}, \bbTheta_i^t)=\frac{1}{L}\sum_{\bbtheta\in \bbTheta_i^t}\nabla_{\bbx_b} f( \bbx^{t}, \bbtheta)
}$ [cf. \eqref{block_sto_grad}]
\INDSTATE Update the coordinates of the decision variable 
$\displaystyle{
\bbx^{t+\tau+1}_{b}  \ = \ \bbx^{t+\tau}_{b}  - \gamma^{t+\tau}\nabla_{\bbx_b} f( \bbx^{t}, \bbTheta_i^t)
}$        
          \STATE {\bf Send updated parameters $\bbx_b^{t+1}$ associated with block $b=b_i^t$ to shared memory}
          \STATE If another processor is also operating on block $b_i^t$ at time $t$, randomly overwrite
\ENDWHILE}
\end{algorithmic}
\end{algorithm}

Up to this point, the RAPSA method dictates that distinct parallel processors select blocks $b_i^t\in \{1,\dots, B\}$ uniformly at random at each time step $t$ as in Figure \ref{fig_diagram}. However, the requirement that each processor operates on a common time index is burdensome for parallel operations on large computing clusters, as it means that nodes must wait for the processor which has the longest computation time at each step before proceeding. Remarkably, we are able to extend the methods developed in Sections \ref{sec:rapsa} and \ref{sec:arapsa} to the case where the parallel processors need not to operate on a common time index (lock-free) and establish that their performance guarantees carry through, so long as the degree of their asynchronicity is bounded in a certain sense. In doing so, we alleviate the computational bottleneck in the parallel architecture, allowing processors to continue processing data as soon as their local task is complete.

\subsection{Asynchronous RAPSA}\label{subsec:asynchronous}
Consider the case where each node operates asynchronously. In this case, at an instantaneous time index $t$, only one processor executes an update, as all others are assumed to be busy. If two processors complete their prior task concurrently, then they draw the same time index at the next available slot, in which case the tie is broken at random. Suppose processor $i$ selects block $b_i^t\in  \{1,\dots, B\}$ at time $t$. Then it grabs the associated component of the decision variable $\bbx_b^t$ and computes the stochastic gradient $ \nabla_{\bbx_b} f( \bbx^{t},\bbTheta_i^{t}) $ associated with the samples $\bbTheta_i^{t}$. This process may take time and during this process other processors may overwrite the variable $\bbx_b$. Consider the case that the process time of computing stochastic gradient or equivalently the descent direction is $\tau$. Thus, when processor $i$ updates the block $b$ using the evaluated stochastic gradient $\nabla_{\bbx_b} f( \bbx^{t},\bbTheta_i^{t}) $, it performs the update
\begin{align} \label{eq:block_stochastic_gradient_asyn}
   \bbx^{t+\tau+1}_{b}  =  \bbx^{t+\tau}_b  - \gamma^{t+\tau} \ \! \nabla_{\bbx_b} f( \bbx^{t}, \bbTheta_i^{t}) 
   \quad   \qquad   b = b_i^{t} .
\end{align}
Thus, the descent direction evaluated based on the available information at step $t$ is used to update the variable at time $t+\tau$. Asynchronous RAPSA is summarized in Algorithm \ref{algo_asyn}.
Note that the delay comes from asynchronous implementation of the algorithm and the fact that other processors are able to modify the variable $\bbx_b$ during the time that processor $i$ computes its descent direction. We assume the the random time $\tau$ that each processor requires to compute its descent direction is bounded above by a constant $\Delta$, i.e., $\tau \leq \Delta$ -- see Assumption \ref{delay_assumption}.

 Despite the minimal coordination of the asynchronous random parallel stochastic algorithm in \eqref{eq:block_stochastic_gradient_asyn}, we may establish the same performance guarantees as that of RAPSA in Section \ref{sec:rapsa}. These analytical properties are investigated at length in Section \ref{sec:convergence_analysis}.

%
\begin{remark}
One may raise the concern that there could be instances that two processors or more work on a same block. Although, this event is not very likely since $I<<B$, there is a positive chance that it might happen. This is true since the available processor picks the block that it wants to operate on uniformly at random from the set $ \{1,\dots, B\}$. We show that this event does not cause any issues and the algorithm can eventually converge to the optimal argument even if more than one processor work on a specific block at the same time -- see Section \ref{sec:asyn_rapsa_convg}. Functionally, this means that if one block is worked on concurrently by two processors, the memory coordination requires that the result of one of the two processors is written to memory with probability $1/2$. This random overwrite rule applies to the case that three or more processors are operating on the same block as well. In this case, the result of one of the conflicting processors is written to memory with probability $1/C$ where $C$ is the number of conflicting processors.
\end{remark}

\subsection{Asynchronous ARAPSA}

%
\begin{algorithm}[t]
\caption{Asynchronous Accelerated RAPSA at processor $i$}\label{algo_asyn_ARAPS} 
\begin{algorithmic}[1] 
{\WHILE {$t< T$}
\STATE {\bf Processor $i\in \{1,\dots,I\}$ at time index $t$ executes the following steps}:
\INDSTATE Select block $b_i^t$ uniformly at random from set of blocks $\{1,\dots,B\}$
\INDSTATE Choose a set of realizations $\bbTheta_i^t$ for the block $\bbx_b$, $b=b_i^t$
\INDSTATE Compute stochastic gradient :
$\displaystyle{
\nabla_{\bbx_b} f( \bbx^{t}, \bbTheta_i^t)=\frac{1}{L}\sum_{\bbtheta\in \bbTheta_i^t}\nabla_{\bbx_b} f( \bbx^{t}, \bbtheta)
}$ [cf. \eqref{block_sto_grad}]
\INDSTATE Compute the initial Hessian inverse approximation: $\displaystyle{\hbB_b^{t,0}=\eta_b^t \bbI}$
\INDSTATE Compute descent direction:  
           $\displaystyle{\hbd_{b}^t = 
            \text{ARAPSA Step}\left(\hbB_b^{t,0}, \   
                              \nabla_{\bbx_b} f( \bbx^{t}, \bbTheta_i^t) , \
                               \{\bbv_b^{u},\hbr_b^{u}\}_{u=t-\tau}^{t-1}\right)}$  
\INDSTATE Update the coordinates of the decision variable 
$\displaystyle{
\bbx^{t+\tau+1}_{b}  \ = \ \bbx^{t+\tau}_{b}  - \gamma^{t+\tau}\ \hbd_{b}^t
}$
   \INDSTATE Compute {\it updated} stochastic gradient:       
          $\displaystyle{
             \nabla_{\bbx_b} f( \bbx^{t+\tau+1}, \bbTheta_i^t)=\frac{1}{L}\sum_{\bbtheta\in \bbTheta_i^t}\nabla_{\bbx_b} f( \bbx^{t+\tau+1}, \bbtheta)
          }$ [cf. \eqref{block_sto_grad}]
   \INDSTATE Update variations    
          $\displaystyle{\bbv_b^{t}= \bbx_b^{t+\tau+1}-\bbx_b^t}$  and    
          $\displaystyle{\hbr_{b}^{t}=\nabla_{\bbx_b} f( \bbx^{t+\tau+1}, \bbTheta_i^t)-\nabla_{\bbx_b} f( \bbx^{t}, \bbTheta_i^t)}$ 
          [ cf.\eqref{var_grad_var_asyn}]
          \INDSTATE Overwrite the oldest pairs of $\bbv_b$ and $\hbr_{b}$ in local memory by $\bbv_b^{t}$ and $\hbr_{b}^{t}$, respectively. 
          \STATE {\bf Send updated parameters $\bbx_b^{t+1}$, $\{\bbv_b^u,\hbr_b^u\}_{u=t-\tau}^{t-1}$ to shared memory.}
          \STATE If another processor is  operating on block $b_i^t$, choose to overwrite with probability $1/2$.
\ENDWHILE}
\end{algorithmic}
\end{algorithm}

In this section, we study the asynchronous implementation of accelerated RAPSA (ARAPSA). The main difference between the synchronous of implementation ARAPSA in Section \ref{sec:arapsa} and the asynchronous version is in the update of the variable $\bbx_b^t$ corresponding to the block $b$. Consider the case that processor $i$ finishes its previous task at time $t$, chooses the block $b=b_{i}^{t}$, and reads the variable $\bbx_{b}^{t}$. Then, it computes the stochastic gradient  $f( \bbx^{t}, \bbTheta_i^{t}) $ using the set of random variables $\bbTheta_{i}^{t}$. Further, processor $i$ computes the descent direction $ \hbB_b^{t}\ \nabla_{\bbx_b} f( \bbx^{t}, \bbTheta_i^{t}) $ using the last $\tau$ sets of curvature information {$\{\bbv_b^u,\hbr_b^{u}\}_{u=t-\tau}^{t-1}$} as shown in Algorithm 1. If we assume that the required time to compute the descent direction $ \hbB_b^{t}\ \nabla_{\bbx_b} f( \bbx^{t}, \bbTheta_i^{t}) $ is $\tau'$, processor $i$ updates the variable $\bbx_b^{t+\tau'}$ as 
\begin{align} \label{eq:block_stochastic_gradient_asyn_arapsa}
   \bbx^{t+\tau'+1}_{b}  =  \bbx^{t+\tau'}_b  - \gamma^{t+\tau'}\ \! \hbB_b^{t}\ \! \nabla_{\bbx_b} f( \bbx^{t}, \bbTheta_i^{t}) 
   \quad   \qquad   b = b_i^{t} .
\end{align}
Note that the update in \eqref{eq:block_stochastic_gradient_asyn_arapsa} is different from the synchronous version in \eqref{ARAPS_update} in the time index of the variable that is updated using the available information at time $t$. In other words, in the synchronous implementation the descent direction $ \hbB_b^{t}\ \nabla_{\bbx_b} f( \bbx^{t}, \bbTheta_i^{t}) $ is used to update the variable $\bbx_b^t$ with the same time index, while this descent direction is executed to update the variable $\bbx_b^{t+\tau'}$ in asynchronous ARAPSA.

Note that the definitions of the variable variation $\bbv_b^t$ and the stochastic gradient variation $\hbr_b^t$ are different in asynchronous setting and they are given by 
\begin{equation}\label{var_grad_var_asyn}
\bbv_b^t \:=\ \bbx^{t+\tau'+1}_{b}  - \bbx^{t}_{b} , \qquad 
\hbr_b^t \:=\ \nabla_{\bbx_b} f( \bbx^{t+\tau'+1}, \bbTheta_i^t)  - \nabla_{\bbx_b} f( \bbx^{t}, \bbTheta_i^t).
\end{equation}
This modification comes from the fact that the stochastic gradient $\nabla_{\bbx_b} f( \bbx^{t}, \bbTheta_i^t)$ is already evaluated for the descent direction in \eqref{eq:block_stochastic_gradient_asyn_arapsa}. Thus, we define the stochastic gradient variation by computing the difference of the stochastic gradient $\nabla_{\bbx_b} f( \bbx^{t}, \bbTheta_i^t)$ and the stochastic gradient associated with the same random set $\bbTheta_i^t$ evaluated at the most recent iterate which is $\bbx^{t+\tau'+1}_{b} $. Likewise, the variable variation is redefined as the difference $\bbx^{t+\tau'+1}_{b}  - \bbx^{t}_{b} $. The steps of asynchronous ARAPSA are summarized in Algorithm \ref{algo_asyn_ARAPS}.


%
\section{Convergence Analysis}\label{sec:convergence_analysis}

We show in this section that the sequence of objective function values $F(\bbx^t)$ generated by RAPSA approaches the optimal objective function value $F(\bbx^{*})$. We further show that the convergence guarantees for synchronous RAPSA generalize to the asynchronous setting. In establishing this result we define the set $\ccalS^t$ corresponding to the components of the vector $\bbx$ associated with the blocks selected at step $t$ defined by indexing set $\ccalI^t \subset \{1,\dots,B\}$. Note that components of the set $\ccalS^t$ are chosen uniformly at random from the set of blocks $\{\bbx_1,\dots,\bbx_B\}$. With this definition, due to convenience for analyzing the proposed methods, we rewrite the time evolution of the RAPSA iterates (Algorithm \ref{algo_RAPSA}) as
\begin{align} \label{eq:block_stochastic_gradient}
   {\bbx^{t+1}_{i}  \ = \ \bbx^{t}_{i}  - \gamma^t\  \nabla_{\bbx_i} f( \bbx^{t}, \bbTheta_i^t) \qquad \forall \ \bbx_i\in \ccalS^t,} 
\end{align}
while the rest of the blocks remain unchanged, i.e., $ \bbx^{t+1}_{i}=\bbx^{t}_{i}$ for $\bbx_i\notin \ccalS^t$. Since the number of updated blocks is equal to the number of processors, the ratio of updated blocks is $r:=|\ccalI^t|/B=I/B$. To prove convergence of RAPSA, we require the following assumptions.

%
\begin{assumption}\label{convexity_assumption} 
The instantaneous objective functions $f(\bbx,\bbtheta)$ are differentiable and the average function $F(\bbx)$ is strongly convex with parameter $m>0$.
%
\end{assumption}

%
\begin{assumption}\label{Lipschitz_assumption} The average objective function gradients $\nabla F(\bbx)$ are Lipschitz continuous with respect to the Euclidian norm with parameter $M$, i.e., for all $\bbx, \hbx \in \reals^p$, it holds that
\begin{equation}
   \| \nabla F(\bbx)-\nabla F(\hbx) \| \ \leq\  M\ \| \bbx- \hbx \|.
\end{equation}
\end{assumption}

%
\begin{assumption}\normalfont\label{ass_bounded_stochastic_gradient_norm} The second moment of the norm of the stochastic gradient is bounded for all $\bbx$, i.e., there exists a constant $K$ such that for all variables $\bbx$, it holds
\begin{equation}\label{ekhtelaf}
   \mbE_{\bbtheta} \big{[} \|\nabla f( \bbx^t, \bbtheta^t)\|^{2} \given{\bbx^t}\big{]} \leq K.
\end{equation} \end{assumption}


Notice that Assumption \ref{convexity_assumption} only enforces strong convexity of the average function $F$, while the instantaneous functions $f_i$ may not be even convex. Further, notice that since the instantaneous functions $f_i$ are differentiable the average function $F$ is also differentiable. The Lipschitz continuity of the average function gradients $\nabla F$ is customary in proving objective function convergence for descent algorithms. The restriction imposed by Assumption \ref{ass_bounded_stochastic_gradient_norm} is a standard condition in stochastic approximation literature \citep{robbins1951}, its intent being to limit the variance of the stochastic gradients \citep{Nemirovski}. 

\subsection{Convergence of RAPSA}

We turn our attention to the random parallel stochastic algorithm defined in \eqref{block_sto_grad}-\eqref{eq:block_stochastic_gradient_1} in Section \ref{sec:rapsa}, establishing performances guarantees in both the diminishing and constant algorithm step-size regimes. Our first result comes in the form of a expected descent lemma that relates the expected difference of subsequent iterates to the gradient of the average function.

\begin{lemma}\label{exp_wrt_blocks}
Consider the random parallel stochastic algorithm defined in \eqref{block_sto_grad}-\eqref{eq:block_stochastic_gradient_1}. Recall the definitions of the set of updated blocks $\ccalI^t$ which are randomly chosen from the total $B$ blocks. 
Define $\ccalF^t$ as a sigma algebra that measures the history of the system up until time $t$. Then, the expected value of the difference $\bbx^{t+1}-\bbx^{t}$ with respect to the random set $\ccalI^t$ given $\ccalF^t$ is 
\begin{equation}\label{lemma_RAPS_dec_claim_1}
\mathbb{E}_{\ccalI^t}\!\left[{\bbx^{t+1}-\bbx^{t}\mid \ccalF^t}\right] =
	 - r\gamma^t \ \nabla f( \bbx^{t}, \bbTheta^t).
\end{equation}
Moreover, the expected value of the squared norm $\|\bbx^{t+1}-\bbx^{t}\|^2$ with respect to the random set $\ccalI^t$ given $\ccalF^t$ can be simplified as
\begin{equation}\label{lemma_RAPS_dec_claim_2}
\mathbb{E}_{\ccalI^t}\!\left[\|{\bbx^{t+1}-\bbx^{t}\|^2\mid \ccalF^t}\right] = 
	 {r(\gamma^t)^2 }\ \left\|\nabla f( \bbx^{t}, \bbTheta^t)\right\|^2.
\end{equation}
\end{lemma}

\begin{proof}
See Appendix \ref{apx_lemma_exp_wrt_blocks}.
\end{proof}

Notice that in the regular stochastic gradient descent method the difference of two consecutive iterates $\bbx^{t+1}-\bbx^{t}$ is equal to the stochastic gradient $\nabla f( \bbx^{t}, \bbTheta^t)$ times the stepsize $\gamma^t$. Based on the first result in Lemma \ref{exp_wrt_blocks}, the expected value of stochastic gradients with respect to the random set of blocks $\ccalI^t$ is the same as the one for SGD except that it is multiplied by the fraction of updated blocks $r$. Expression in \eqref{lemma_RAPS_dec_claim_2} shows the same relation for the expected value of the squared difference $\|\bbx^{t+1}-\bbx^{t}\|^2$. These relationships confirm that in expectation RAPSA behaves as SGD which allows us to establish the global convergence of RAPSA.

\begin{proposition}\label{martingale_prop}
Consider the random parallel stochastic algorithm defined in \eqref{block_sto_grad}-\eqref{eq:block_stochastic_gradient_1}. 
If Assumptions \ref{convexity_assumption}-\ref{ass_bounded_stochastic_gradient_norm} hold, then the objective function error sequence $F(\bbx^t)-F(\bbx^*)$ satisfies 
\begin{align}\label{martingale_prop_claim}
\mathbb{E}\left[F(\bbx^{t+1})-F(\bbx^*)\mid \ccalF^t\right]\leq 
	\left( 1- {2m r\gamma^t}{} \right)\left( F(\bbx^{t}) -F(\bbx^*)\right)+ \frac{rM  K(\gamma^t)^2}{2 }.
\end{align}
\end{proposition}

\begin{proof}
See Appendix \ref{apx_martingale_prop}.
\end{proof}

Proposition \ref{martingale_prop} leads to a supermartingale relationship for the sequence of objective function errors $F(\bbx^t)-F(\bbx^*)$. In the following theorem we show that if the sequence of stepsize satisfies standard stochastic approximation diminishing step-size rules (non-summable and squared summable), the sequence of objective function errors $F(\bbx^t)-F(\bbx^*)$ converges to null almost surely. Considering the strong convexity assumption this result implies almost sure convergence of the sequence $\| \bbx^{t}-\bbx^{*} \|^{2}$ to null.

\begin{theorem}\label{RAPSA_convg_thm}
Consider the random parallel stochastic algorithm defined in \eqref{block_sto_grad}-\eqref{eq:block_stochastic_gradient_1} (Algorithm \ref{algo_RAPSA}). If Assumptions \ref{convexity_assumption}-\ref{ass_bounded_stochastic_gradient_norm} hold true and the sequence of stepsizes are non-summable $\sum_{t=0}^\infty \gamma^t=\infty$ and square summable $\sum_{t=0}^\infty (\gamma^t)^2<\infty$, then sequence of the variables $\bbx^t$ generated by RAPSA converges almost surely to the optimal argument $\bbx^*$, 
\begin{equation}\label{rapsa_as_convg}
\lim_{t\to \infty} \|\bbx^t-\bbx^*\|^2 \ =\ 0 \qquad \text{a.s.}
\end{equation}
Moreover, if stepsize is defined as $\gamma^t:=\gamma^0T^0/(t+T^0)$ and the stepsize parameters are chosen such that $2mr \gamma^0 T^0>1$, then the expected average function error $\E{F(\bbx^t)-F(\bbx^*)}$ converges to null at least with a sublinear convergence rate of order $\ccalO(1/t)$,
\begin{equation}\label{rapsa_rate}
\E{F(\bbx^t)-F(\bbx^*)} \ \leq \ \frac{ C}{t+T^0},
\end{equation}
where the constant $C$ is defined as 
\begin{equation}\label{Constant_C}
C= \max\left\{\frac{rMK (\gamma^0 T^0)^2}{4mr\gamma^0 T^0-2},\ T^0(F(\bbx^0)-F(\bbx^*))\right\}.
\end{equation}
\end{theorem}
\begin{proof}
See Appendix \ref{apx_RAPSA_convg_thm}.
\end{proof}

The result in Theorem \ref{RAPSA_convg_thm} shows that when the sequence of stepsize is diminishing as $\gamma^t=\gamma^0T^0/(t+T^0)$, the average objective function value $F(\bbx^t)$ sequence converges to the optimal objective value $F(\bbx^*)$ with probability 1. Further, the rate of convergence in expectation is at least in the order of $\ccalO(1/t)$. \footnote{The expectation on the left hand side of \eqref{rapsa_rate}, and throughout the subsequent convergence rate analysis, is taken with respect to the full algorithm history $\ccalF_0$, which all realizations of both $\bbTheta_t$ and $\ccalI_t$ for all $t\geq 0$.} 
Diminishing stepsizes are useful when exact convergence is required, however, for the case that we are interested in a specific accuracy $\eps$ the more efficient choice is using a constant stepsize. In the following theorem we study the convergence properties of RAPSA for a constant stepsize $\gamma^t=\gamma$.

\begin{theorem}\label{RAPSA_convg_thm_finite}
Consider the random parallel stochastic algorithm defined in \eqref{block_sto_grad}-\eqref{eq:block_stochastic_gradient_1} (Algorithm \ref{algo_RAPSA}). If Assumptions \ref{convexity_assumption}-\ref{ass_bounded_stochastic_gradient_norm} hold true and the stepsize is constant $\gamma^t=\gamma$, then a subsequence of the variables $\bbx^t$ generated by RAPSA converges almost surely to a neighborhood of the optimal argument $\bbx^*$ as
\begin{equation}\label{rapsa_as_convg_finite}
\liminf_{t\to \infty}\ F(\bbx^t)-F(\bbx^*)\ \leq\ \frac{\gamma M K}{4m} \qquad \text{a.s.}
\end{equation}
Moreover, if the constant stepsize $\gamma$ is chosen such that $2m r\gamma<1$ then the expected average function value error $\E{F(\bbx^t)-F(\bbx^*)}$ converges \textit{linearly} to an error bound as
\begin{align}\label{rapsa_rate_finite}
\E{F(\bbx^{t})-F(\bbx^*)}
	&\leq 
	\left( 1- {2m\gamma r}{} \right)^{t}( F(\bbx^{0}) -F(\bbx^*))
	+ \frac{\gamma M  K}{4 m}.
\end{align}
\end{theorem}
\begin{proof}
See Appendix \ref{apx_RAPSA_convg_thm_finite}.
\end{proof}

Notice that according to the result in \eqref{rapsa_rate_finite} there exits a trade-off between accuracy and speed of convergence. Decreasing the constant stepsize $\gamma$ leads to a smaller error bound ${\gamma M  K}/{4 m}$ and a more accurate convergence, while the linear convergence constant $\left( 1- {2m\gamma r}{} \right)$ increases and the convergence rate becomes slower. Further, note that the error of convergence ${\gamma M  K}/{4 m}$ is independent of the ratio of updated blocks $r$, while the constant of linear convergence $1- {2m\gamma r}$ depends on $r$. Therefore, updating a fraction of the blocks at each iteration decreases the speed of convergence for RAPSA relative to SGD that updates all of the blocks, however, both of the algorithms reach the same accuracy. 

To achieve accuracy $\eps$ the sum of two terms in the right hand side of \eqref{rapsa_rate_finite} should be smaller than $\eps$. Let's consider $\phi$ as a positive constant that is strictly smaller than $1$, i.e., $0<\phi<1$. Then, we want to have
\begin{equation}\label{condition_on_gamma}
 \frac{\gamma M  K}{4 m}\leq \phi \eps , \quad\!\! \left( 1- {2m\gamma r}{} \right)^{t}( F(\bbx^{0}) -F(\bbx^*)) \leq (1-\phi)\eps.
\end{equation}
Therefore, to satisfy the first condition in \eqref{condition_on_gamma} we set the stepsize as $\gamma=4m\phi\eps/MK$. Apply this substitution into the second inequality in  \eqref{condition_on_gamma} and consider the inequality $a+\ln(1-a)<0$  for $0<a<1$, to obtain that 
\begin{equation}\label{num_ite}
t\geq\frac{MK}{8m^2r\phi \eps}\ln\left(\frac{ F(\bbx^{0}) -F(\bbx^*)}{(1-\phi)\eps}\right).
\end{equation}
The lower bound in \eqref{num_ite} shows the minimum number of required iterations for RAPSA to achieve accuracy $\eps$.

\subsection{Convergence of Asynchronous RAPSA}\label{sec:asyn_rapsa_convg}

In this section, we study the convergence of Asynchronous RAPSA (Algorithm \ref{algo_asyn}) developed in Section \ref{sec:asyn} and we characterize the effect of delay in the asynchronous implementation. To do so, the following condition on the delay $\tau$ is required. 

%
\begin{assumption}\label{delay_assumption} 
The random variable $\tau$ which is the delay between reading and writing for processors does not exceed the constant $\Delta$, i.e., 
\begin{equation}\label{delay_bounds}
\tau\leq \Delta. 
\end{equation}
\end{assumption}

The condition in Assumption \ref{delay_assumption} implies that processors can finish their tasks in a time that is bounded by the constant $\Delta.$ This assumption is typical in the analysis of asynchronous algorithms.

To establish the convergence properties of asynchronous RAPSA recall the set $\ccalS^t$ containing the blocks that are updated at step $t$ with associated indices $\ccalI^t \subset \{1,\dots,B\}$. Therefore, the update of asynchronous RAPSA can be written as 
\begin{align} \label{eq:block_stochastic_gradient_asyn_2}
   {\bbx^{t+1}_{i}  \ = \ \bbx^{t}_{i}  - \gamma^t\  \nabla_{\bbx_i} f( \bbx^{t-\tau}, \bbTheta_i^{t-\tau}) \qquad \forall \ \bbx_i\in \ccalS^t,} 
\end{align}
and the rest of the blocks remain unchanged, i.e., $ \bbx^{t+1}_{i}=\bbx^{t}_{i}$ for $\bbx_i\notin \ccalS^t$. 

Note that the random set $\ccalI^t$ and the associated block set $\ccalS^t$ are chosen at time $t-\tau$ in practice; however, for the sake of analysis we can assume that these sets are chosen at time $t$. In other words, we can assume that at step $t-\tau$ processor $i$ computes the full (for all blocks) stochastic gradient  $\nabla f( \bbx^{t-\tau}, \bbTheta_i^{t-\tau})$ and after finishing this task at time $t$, it chooses uniformly at random the block that it wants to update. Thus, the block $\bbx_i$ in \eqref{eq:block_stochastic_gradient_asyn_2} is chosen at step $t$. This new interpretation of the update of asynchronous RAPSA is only important for the convergence analysis of the algorithm and we use it in the proof of following lemma which is similar to the result in Lemma \ref{exp_wrt_blocks} for synchronous RAPSA.

\begin{lemma}\label{exp_wrt_blocks_asyn}
Consider the asynchronous random parallel stochastic algorithm (Algorithm \ref{algo_asyn}) defined in \eqref{eq:block_stochastic_gradient_asyn}. Recall the definitions of the set of updated blocks $\ccalI^t$ which are randomly chosen from the total $B$ blocks. 
Define $\ccalF^t$ as a sigma algebra that measures the history of the system up until time $t$. Then, the expected value of the difference $\bbx^{t+1}-\bbx^{t}$ with respect to the random set $\ccalI^{t}$ given $\ccalF^{t}$ is 
\begin{equation}\label{lemma_RAPS_dec_claim_1_asyn}
\mathbb{E}_{\ccalI^t}\!\left[{\bbx^{t+1}-\bbx^{t}\mid \ccalF^{t}}\right] =
	 - \frac{\gamma^t}{B} \ \nabla f( \bbx^{t-\tau}, \bbTheta^{t-\tau}).
\end{equation}
Moreover, the expected value of the squared norm $\|\bbx^{t+1}-\bbx^{t}\|^2$ with respect to the random set $\ccalS^t$ given $\ccalF^t$ satisfies the identity
\begin{equation}\label{lemma_RAPS_dec_claim_2_asyn}
\mathbb{E}_{\ccalI^t}\!\left[\|{\bbx^{t+1}-\bbx^{t}\|^2\mid \ccalF^{t}}\right] = 
	 {\frac{(\gamma^t)^2}{B} }\ \left\|\nabla f( \bbx^{t-\tau}, \bbTheta^{t-\tau})\right\|^2.
\end{equation}
\end{lemma}

\begin{myproof}
See Appendex \ref{apx_exp_wrt_blocks_asyn}.
\end{myproof}

The results in Lemma \ref{exp_wrt_blocks_asyn} is a natural extension of the results in Lemma \ref{exp_wrt_blocks} for the lock-free setting, since in the asynchronous scheme only one of the blocks is updated at each iteration and the ratio $r$ can be simplified as $1/B$. We use the result in Lemma \ref{exp_wrt_blocks_asyn} to characterize the decrement in the expected sub-optimality in the following proposition. 

\begin{proposition}\label{martingale_prop_asyn}
Consider the asynchronous random parallel stochastic algorithm defined in \eqref{eq:block_stochastic_gradient_asyn} (Algorithm \ref{algo_asyn}) . 
If Assumptions \ref{convexity_assumption}-\ref{ass_bounded_stochastic_gradient_norm} hold, then for any arbitrary $\rho>0$ we can write that the objective function error sequence $F(\bbx^t)-F(\bbx^*)$ satisfies 
\begin{align}\label{martingale_prop_claim_asyn}
&\mathbb{E}\left[F(\bbx^{t+1})-F(\bbx^*)\mid \ccalF^{t-\tau}\right]\nonumber\\
&\quad	\leq 
	\left( 1-\frac{2m \gamma^t}{B}  \left[1-\frac{\rho M}{2}\right] \right)\mathbb{E}\left[F(\bbx^{t}) -F(\bbx^*)\mid \ccalF^{t-\tau}\right]
	+ \frac{M  K(\gamma^t)^2}{2B }+  \frac{\tau^2 MK \gamma^t (\gamma^{t-\tau})^2}{2\rho B^2}.
\end{align}
\end{proposition}

\begin{myproof}
See Appendix \ref{apx_RAPSA_convg_thm_finite_asyn}.
\end{myproof}

We proceed to use the result in Proposition \ref{martingale_prop_asyn} to prove that the sequence of iterates generated by asynchronous RAPSA converges to the optimal argument $\bbx^*$ defined by \eqref{eq:block_stoch_opt}.

\begin{theorem}\label{RAPSA_convg_thm_asyn}
Consider the asynchronous RAPSA defined in \eqref{eq:block_stochastic_gradient_asyn} (Algorithm \ref{algo_asyn}) . If Assumptions \ref{convexity_assumption}-\ref{ass_bounded_stochastic_gradient_norm} hold true and the sequence of stepsizes are non-summable $\sum_{t=0}^\infty \gamma^t=\infty$ and square summable $\sum_{t=0}^\infty (\gamma^t)^2<\infty$, then sequence of the variables $\bbx^t$ generated by RAPSA converges almost surely to the optimal argument $\bbx^*$, 
\begin{equation}\label{rapsa_as_convg_asyn}
\liminf_{t\to \infty} \|\bbx^t-\bbx^*\|^2 \ =\ 0 \qquad \text{a.s.}
\end{equation}
Moreover, if stepsize is defined as $\gamma^t:=\gamma^0T^0/(t+T^0)$ and the stepsize parameters are chosen such that $(2m \gamma^0 T^0/B)(1-\rho M/2)>1$, then the expected average function error $\E{F(\bbx^t)-F(\bbx^*)}$ converges to null at least with a sublinear convergence rate of order $\ccalO(1/t)$,
\begin{equation}\label{rapsa_rate}
\E{F(\bbx^t)-F(\bbx^*)} \ \leq \ \frac{ C}{t+T^0},
\end{equation}
where the constant $C$ is defined as 
\begin{equation}\label{Constant_C2}
C= \max \left\{\frac{M  K(\gamma^0 T^0)^2/2B+(\tau^2 MK (\gamma^0 T^0)^3)(2\rho B^2)}{(2m \gamma^0 T^0/B)(1-\rho M/2)-1},\ T^0(F(\bbx^0)-F(\bbx^*))\right\}.
\end{equation}
\end{theorem}


\begin{myproof}
See Appendix \ref{apx_RAPSA_convg_thm_asyn}.
\end{myproof}

Theorem \ref{RAPSA_convg_thm_asyn} establishes that the RAPSA algorithm when run on a lock-free computing architecture, still yields convergence to the optimal argument $\bbx^*$ defined by \eqref{eq:block_stoch_opt}. Moreover, the expected objective error sequence converges to null as $\ccalO(1/t)$. These results, which correspond to the diminishing step-size regime, are comparable to the performance guarantees (Theorem \ref{RAPSA_convg_thm}) previously established for RAPSA on a synchronous computing cluster, meaning that the algorithm performance does not degrade significantly when implemented on an asynchronous system. This issue is explored numerically in Section \ref{sec:simulations}.

\section{Numerical analysis}
\label{sec:simulations}

In this section we study the numerical performance of the doubly stochastic approximation algorithms developed in Sections \ref{sec:rapsa}-\ref{sec:asyn} by first considering a linear regression problem. We then use RAPSA to develop a visual classifier to distinguish between distinct hand-written digits. 

\subsection{Linear Regression}\label{subsec:lmmse}

We consider a setting in which observations $\bbz_n\in\reals^q$ are collected which are noisy linear transformations $\bbz_{n}=\bbH_n \bbx + \bbw_{n}$ of a signal $\bbx\in\reals^p$ which we would like to estimate, and $\bbw \sim \ccalN(0, \sigma^2 I_q)$ is a Gaussian random variable. For a finite set of samples $N$, the optimal $\bbx^*$ is computed as the least squares estimate $\bbx^*:=\argmin_{\bbx\in\reals^p } ({1}/{N})\sum_{n=1}^N \|\bbH_{n}\bbx- \bbz_{n}\|^2$. We run RAPSA on LMMSE estimation problem instances where $q=1$, $p=1024$, and $N=10^4$ samples are given. The observation matrices $\bbH_n \in \reals^{q\times p}$, when stacked over all $n$ (an $N \times p$ matrix), are generated from a matrix normal distribution whose mean is a tri-diagonal matrix. The main diagonal is $2$, while the super and sub-diagonals are all set to $-1/2$. Moreover, the true signal has entries chosen uniformly at random from the fractions $\bbx\in \{1,\dots,p\}/p$. Additionally, the noise variance perturbing the observations is set to $\sigma^2=10^{-2}$. We assume that the number of processors $I=16$ is fixed and each processor is in charge of $1$ block. We consider different number of blocks $B=\{16,32,64,128\}$. Note that when the number of blocks is $B$, there are $p/B=1024/B$ coordinates in each block. 
\begin{figure}\centering
\subfigure[Excess Error $F(\bbx^t) - F(\bbx^*)$ vs. iteration $t$]{
\includegraphics[width=0.48\linewidth,height=0.28\linewidth]
{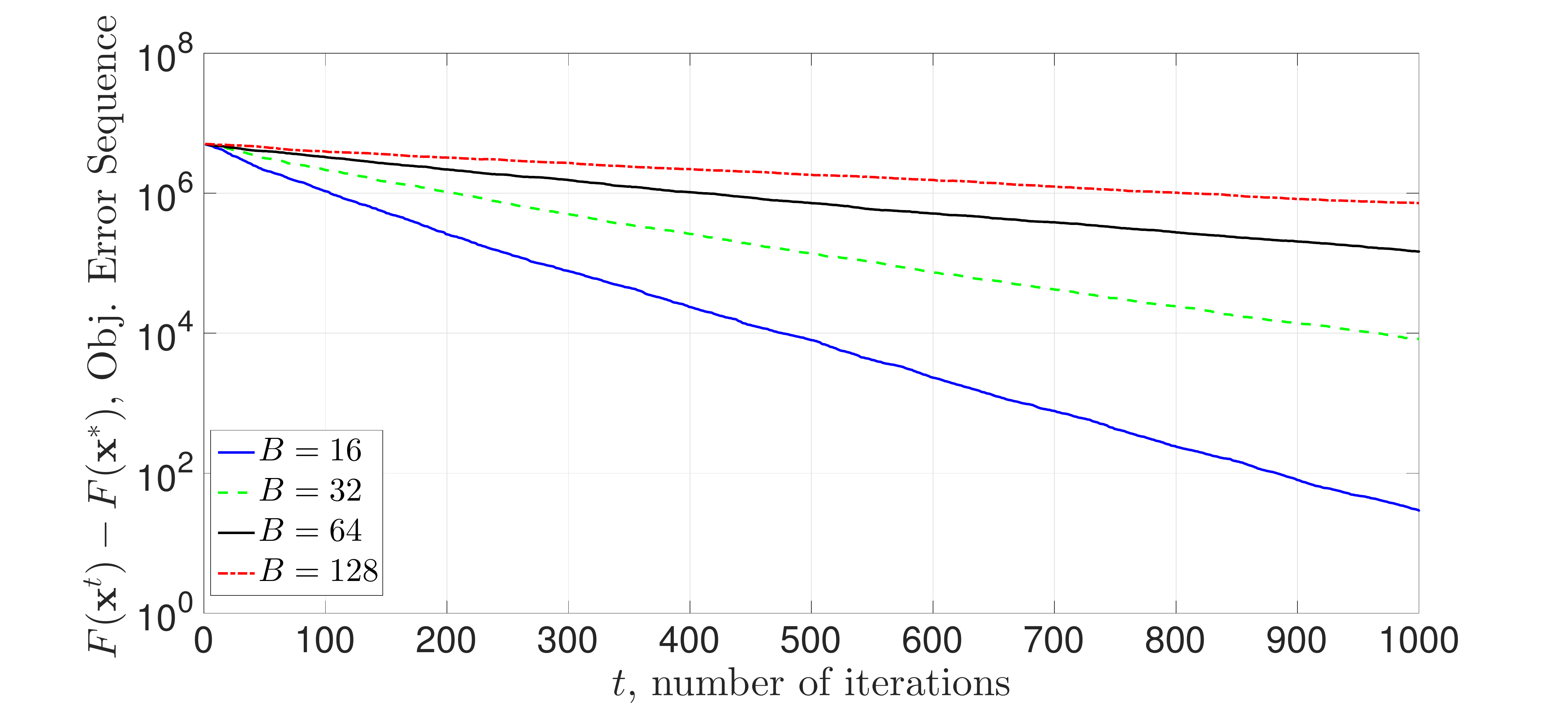}\label{subfig:rapsa_linear_constant_a}}
\subfigure[Excess Error $F(\bbx^t) - F(\bbx^*)$ vs. feature $\tilde{p}_t$]{
\includegraphics[width=0.48\linewidth,height=0.28\linewidth]
{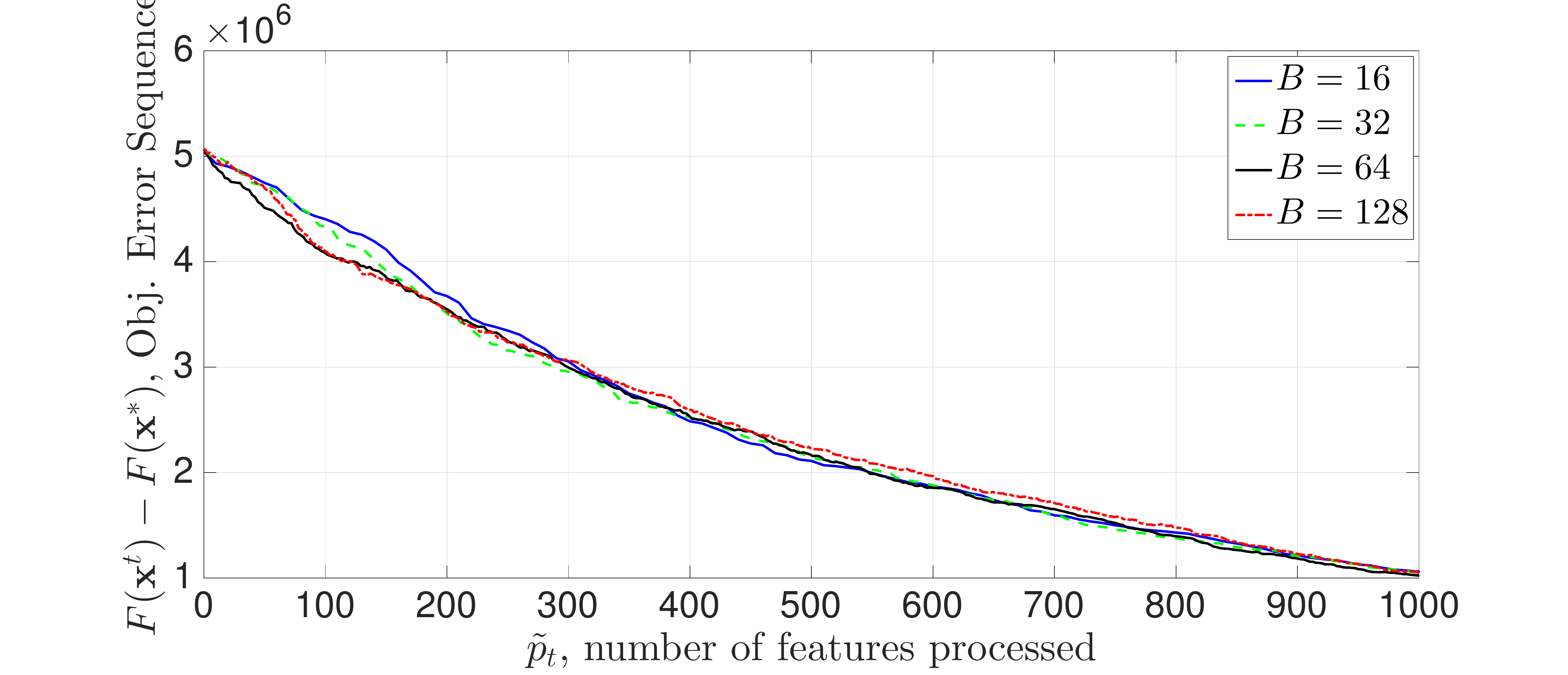}\label{subfig:rapsa_linear_constant_b}}
\caption{RAPSA on a linear regression (quadratic minimization) problem with signal dimension $p=1024$ for $N=10^3$ iterations with mini-batch size $L=10$ for different number of blocks $B=\{16,32,64,128\}$ initialized as $10^4 \times \bbone$. We use constant step-size $\gamma^t= \gamma=10^{-2}$. Convergence is in terms of number of iterations is best when the number of blocks updated per iteration is equal to the number of processors ($B=16$, corresponding to parallelized SGD), but comparable across the different cases in terms of number of features processed. This shows that there is no price payed in terms of convergence speed for reducing the computation complexity per iteration.} \label{fig:rapsa_linear_constant}
\end{figure}

{\bf Results for RAPSA} We first consider the algorithm performance of RAPSA (Algorithm \ref{algo_RAPSA}) when using a constant step-size $\gamma^t=\gamma=10^{-2}$. The size of mini-batch is set as $L=10$ in the subsequent experiments. To determine the advantages of incomplete randomized parallel processing, we vary the number of coordinates updated at each iteration. In the case that $B=16$, $B=32$, $B=64$, and $B=128$, in which case the number of updated coordinates per iteration are $1024$, $512$, $256$, and $128$, respectively. Notice that the case that $B=16$ can be interpreted as parallel SGD, which is mathematically equivalent to Hogwild! \citep{Niu11hogwild:a}, since all the coordinates are updated per iteration, while in other cases $B>16$ only a subset of $1024$ coordinates are updated. 

 Fig. \ref{subfig:rapsa_linear_constant_a} illustrates the convergence path of RAPSA's objective error sequence defined as $F(\bbx^t) - F(\bbx^*)$ with $F(\bbx) = ({1}/{N})\sum_{n=1}^N \|\bbH_{n}\bbx- \bbz_{n}\|^2 $
  as compared with the number of iterations $t$. In terms of iteration $t$, we observe that the algorithm performance is best when the number of processors equals the number of blocks, corresponding to parallelized stochastic gradient method. However, comparing algorithm performance over iteration $t$ across varying numbers of blocks updates is unfair. If RAPSA is run on a problem for which $B=32$, then at iteration $t$ it has only processed {\it half} the data that parallel SGD, i.e., $B=16$, has processed by the same iteration. Thus for completeness we also consider the algorithm performance in terms of number of features processed $\tdp_t$ which is given by $\tdp_t=ptI/B$.


\begin{figure}\centering
\subfigure[Excess Error $F(\bbx^t) - F(\bbx^*)$ vs. iteration $t$]{
\includegraphics[width=0.48\linewidth,height=0.28\linewidth]
{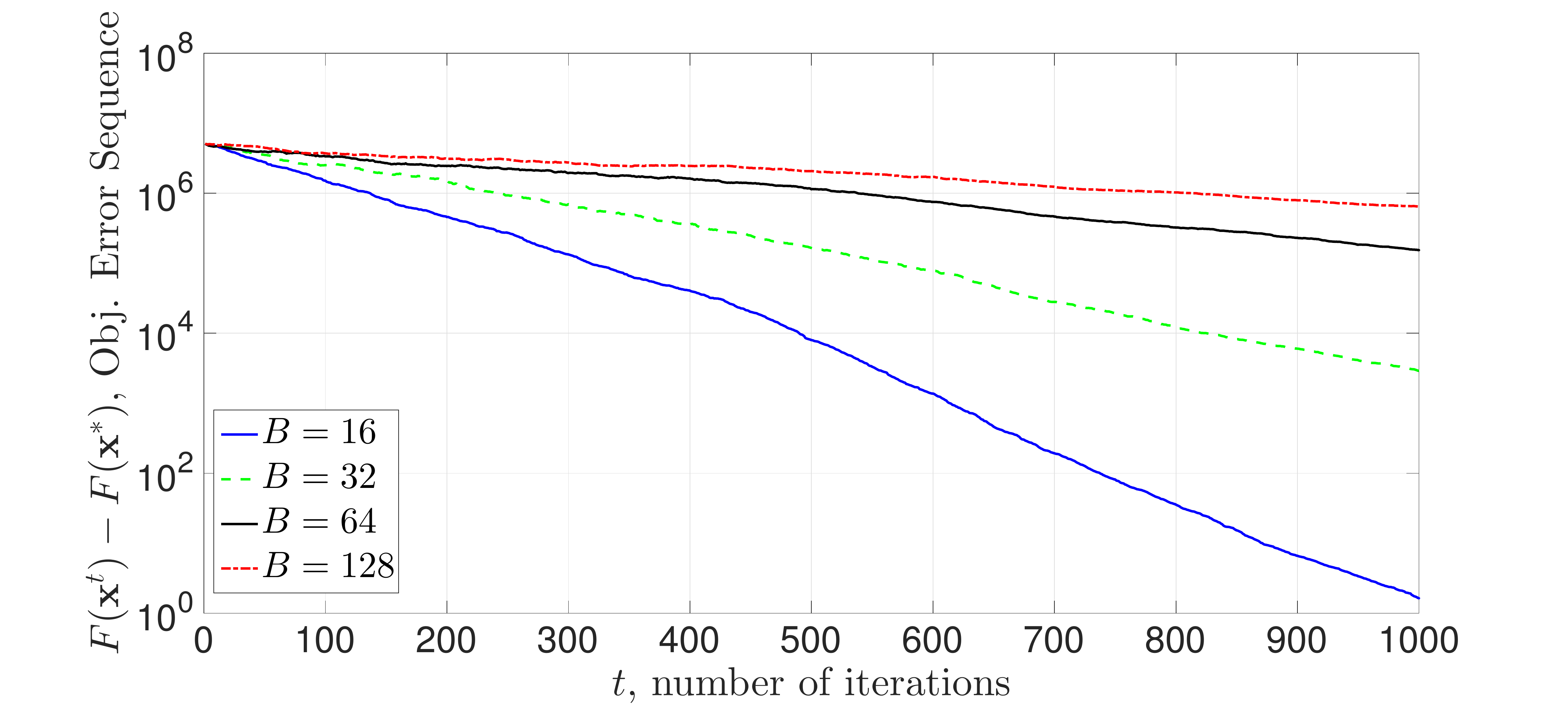}\label{subfig:rapsa_linear_hybrid_a}}
\subfigure[Excess Error $F(\bbx^t) - F(\bbx^*)$ vs. feature $\tilde{p}_t$]{
\includegraphics[width=0.48\linewidth,height=0.28\linewidth]
{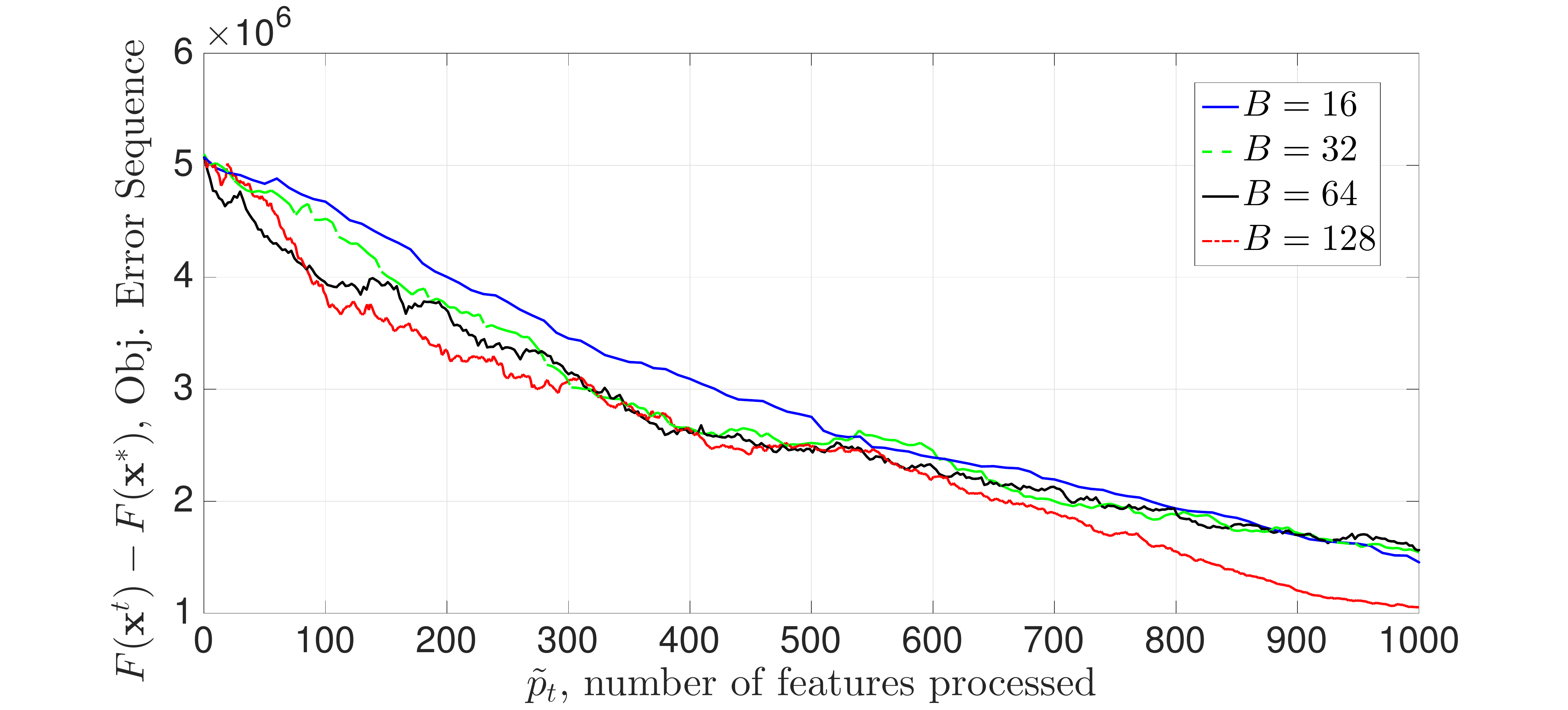}\label{subfig:rapsa_linear_hybrid_b}}
\caption{RAPSA on a linear regression problem with signal dimension $p=1024$ for $N=10^3$ iterations with mini-batch size $L=1$0 for different number of blocks $B=\{16,32,64,128\}$ using initialization $\bbx_0 = 10^4 \times \bbone$. We use hybrid step-size $\gamma^t= \min(10^{-1.5},10^{-1.5} \tilde{T}_0/t)$ with annealing rate $\tilde{T}_0=400$. Convergence is faster with smaller $B$ which corresponds to the proportion of blocks updated per iteration $r$  closer to $1$ in terms of number of iterations. Contrarily, in terms of number of features processed $B=128$ has the best performance and $B=16$ has the worst performance. This shows that updating less features/coordinates per iterations can lead to faster convergence in terms of number of processed features.} \label{fig:rapsa_linear_hybrid}
\end{figure}

In Fig. \ref{subfig:rapsa_linear_constant_b}, we display the convergence of the excess mean square error $F(\bbx^t) - F(\bbx^*)$ in terms of number of features processed $\tilde{p}_t$. In doing so, we may clearly observe the advantages of updating fewer features/coordinates per iteration. Specifically, the different algorithms converge in a nearly identical manner, but RAPSA with $I<<B$ may be implemented without any complexity bottleneck in the dimension of the decision variable $p$ (also the dimension of the feature space).

We observe a comparable trend when we run RAPSA with a hybrid step-size scheme $\gamma^t= \min(\eps, \eps\tilde{T}_0/t)$ which is a constant $\eps=10^{-1.5}$ for the first $\tilde{T}_0=400$ iterations, after which it diminishes as $O(1/t)$. We again observe in Figure \ref{subfig:rapsa_linear_hybrid_a} that convergence is fastest in terms of excess mean square error versus iteration $t$ when all blocks are updated at each step. However, for this step-size selection, we see that updating fewer blocks per step is \emph{faster} in terms of number of features processed. 
%
%
This result shows that updating fewer coordinates per iteration yields convergence gains in terms of number of features processed. This advantage comes from the advantage of Gauss-Seidel style block selection schemes in block coordinate methods as compared with Jacobi schemes. In particular, it's well understood that for problems settings with specific conditioning, cyclic block updates are superior to parallel schemes, and one may respectively interpret RAPSA as compared to parallel SGD as executing variants of cyclic or parallel block selection schemes. We note that the magnitude of this gain is dependent on the condition number of the Hessian of the expected objective $F(\bbx)$.


{\bf Results for Accelerated RAPSA} We now study the benefits of incorporating approximate second-order information about the objective $F(\bbx)$ into the algorithm in the form of ARAPSA (Algorithm \ref{algo_ARAPS}). We first run ARAPSA for the linear regression problem outlined above when using a constant step-size $\gamma^t=\gamma=10^{-2}$ with fixed mini-batch size $L=10$. Moreover, we again vary the number of blocks as  $B=16$, $B=32$, $B=64$, and $B=128$, corresponding to updating all, half, one-quarter, and one-eighth  of the elements of vector $\bbx$ per iteration, respectively. 


\begin{figure}\centering
\subfigure[Excess Error $F(\bbx^t) - F(\bbx^*)$ vs. iteration $t$]{
\includegraphics[width=0.48\linewidth,height=0.28\linewidth]
{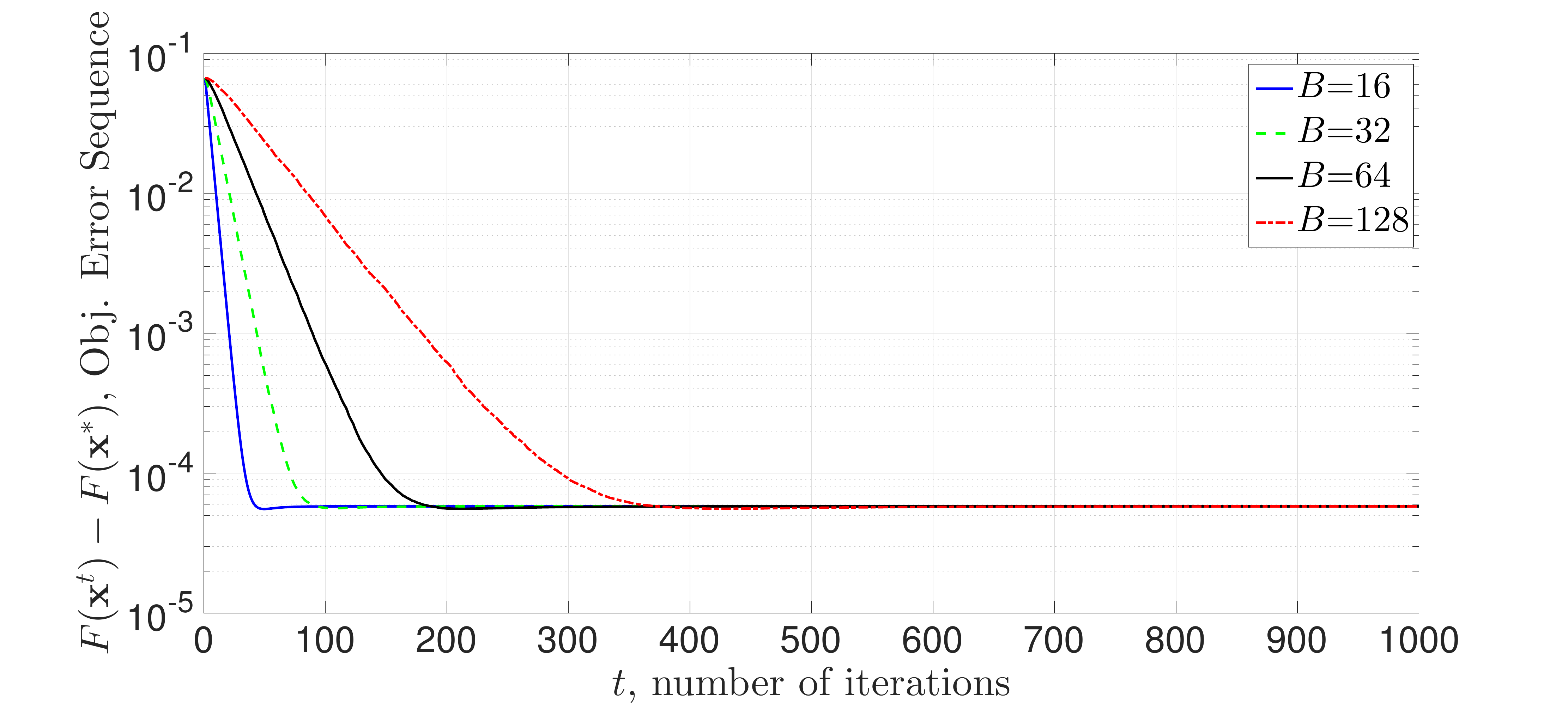}\label{subfig:arapsa_linear_constant_a}}
\subfigure[Excess Error $F(\bbx^t) - F(\bbx^*)$ vs. feature $\tilde{p}_t$]{
\includegraphics[width=0.48\linewidth,height=0.28\linewidth]
{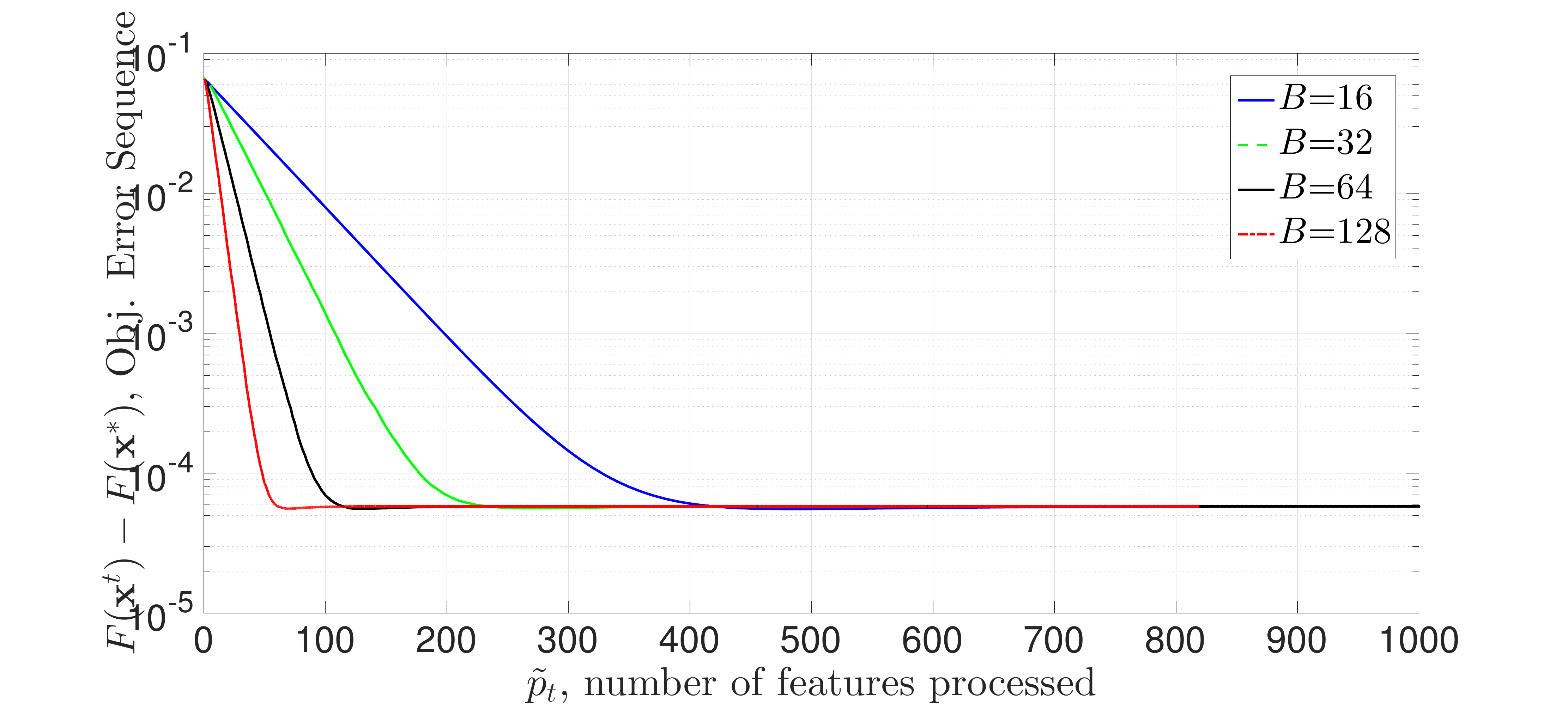}\label{subfig:arapsa_linear_constant_b}}
\caption{ARAPSA on a linear regression problem with signal dimension $p=1024$ for $N=10^3$ iterations with mini-batch size $L=10$ for different number of blocks $B=\{16,32,64,128\}$. We use constant step-size $\gamma^t= \gamma=10^{-1}$ using initialization $10^4 \times \bbone$. Convergence is comparable across the different cases in terms of number of iterations, but in terms of number of features processed $B=128$ has the best performance and $B=16$ (corresponding to parallelized oL-BFGS) converges slowest. We observe that using fewer coordinates per iterations leads to faster convergence in terms of number of processed elements of $\bbx$.} \label{fig:arapsa_linear_constant}
\end{figure}

 Fig. \ref{subfig:arapsa_linear_constant_a} displays the convergence path of ARAPSA's excess mean-square error $F(\bbx^t) - F(\bbx^*)$ versus the number of iterations $t$. We observe that parallelized oL-BFGS ($I=B$) converges fastest in terms of iteration index $t$. On the contrary, in Figure \ref{subfig:arapsa_linear_constant_b}, we may clearly observe that larger $B$, which corresponds to using \emph{fewer} elements of $\bbx$ per step, converges faster in terms of number of features processed. The Gauss-Seidel effect is more substantial for ARAPSA as compared with RAPSA due to the fact that the $\argmin$ of the instantaneous objective computed in block coordinate descent is better approximated by its second-order Taylor-expansion (ARAPSA, Algorithm \ref{algo_ARAPS}) as compared with its linearization (RAPSA, Algorithm \ref{algo_RAPSA}).


\begin{figure}\centering
\subfigure[Excess Error $F(\bbx^t) - F(\bbx^*)$ vs. iteration $t$]{
\includegraphics[width=0.48\linewidth,,height=0.28\linewidth]
{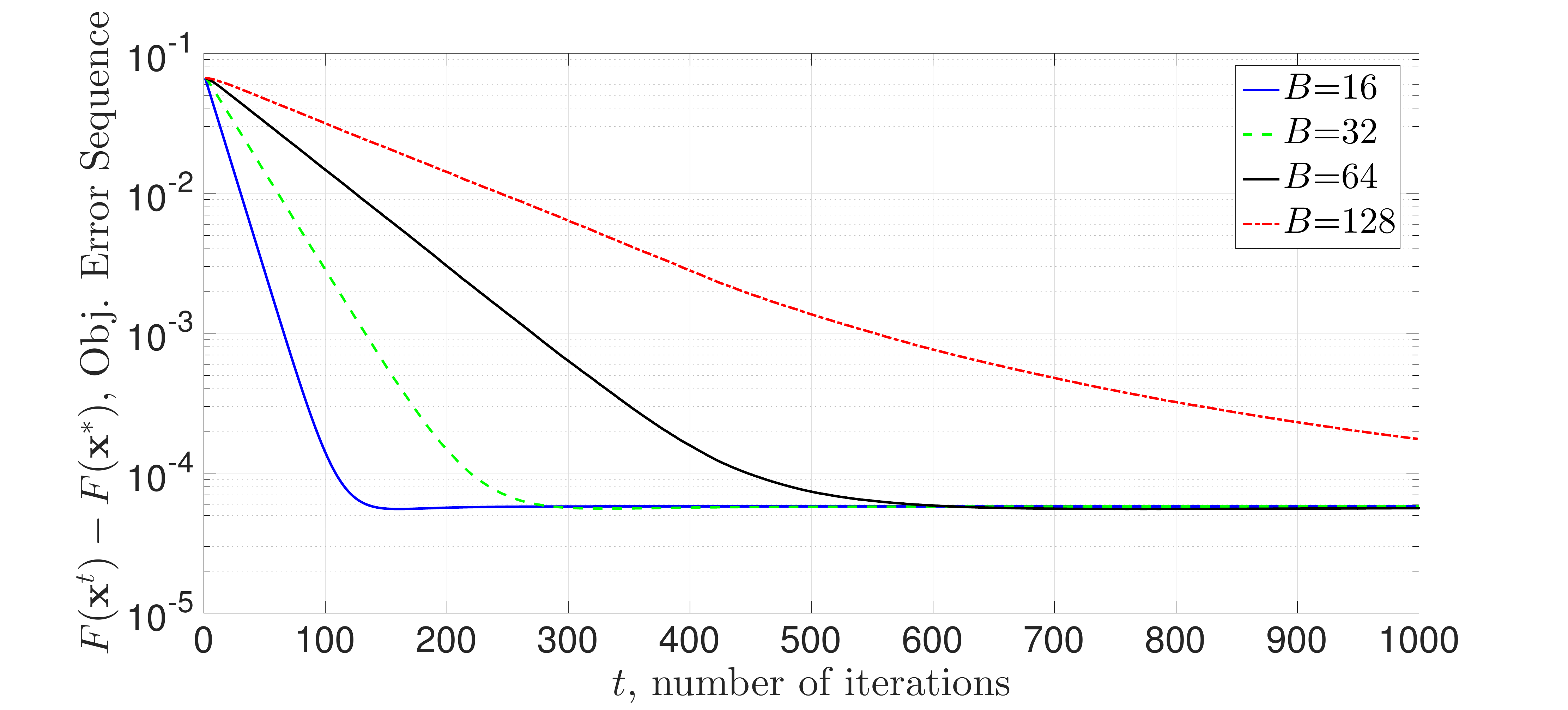}\label{subfig:arapsa_linear_hybrid_a}}
\subfigure[Excess Error $F(\bbx^t) - F(\bbx^*)$ vs. feature $\tilde{p}_t$]{
\includegraphics[width=0.48\linewidth,height=0.28\linewidth]
{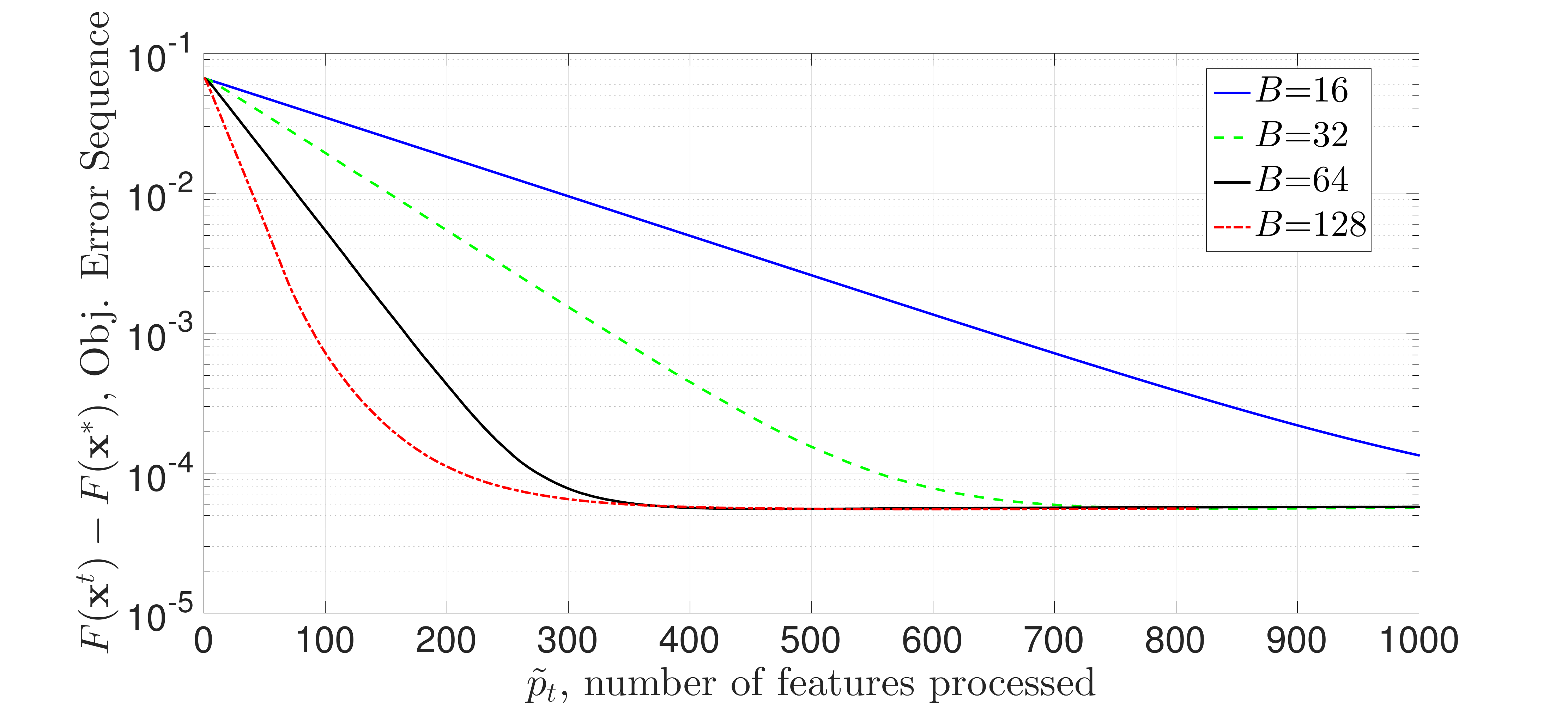}
\label{subfig:arapsa_linear_hybrid_b}}
\caption{ARAPSA on a linear regression problem with signal dimension $p=1024$ for $N=10^4$ iterations with mini-batch size $L=10$ for different number of blocks $B=\{16,32,64,128\}$. We use hybrid step-size $\gamma^t= \min(10^{-1.5},10^{-1.5} \tilde{T}_0/t)$ with annealing rate $\tilde{T}_0=400$. Convergence is comparable across the different cases in terms of number of iterations, but in terms of number of features processed $B=128$ has the best performance and $B=16$ has the worst performance. This shows that updating less features/coordinates per iterations leads to faster convergence in terms of number of processed features.} \label{fig:arapsa_linear_hybrid}
\end{figure}

We now consider the performance of ARAPSA when a hybrid algorithm step-size is used, i.e. $\gamma^t= \min(10^{-1.5},10^{-1.5} \tilde{T}_0/t)$ with attenuation threshold $\tilde{T}_0=400$. The results of this numerical experiment are given in Figure \ref{fig:arapsa_linear_hybrid}. We observe that the performance gains of ARAPSA as compared to parallelized oL-BFGS apparent in the constant step-size scheme are more substantial in the hybrid setting. That is, in Figure \ref{subfig:arapsa_linear_hybrid_a} we again see that parallelized oL-BFGS is best in terms of iteration index $t$ -- to achieve the benchmark $F(\bbx^t) - F(\bbx^*)\leq 10^{-4}$, the algorithm requires $t=100$, $t=221$, $t=412$, and $t>1000$ iterations for $B=16$, $B=32$, $B=64$, and $B=128$, respectively. However, in terms of $\tilde{p}_t$, the number of elements of $\bbx$ processed, to reach the benchmark $F(\bbx^t) - F(\bbx^*)\leq 0.1$, we require $\tilde{p}_t > 1000$, $\tilde{p}_t=570$, $\tilde{p}_t=281$, and $\tilde{p}_t=203$, respectively, for $B=16$, $B=32$, $B=64$, and $B=128$.


{\bf Comparison of RAPSA and ARAPSA} 
We turn to numerically analyzing the performance of Accelerated RAPSA and RAPSA on the linear estimation problem for the case that parameter vectors $\bbx\in\reals^p$ are $p=500$ dimensional for $N=10^4$ iterations in the constant step-size case $\gamma=10^{-2}$. Both algorithms are initialized as $\bbx_0 = 10^3 \times \bbone$ with mini-batch size $L=10$, and ARAPSA uses the curvature memory level $\tau=10$. The number of processors is fixed again as $I=16$, and the number of blocks is $B=64$, meaning that $r=1/4$ of the elements of $\bbx$ are operated on at each iteration.  

\begin{figure}\centering
\subfigure[Excess Error $F(\bbx^t) - F(\bbx^*)$ vs. iteration $t$]{
\includegraphics[width=0.48\linewidth,,height=0.28\linewidth]
{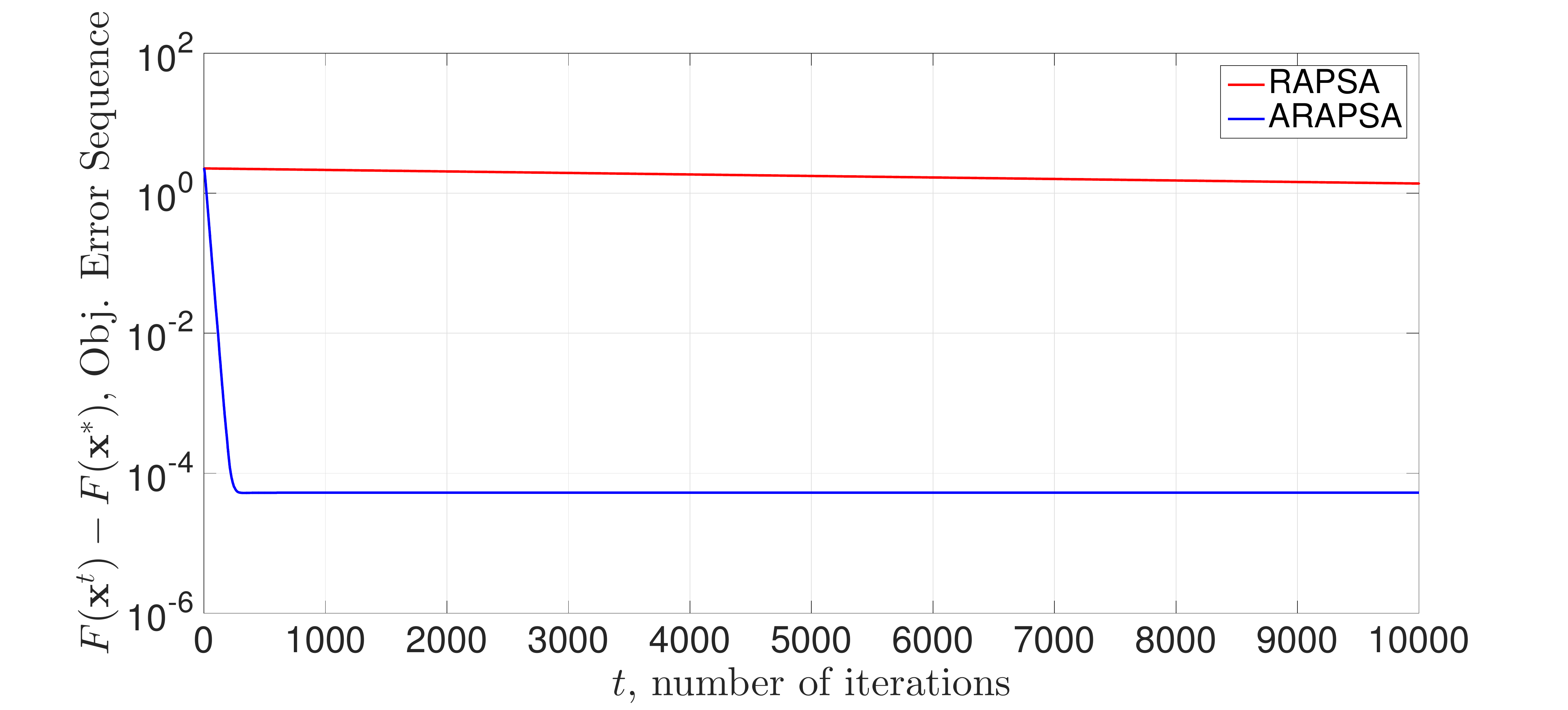}\label{subfig:rapsa_vs_arapsa_linear_a}}
\subfigure[Excess Error $F(\bbx^t) - F(\bbx^*)$ vs. clock time (s)]{
\includegraphics[width=0.48\linewidth,,height=0.28\linewidth]
{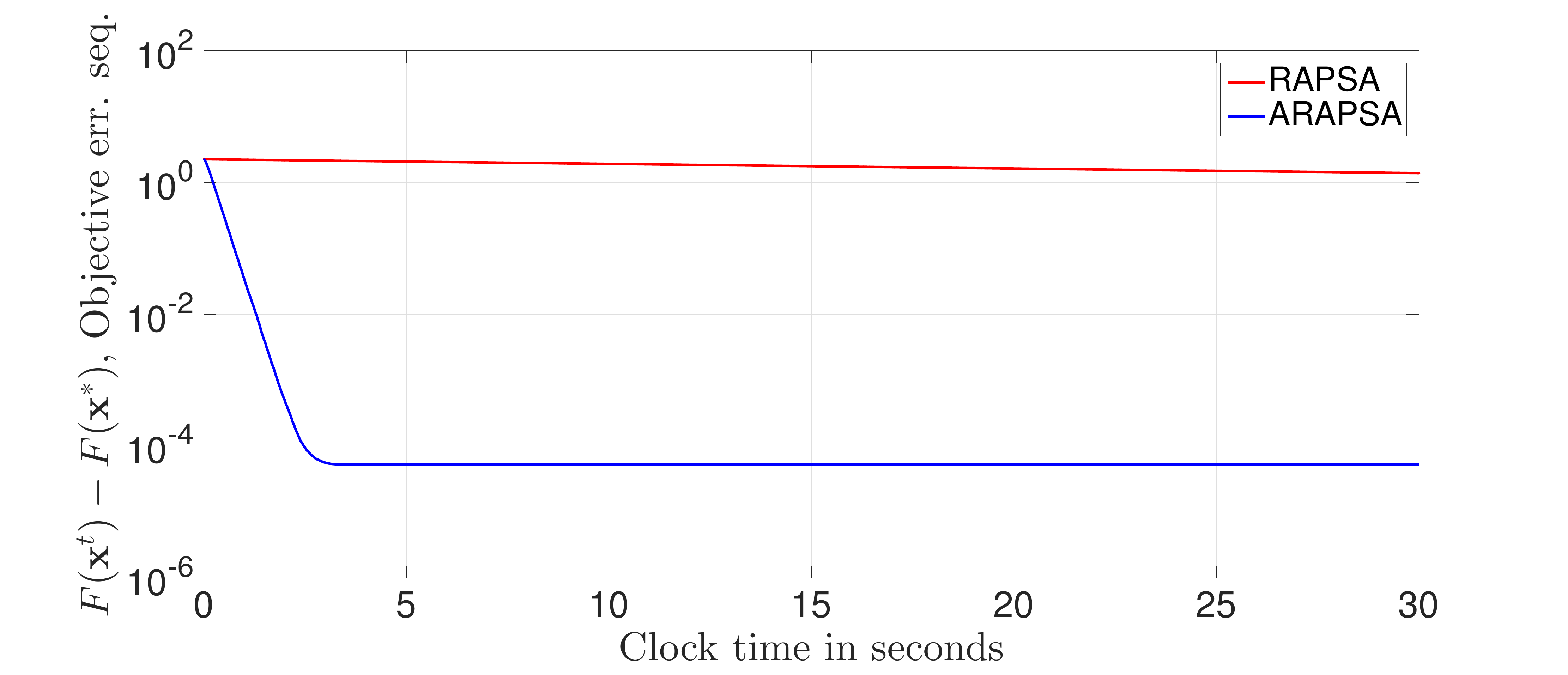}\label{subfig:rapsa_vs_arapsa_linear_b}}
\caption{A numerical comparison of RAPSA and ARAPSA on the linear estimation problem stated at the beginning of Section \ref{subsec:lmmse} for $N=10^4$ iterations with signal dimension $p=500$ with constant step-size $\gamma=10^{-2}$ when there are $I=16$ processors and $B=64$ blocks, meaning that one quarter of the elements of $\bbx$ are updated per iteration. Observe that the rate of convergence for ARAPSA is empirically orders of magnitude higher than RAPSA.} \label{fig:rapsa_vs_arapsa_linear}
\end{figure}

The results of this numerical evaluation are given in Figure \ref{fig:rapsa_vs_arapsa_linear}. We plot the objective error sequence versus iteration $t$ in Figure \ref{subfig:rapsa_vs_arapsa_linear_a}. Observe that ARAPSA converges to within $10^{-4}$ of the optimum by $t=300$ iterations in terms of $F(\bbx^t) - F(\bbx^*)$, whereas RAPSA, while descending slowly, approaches within $10$ of the optimum by $t= 10^4$ iterations. The performance advantages of ARAPSA as compared to RAPSA are also apparent in Figure \ref{subfig:rapsa_vs_arapsa_linear_b}, which readjusts the results of Figure \ref{subfig:rapsa_vs_arapsa_linear_a} to be in terms of \emph{actual} elapsed time. We see that despite the higher complexity of ARAPSA per iteration, its empirical performance results in extremely fast convergence on linear estimation problems. That is, in about $3$ seconds, the algorithm converges to within $10^{-4}$ of the optimal estimator in terms of objective function evaluation.

{\bf Results for Asynchronous RAPSA} 
We turn to studying the empirical performance of the asynchronous variant of RAPSA (Algorithm \ref{algo_asyn}) proposed in Section \ref{subsec:asynchronous}. The model we use for asynchronicity is modeled after a random delay phenomenon in physical communication systems in which each local server has a distinct clock which is not locked to the others. Each processor's clock begins at time $t_0^i=t_0$ for all processors $i=1,\dots,I$ and selects subsequent times as $t_{k} = t_{k-1} + w_k^i$, where $w_k^i \sim \ccalN(\mu,\sigma^2)$ is a normal random variable with mean $\mu$ and variance $\sigma^2$. The variance in this model effectively controls the amount of variability between the clocks of distinct processors. 

We run Asynchronous RAPSA for the linear estimation problem when the parameter vector $\bbx$ is $p=500$ dimensional for $N=10^3$ iterations with no mini-batching $L=1$ for both the case that the algorithm step-size is diminishing and constant step-size regimes for the case that the noise distribution perturbing the collected observations has variance $\sigma^2=10^{-2}$, and the observation matrix is as discussed at the outset of Section \ref{subsec:lmmse}. Further, the algorithm is initialized as $\bbx_0 = 10^3\bbone$. We run the algorithm for a few different instantiations of asynchronicity, that is, $w_k^i \sim \ccalN(\mu,\sigma^2)$ with $\mu = 1$ or $\mu = 2$, and $\sigma=.1$ or $\sigma = .3$.

\begin{figure}[t]\centering 
\subfigure[Excess Error $F(\bbx^t) - F(\bbx^*)$ vs. iteration $t$.]{
\includegraphics[width=0.45\linewidth,height=0.25\linewidth]
		{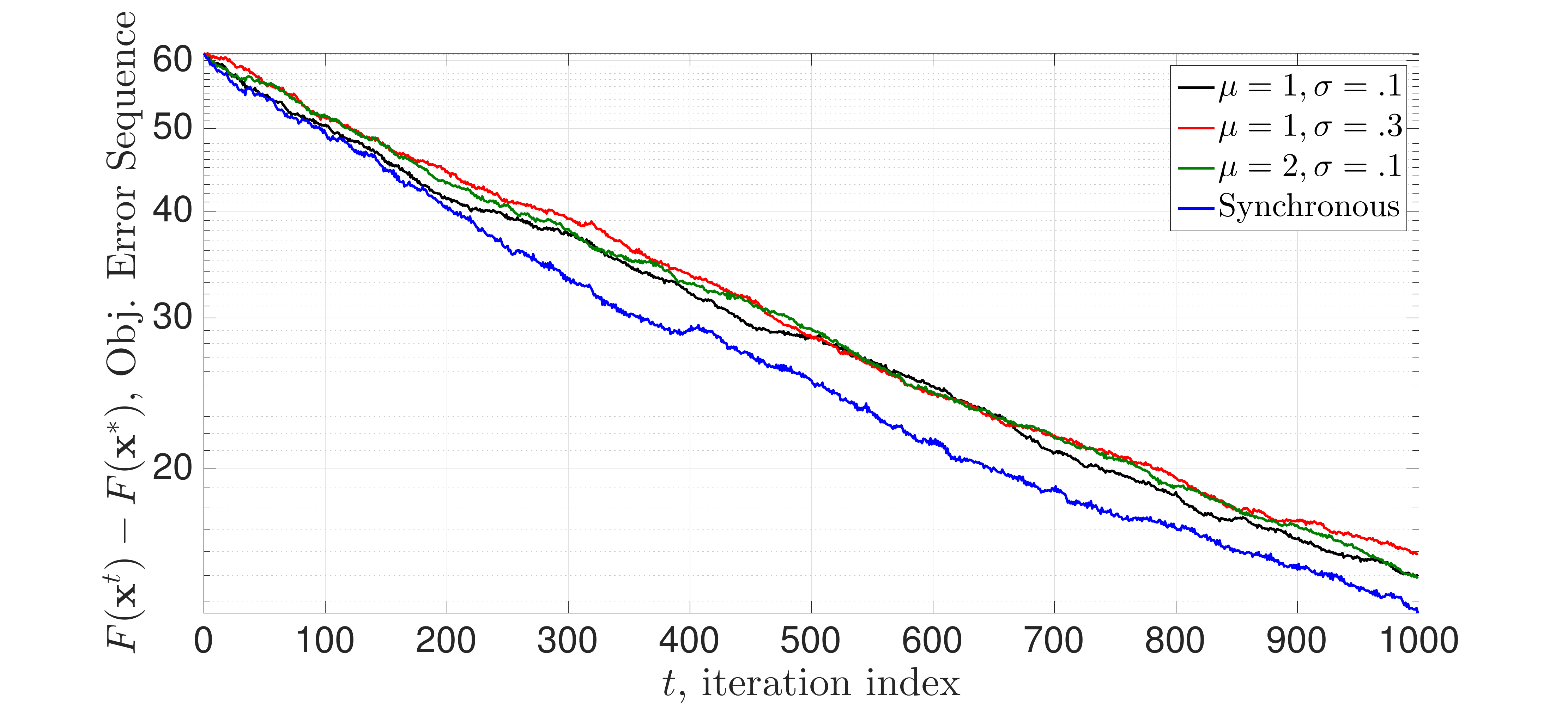}\label{subfig:asyn_rapsa_linear_a}}
\subfigure[Excess Error $F(\bbx^t) - F(\bbx^*)$ vs. iteration $t$.]{
\includegraphics[width=0.45\linewidth,height=0.25\linewidth]
                {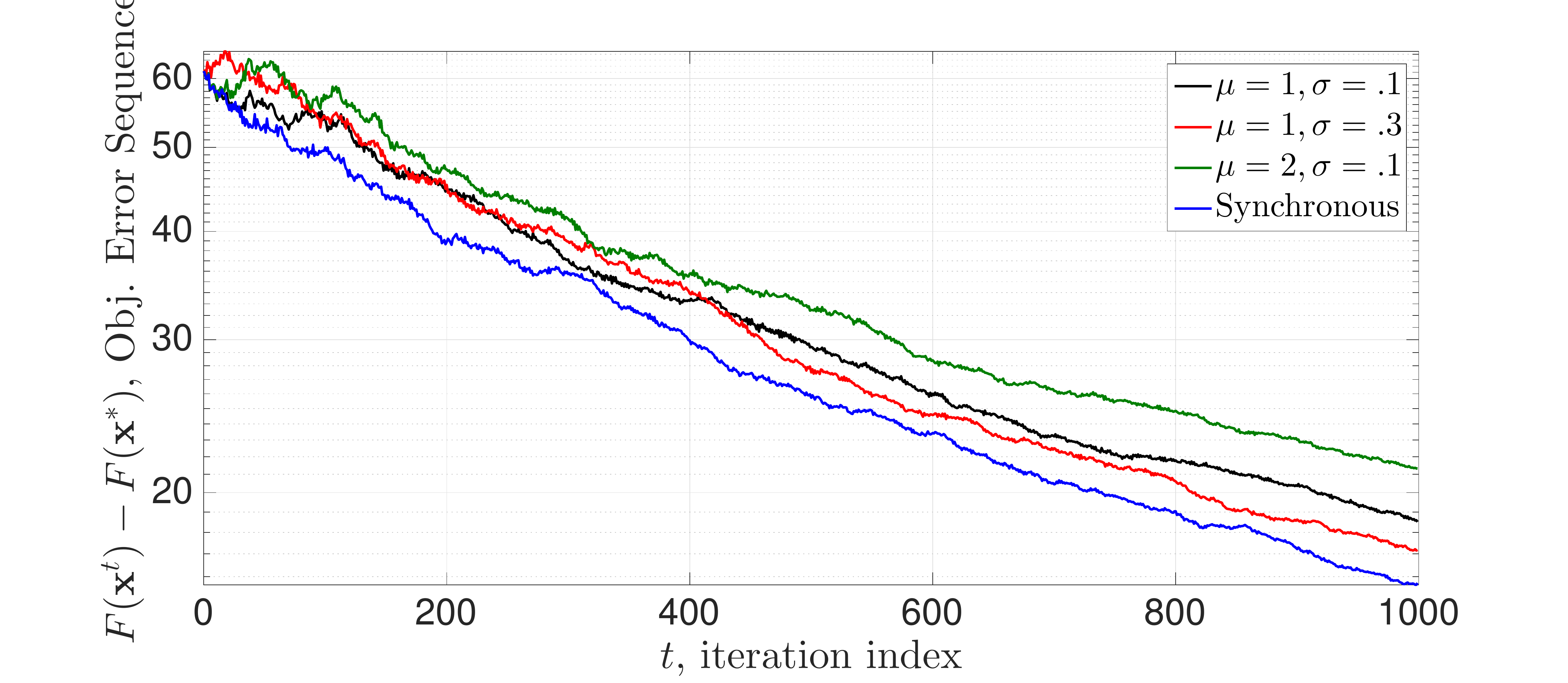}\label{subfig:asyn_rapsa_linear_b}}       
\caption{Asynchronous RAPSA (Algorithm \ref{algo_asyn}) on the linear estimation problem in the constant ($\gamma=10^{4}$, left) and diminishing ($\gamma_t=10^6/(t + 250)$, right) step-size schemes with no mini-batching $L=1$ for a binary training subset of size $N=10^3$ with no regularization $\lambda=0$ when the algorithm is initialized as $\bbx_0 =10^3\times\bbone $. Varying the asynchronicity distribution has little effect, but we find that convergence behavior is slower than its synchronized counterpart, as expected.}\label{fig:asyn_rapsa_linear}
\end{figure}

The results of this numerical experiment are given in Figure \ref{fig:asyn_rapsa_linear} for both the constant and diminishing step-size schemes. We see that the performance of the asynchronous parallel scheme is comparable across different levels of variability among the local clocks of each processor. In particular, in Figure \ref{subfig:asyn_rapsa_linear_a} which corresponds to the case where the algorithm is run with constant step-size $\gamma=10^{-2}$, we observe comparable performance in terms of the objective function error sequence  $F(\bbx^t) - F(\bbx^*)$ with iteration $t$ -- across the varying levels of asynchrony we have  $F(\bbx^t) - F(\bbx^*)\leq 10$ by $t=10^3$. This trend may also be observed in the diminishing step-size scheme $\gamma^t=1/t$ which is given in Figure \ref{subfig:asyn_rapsa_linear_b}. That is, the distance to the optimal objective is nearly identical across differing levels of asynchronicity. In both cases, the synchronized algorithm performs better than its asynchronous counterpart.

%
%
%

%
\subsection{Hand-Written Digit Recognition}\label{subsec:mnist}

We now make use of RAPSA for visual classification of written digits. To do so, let $\bbz\in\reals^p$ be a feature vector encoding pixel intensities (elements of the unit interval $[0,1]$ with smaller values being closer to black) of an image and let $y\in \{-1,1\}$ be an indicator variable of whether the image contains the digit $0$ or $8$, in which case the binary indicator is respectively $y=-1$ or $y=1$.
We model the task of learning a hand-written digit detector as a logistic regression problem, where one aims to train a classifier $\bbx \in \reals^p$ to determine the relationship between feature vectors $\bbz_n \in \reals^p$ and their associated labels $y_n \in \{-1,1\}$  for $n=1,\dots,N$. 
The instantaneous function $f_n$ in \eqref{eq:empirical_min} for this setting is the $\lambda$-regularized negative log-likelihood of a generalized linear model of the odds ratio of whether the label is $y_n=1$ or $y_n=-1$. 
The empirical risk minimization associated with training set $\ccalT=\{(\bbz_n,y_n)\}_{n=1}^N$ is to find $\bbx^*$ as the maximum a posteriori estimate
\begin{equation}\label{logistic_regression}
\bbx^* := \argmin_{\bbx\in\reals^p}  \frac{\lambda}{2}\|\bbx\|^2 + \frac{1}{N}\sum_{n=1}^N \log (1 + \exp({-y_n \bbx^T \bbz_n}))\; ,
\end{equation}
where the regularization term $({\lambda}{/2})\|\bbx\|^2$ encodes a prior belief on the joint distribution of $(\bbz, y)$ and helps to avoid overfitting. We use the MNIST dataset \citep{citeulike:599493}, in which feature vectors $\bbz_n \in \reals^p$ are $p=28^2=784$ pixel images whose values are recorded as intensities, or elements of the unit interval $[0,1]$. Considered here is the subset associated with digits $0$ and $8$, a training set $\ccalT=\{\bbz_n,y_n\}_{n=1}^N$ with $N=1.76\times 10^4$ sample points. 
\begin{figure}[t]\centering 
\subfigure[Objective $F(\bbx^t)$ vs. iteration $t$.]{
\includegraphics[width=0.32\linewidth,height=0.2\linewidth]
		{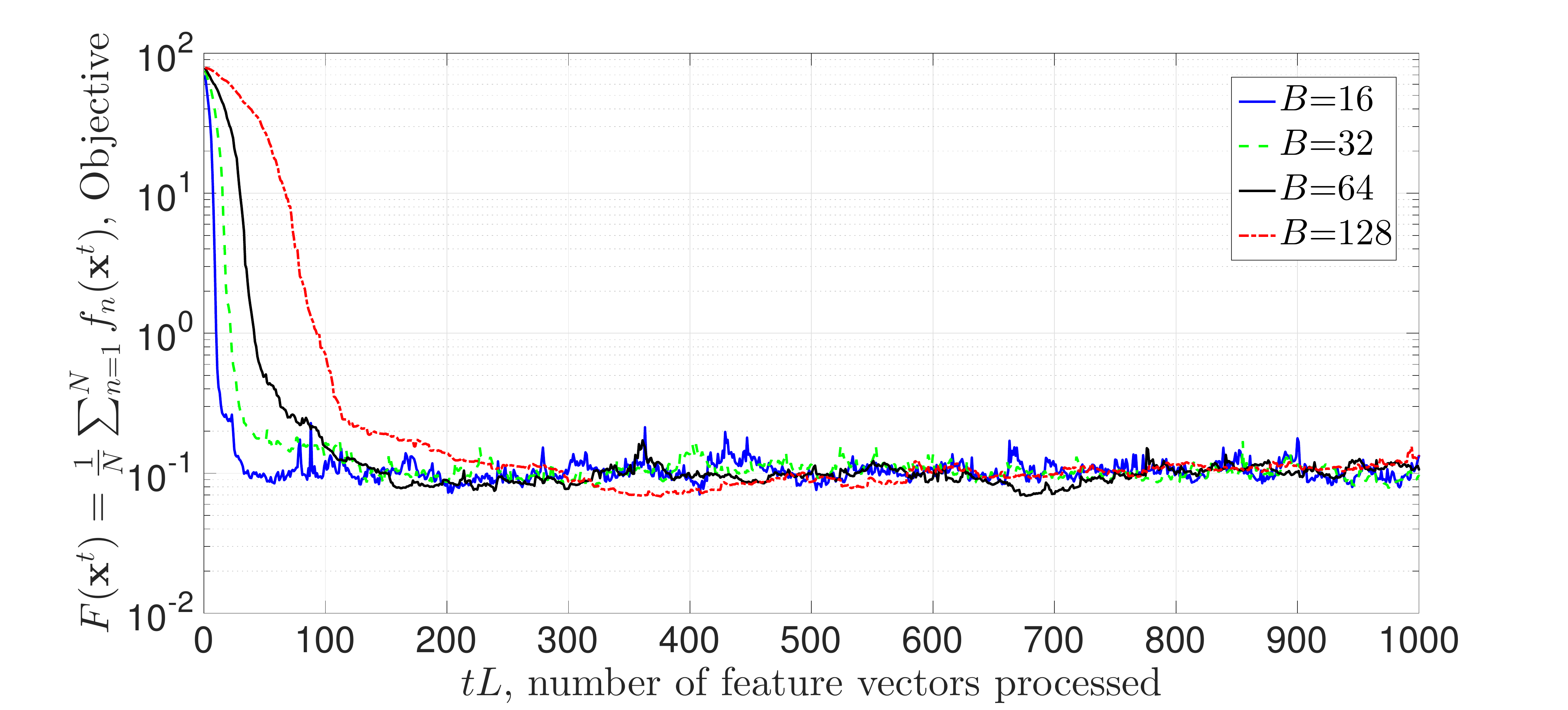}\label{subfig:rapsa_mnist_constant_a}}
\subfigure[Objective $F(\bbx^t)$ vs. feature $\tilde{p}_t$.]{
\includegraphics[width=0.32\linewidth,height=0.2\linewidth]
                {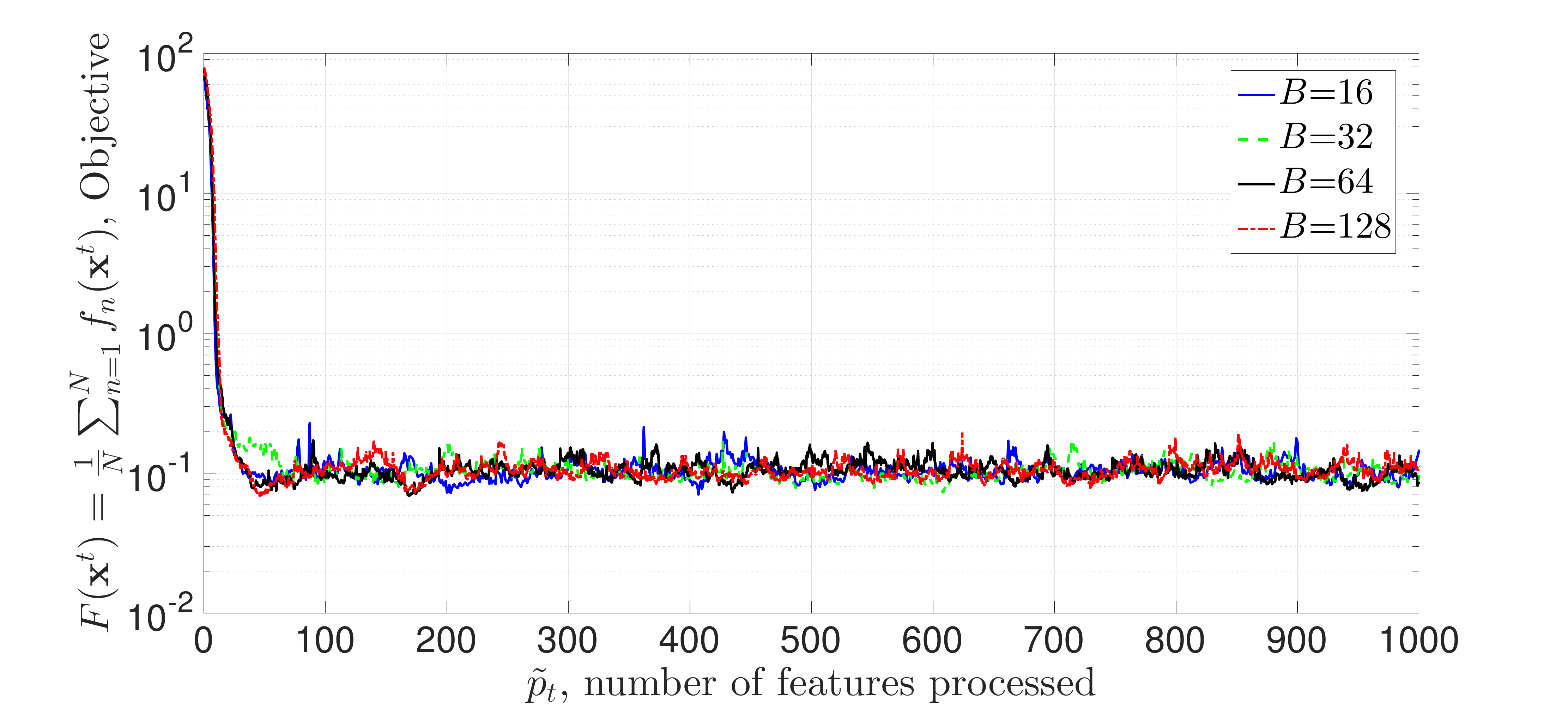}\label{subfig:rapsa_mnist_constant_b}}
\subfigure[Test Set Accuracy vs. feature $\tilde{p}_t$]{
\includegraphics[width=0.32\linewidth, height=0.2\linewidth]
                {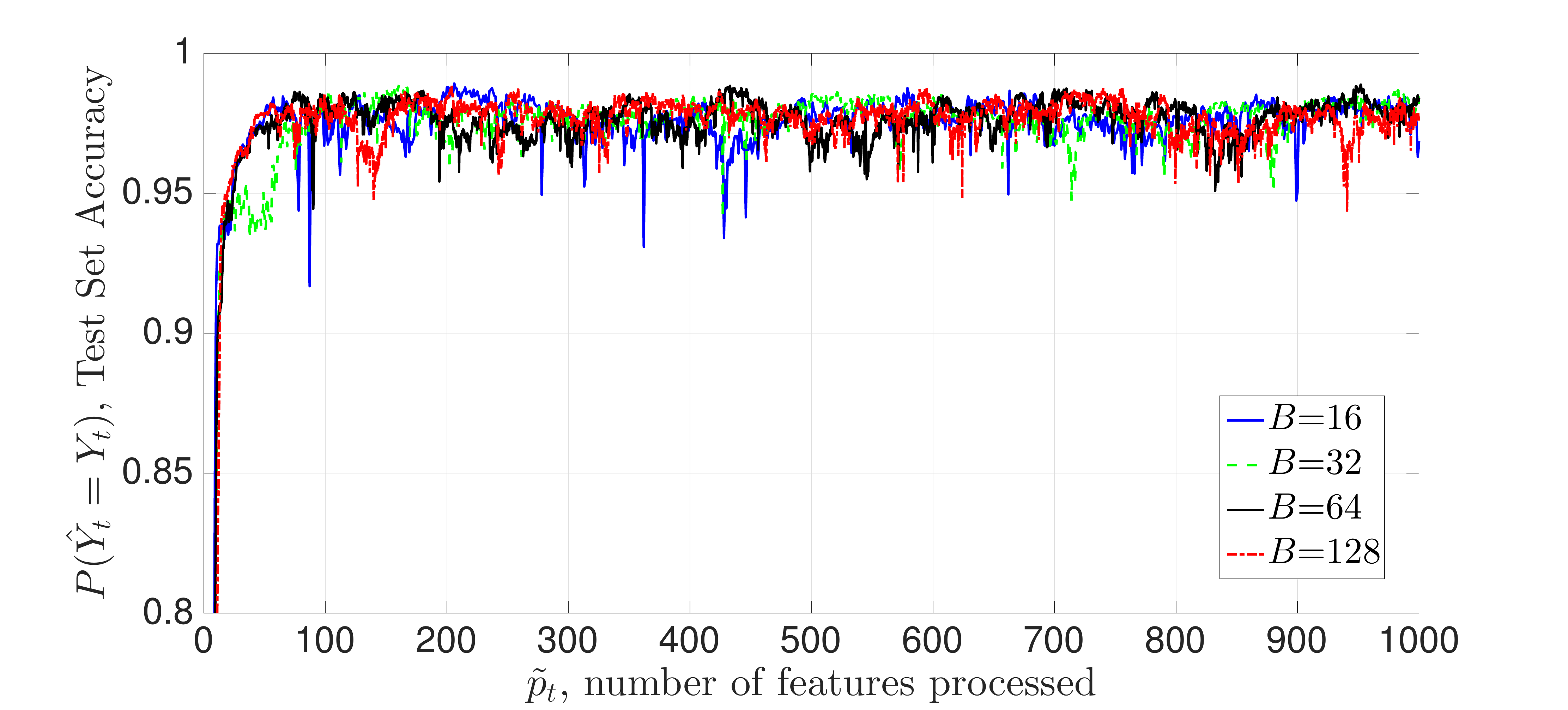}\label{subfig:rapsa_mnist_constant_c}}                
\caption{RAPSA on MNIST data with constant step-size $\gamma^t= \gamma=10^{-.5}$ with no mini-batching $L=1$. Algorithm performance is best in terms of number of iterations $t$ when all blocks are used per step (parallelized SGD), but in terms of number of features processed, the methods perform comparably. Thus RAPSA performs as well as SGD while breaking the complexity bottleneck in $p$, the dimension of decision variable $\bbx$.} \label{fig:rapsa_mnist_constant}
\end{figure}
%

\begin{figure}[t]\centering 
\subfigure[Objective $F(\bbx^t)$ vs. iteration $t$.]{
\includegraphics[width=0.32\linewidth,height=0.2\linewidth]
		{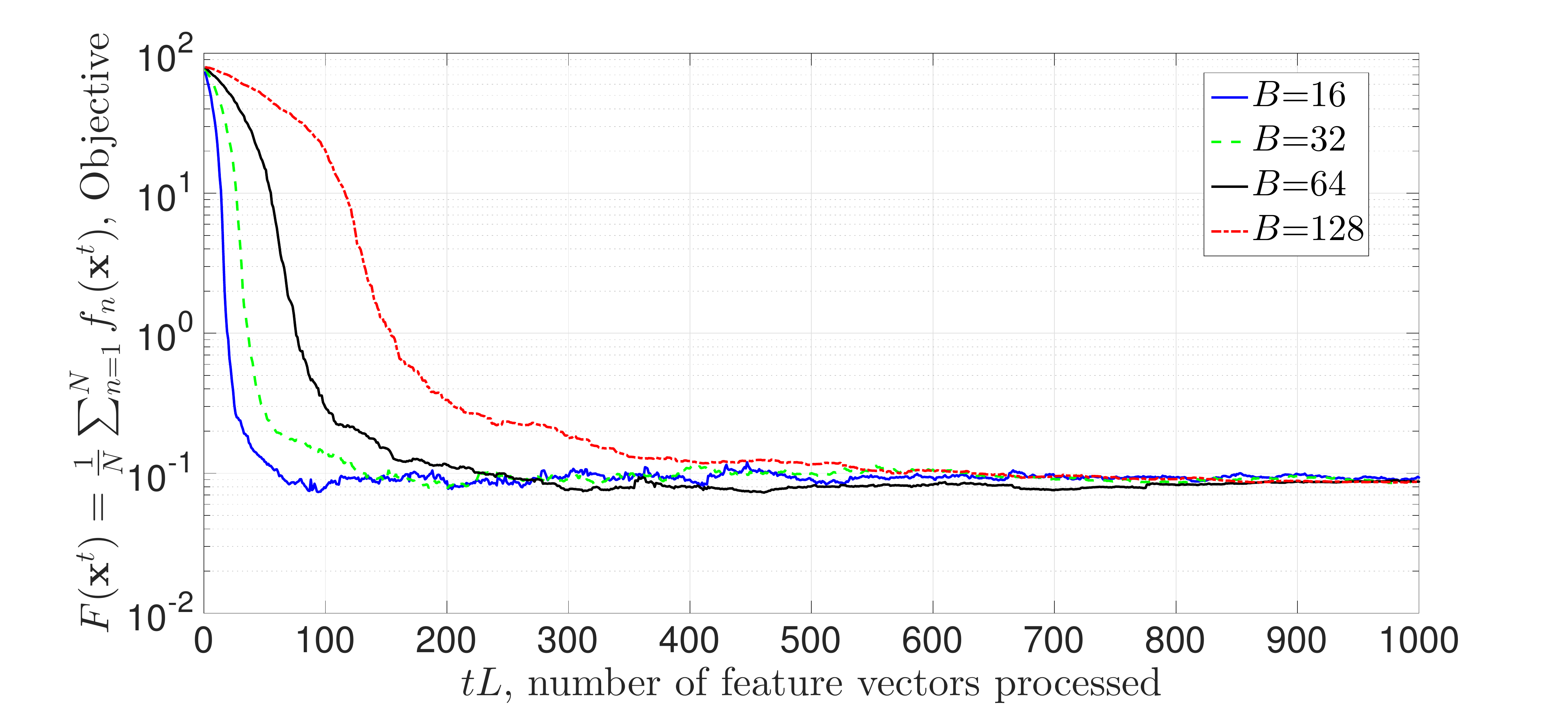}\label{subfig:rapsa_mnist_hybrid_a}}
\subfigure[Objective $F(\bbx^t)$ vs. feature $\tilde{p}_t$.]{
\includegraphics[width=0.32\linewidth,height=0.2\linewidth]
                {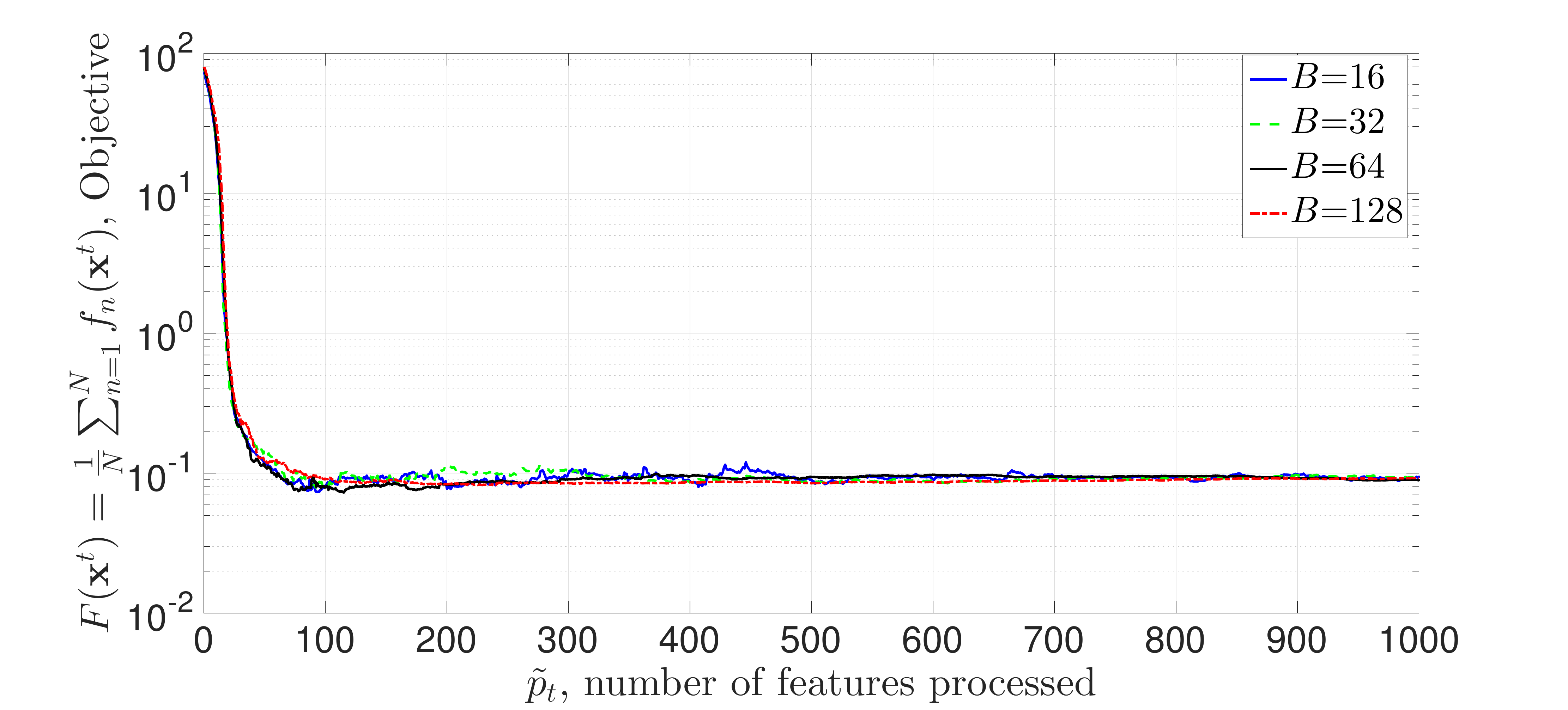}\label{subfig:rapsa_mnist_hybrid_b}}
\subfigure[Test Set Accuracy vs. feature $\tilde{p}_t$]{
\includegraphics[width=0.32\linewidth, height=0.2\linewidth]
                {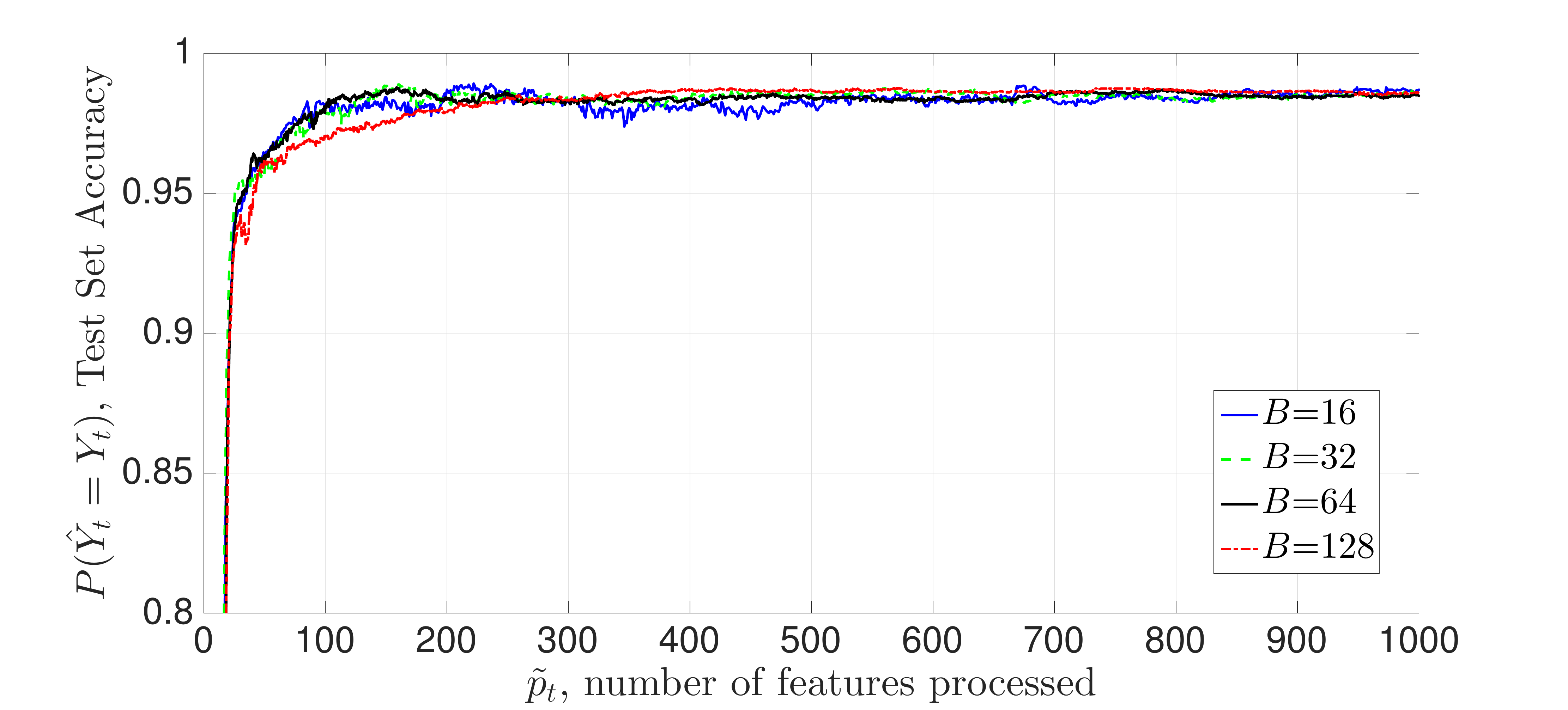}\label{subfig:rapsa_mnist_hybrid_c}}                
\caption{RAPSA on MNIST data with hybrid step-size $\gamma^t= \min(10^{-3/4},10^{-3/4} \tilde{T}_0/t)$, with $\tilde{T}_0=300$ and no mini-batching $L=1$. As with the constant step-size selection, we observe that updating all blocks per iteration is best in terms of $t$, but in terms of elements of $\bbx$ updated, algorithm performance is nearly identical, meaning that no price is payed for breaking the complexity bottleneck in $p$.} \label{fig:rapsa_mnist_hybrid}
\end{figure}
%
{\bf Results for RAPSA} We run RAPSA on this training subset for the cases that $B=16$, $B=32$, $B=64$, and $B=128$, which are associated with updating $p$, $p/2$, $p/4$, and $p/8$ features per iteration. We consider the use of RAPSA with both constant and hybrid step-size selections. In Figure \ref{fig:rapsa_mnist_constant}, we display the results when we select a constant learning rate $\gamma^t=\gamma=10^{-.5}=0.316$. In Figure \ref{subfig:rapsa_mnist_constant_a} we plot the objective $F(\bbx^t)$ versus iteration $t$, and observe that algorithm performance improves with using more elements of $\bbx$ per iteration. That is, using all $p$ coordinates of $\bbx$ achieves superior convergence with respect to iteration $t$. 
However, as previously noted, iteration index $t$ is an unfair comparator for objective convergence since the four different setting process different number of features per iteration. In Figure \ref{subfig:rapsa_mnist_constant_b}, we instead consider $F(\bbx^t)$ versus the number of coordinates of $\bbx$, denoted as $\tilde{p}_t$, that algorithm performance is comparable across the different selections of $B$. This demonstrates that RAPSA breaks the computational bottleneck in $p$ while suffering no reduction in convergence speed with respect to $\tilde{p}_t$.

We consider further the classification accuracy on a test subset of size $\tilde{N}=5.88\times10^3$, the results of which are shown in Fig. \ref{subfig:rapsa_mnist_hybrid_c}. We see that the result for classification accuracy on a test set is consistent with the results for the convergence of the objective function value, and asymptotically reach approximately $98\%$ across the different instances of RAPSA.

In Figure \ref{fig:rapsa_mnist_hybrid} we show the result of running RAPSA for this logistic regression problem with hybrid step-size $\gamma^t= \min(10^{-3/4},10^{-3/4} \tilde{T}_0/t)$, with $\tilde{T}_0=300$ and no mini-batching $L=1$. In Fig. \ref{subfig:rapsa_mnist_hybrid_a}, which displays the objective $F(\bbx^t)$ versus iteration $t$, that using full stochastic gradients is better than only updating \emph{some} of the coordinates in terms of the number of iterations $t$. In particular, to reach the objective benchmark $F(\bbx^t) \leq 10^{-1}$, we have to run RAPSA $t=74$, $t=156$, and $t=217$, and $t=631$ iterations, for the cases that $B=16$, $B=32$, $B=64$, and $B=128$.
We illustrate  the objective $F(\bbx^t)$ vs. feature $\tilde{p}_t$ in Fig. \ref{subfig:rapsa_mnist_hybrid_b}. Here we recover the advantages of randomized incomplete parallel processing: updating fewer blocks per iteration yields comparable algorithm performance. 


We additionally display the algorithm's achieved test-set accuracy on a test subset of size $\tilde{N}=5.88\times10^3$ in Fig. \ref{subfig:rapsa_mnist_hybrid_c} under the hybrid step-size regime. We again see that after a burn-in period, the classifier achieves the highly accurate asymptotic error rate of between $1-2\%$ across the different instantiations of RAPSA. We note that the test set accuracy achieved by the hybrid scheme is superior to the constant step-size setting.

%
%

{\bf Results for Accelerated RAPSA} We now run Accelerated RAPSA (Algorithm \ref{algo_ARAPS}) as stated in Section \ref{sec:arapsa} for this problem setting for the entire MNIST binary training subset associated with digits $0$ and $8$, with mini-batch size $L=10$ and the level of curvature information set as  $\tau=10$. We further select regularizer $\lambda=1/\sqrt{N}=7.5\times 10^{-3}$, and consider both constant and hybrid step-size regimes. As before, we study the advantages of incomplete randomized parallel processing by varying the number of blocks $B\in \{16, 32, 64,128\}$ on an architecture with a fixed number $|\ccalI_t|=I=16$ of processors. This setup is associated with using all $p$ entries of vector $\bbx$ at each iteration as compared with $1/2$, $1/4$, and $1/8$ of its entries.

\begin{figure}[t]\centering 
\subfigure[Objective $F(\bbx^t)$ vs. iteration $t$.]{
\includegraphics[width=0.32\linewidth,height=0.2\linewidth]
		{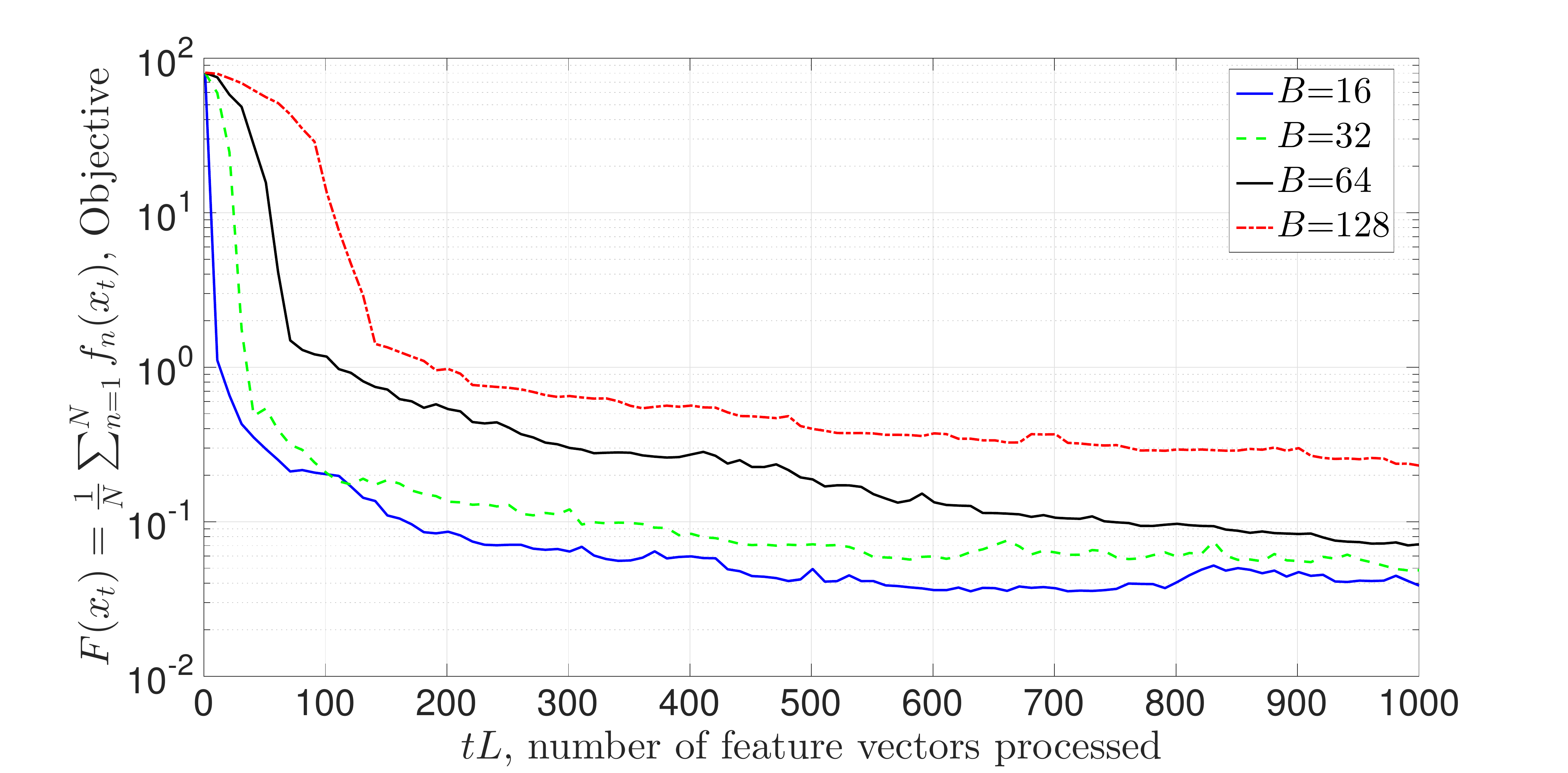}\label{subfig:arapsa_mnist_constant_a}}
\subfigure[Objective $F(\bbx^t)$ vs. feature $\tilde{p}_t$.]{
\includegraphics[width=0.32\linewidth,height=0.2\linewidth]
                {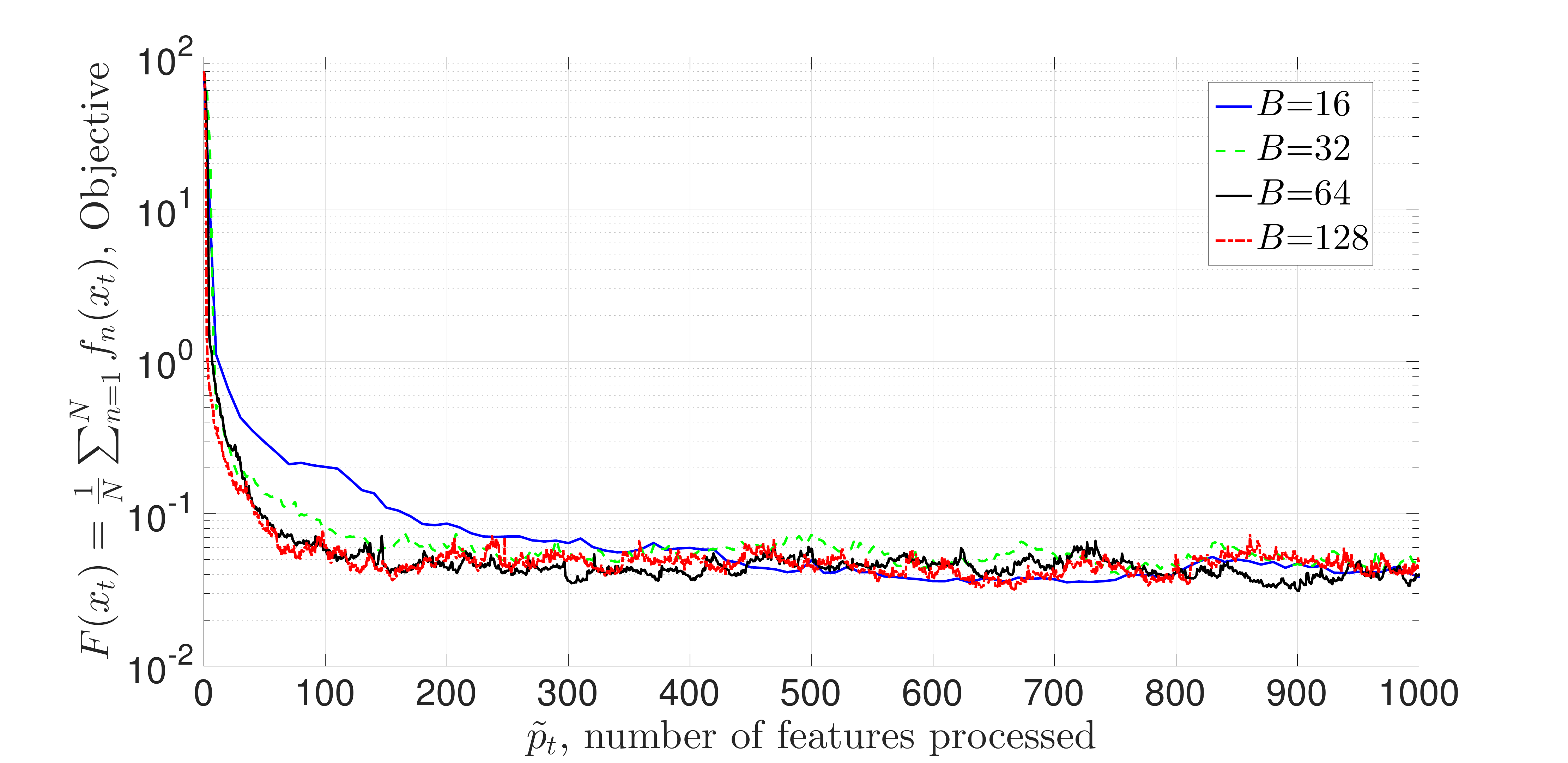}\label{subfig:arapsa_mnist_constant_b}}
\subfigure[Test Set Accuracy vs. feature $\tilde{p}_t$]{
\includegraphics[width=0.32\linewidth, height=0.2\linewidth]
                {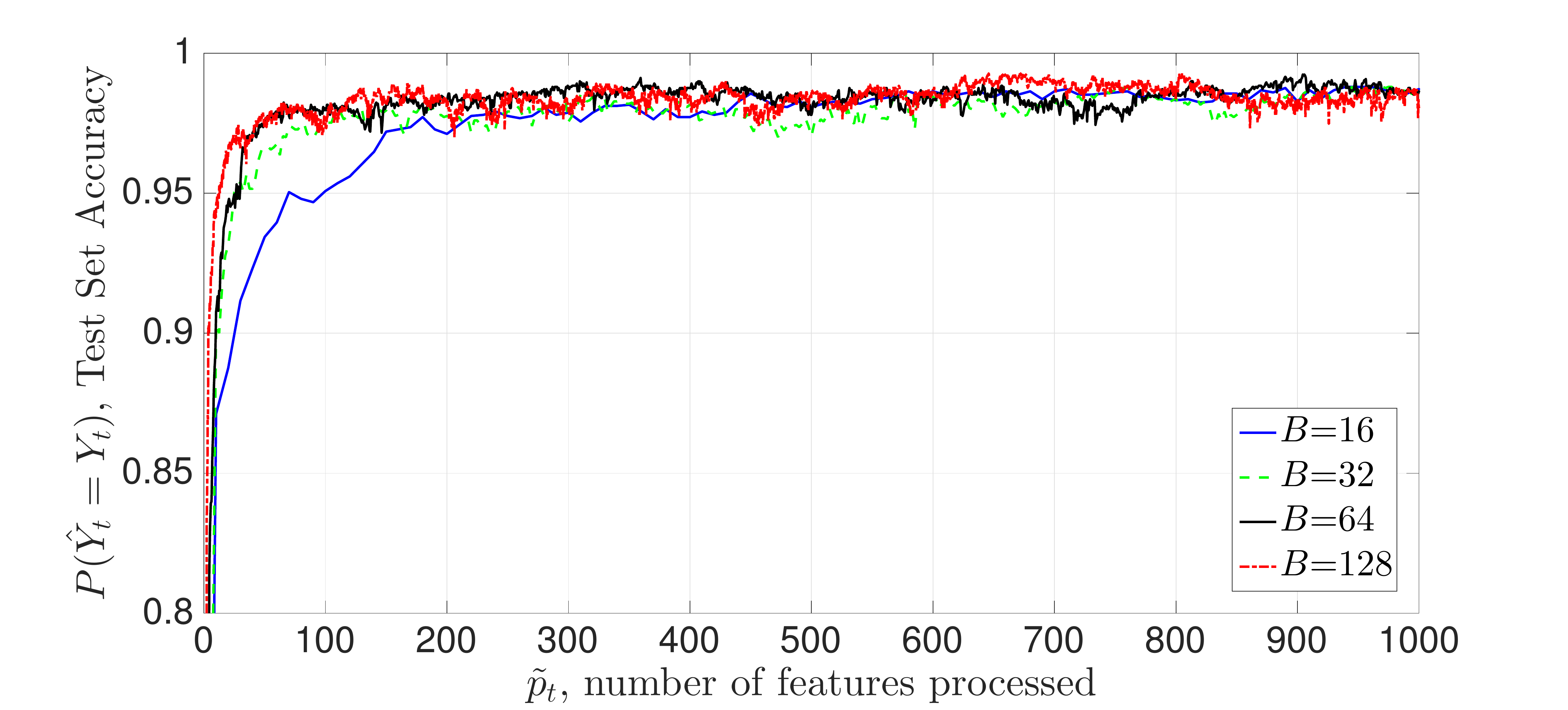}\label{subfig:arapsa_mnist_constant_c}}                
\caption{ARAPSA on MNIST data with constant step-size $\gamma^t= \gamma=10^{-2}$ and mini-batch size $L=10$, curvature memory $\tau=10$, and regularizer $\lambda=7.5\times10^{-3}$. Algorithm performance is comparable across different numbers of decision variable coordinates updated per iteration $t$, but in terms of number of features processed, ARAPSA performance best when using the least information per update.}\label{fig:arapsa_mnist_constant}
\end{figure}
%

\begin{figure}[t]\centering 
\subfigure[Objective $F(\bbx^t)$ vs. iteration $t$.]{
\includegraphics[width=0.32\linewidth,height=0.2\linewidth]
		{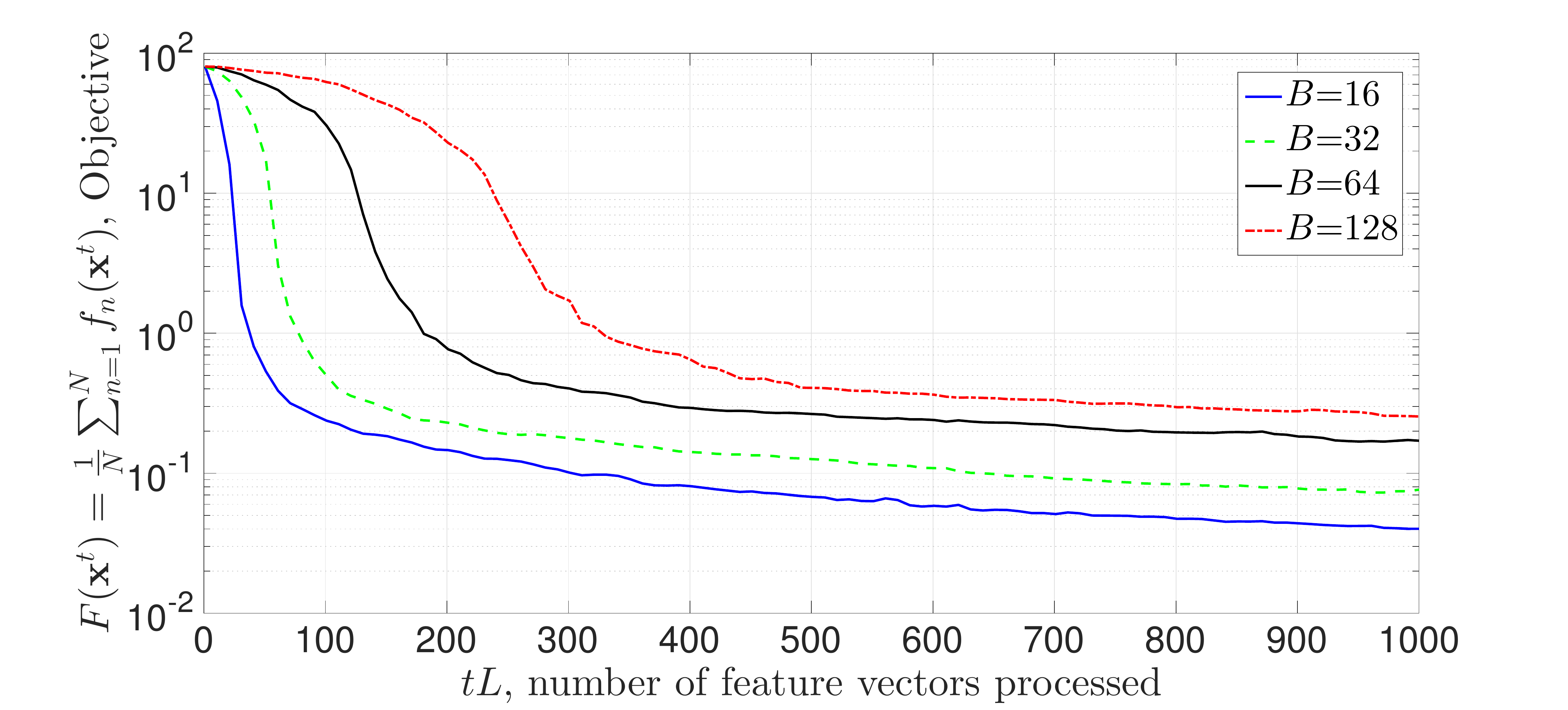}\label{subfig:arapsa_mnist_hybrid_a}}
\subfigure[Objective $F(\bbx^t)$ vs. feature $\tilde{p}_t$.]{
\includegraphics[width=0.32\linewidth,height=0.2\linewidth]
                {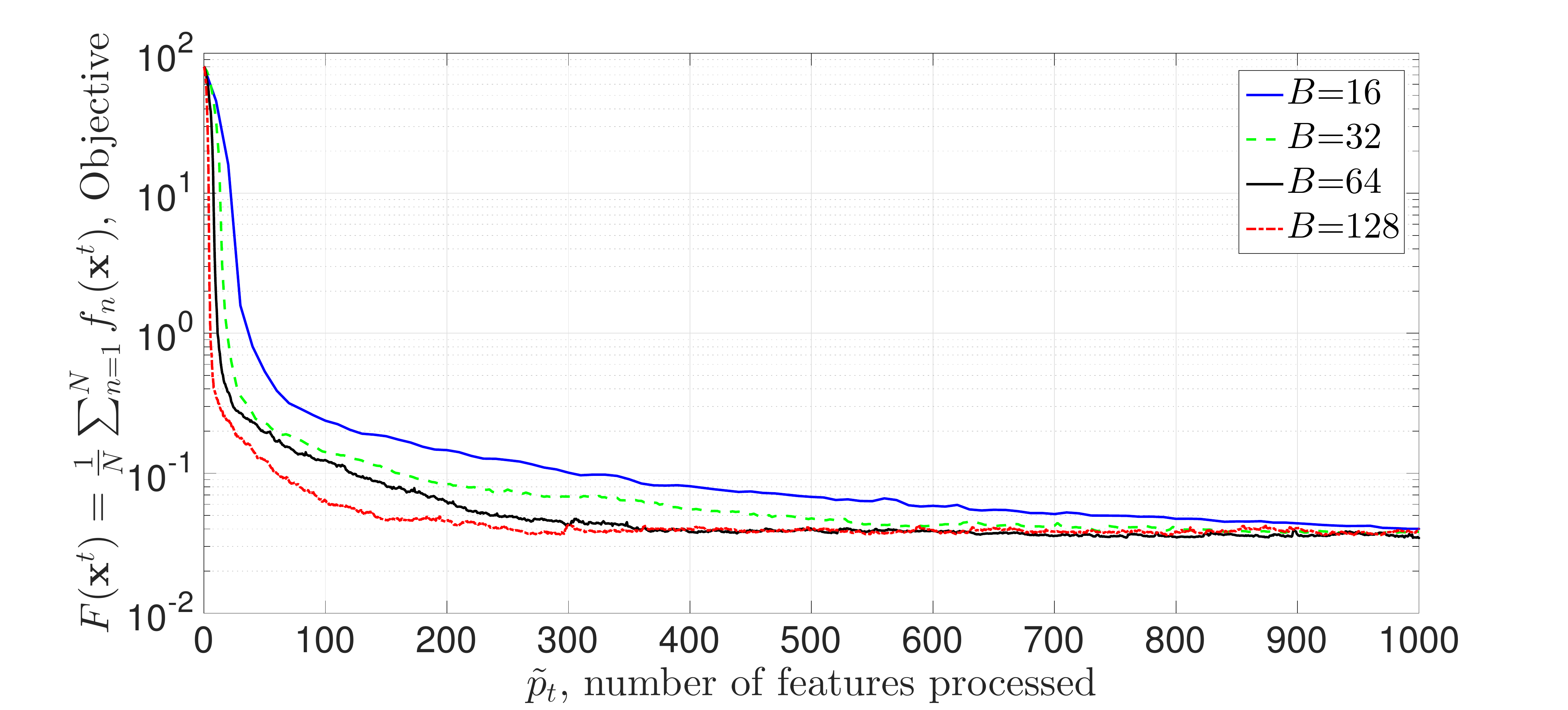}\label{subfig:arapsa_mnist_hybrid_b}}
\subfigure[Test Set Accuracy vs. feature $\tilde{p}_t$]{
\includegraphics[width=0.32\linewidth, height=0.2\linewidth]
                {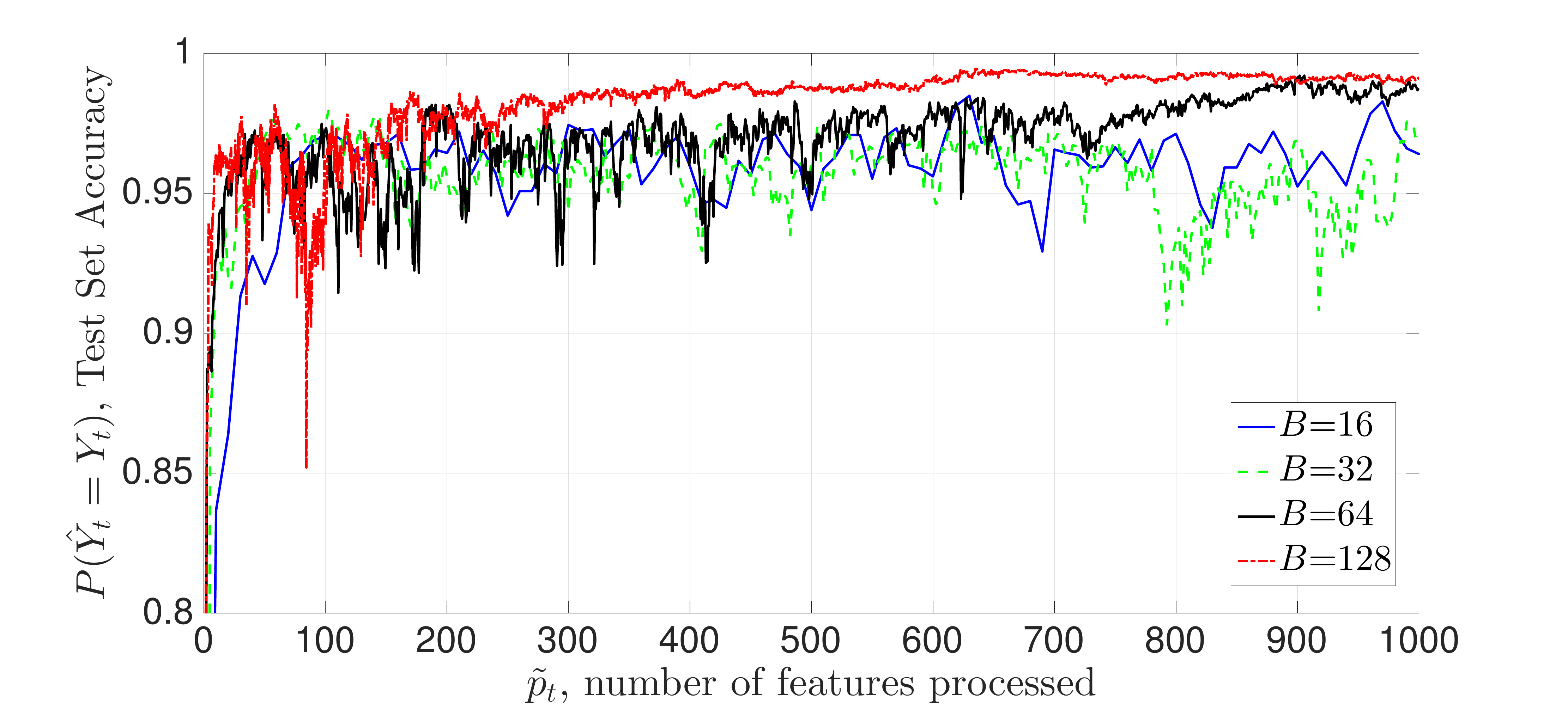}\label{subfig:arapsa_mnist_hybrid_c}}                
\caption{ARAPSA on MNIST data with hybrid step-size $\gamma^t= \min(10^{-1},10^{-1} \tilde{T}_0/t)$, with $\tilde{T}_0=500$, mini-batch size $L=10$, curvature memory $\tau=10$, and regularizer $\lambda=7.5\times10^{-3}$. Algorithm performance is comparable across different numbers of decision variable coordinates updated per iteration $t$, but in terms of number of features processed, RAPSA performance best when using the least information per update.}\label{fig:arapsa_mnist_hybrid}
\end{figure}

Figures \ref{fig:arapsa_mnist_constant} the results of this algorithm run when a constant step-size $\gamma=10^{-2}$ is used. Observe in Figure \ref{subfig:arapsa_mnist_constant_a} that the algorithm achieves convergence across the differing numbers of blocks $B$ in terms of iteration $t$, with faster learning rates achieved with smaller $B$. In particular, to reach the benchmark $F(\bbx^t) \leq 10^{-1}$, we require $t=145$, $t=311$, and $t=701$ iterations for $B=16$, $B=32$, and $B=64$, respectively, whereas the case $B=128$ does not achieve this benchmark by $t=10^3$. This trend is inverted, however, in  Figure \ref{subfig:arapsa_mnist_constant_b}, which displays the objective $F(\bbx^t)$ with $\tilde{p}_t$ the number of coordinates of $\bbx$ on which the algorithm operates per step. Observe that using \emph{fewer} entries of $\bbx$ per iteration is better in terms of number of features processed $\tilde{p}_t$. Furthermore, ARAPSA achieves comparable accuracy on a test set of images, approximately near $98\%$ across different selections of $B$, as is displayed in Figure \ref{subfig:arapsa_mnist_constant_c}.

We now run Accelerated RAPSA when the learning rate is hand-tuned to optimize performance via a hybrid scheme $\gamma^t= \min(10^{-1},10^{-1} \tilde{T}_0/t)$, with attenuation threshold $\tilde{T}_0=500$. The results of this experiment are given in Figure \ref{fig:arapsa_mnist_hybrid}. In particular, in Figure \ref{subfig:arapsa_mnist_hybrid_a} we plot the objective $F(\bbx^t)$ with iteration $t$ when the number of blocks $B$ is varied. We see that parallelized oL-BFGS ($I=B$ so that $r=1$) performs best in terms of $t$: to achieve the threshold condition $F(\bbx^t) \leq 10^{-1}$, we require $t=278$, $t=522$ iterations for $B=16$ and $B=32$, respectively, whereas the cases $B=64$ and $B=128$ do not achieve this benchmark by $t=10^3$. However, the instance of ARAPSA with the fastest and most accurate convergence uses the \emph{least} coordinates of $\bbx$ when we compare the objective with $\tilde{p}_t$, as may be observed in Figure \ref{subfig:arapsa_mnist_hybrid_b}.  This trend is corroborated in Figure \ref{subfig:arapsa_mnist_hybrid_c}, where we observe that ARAPSA with $B=128$ achieves $99\%$ test-set accuracy the fastest, followed by $B=64$, $B=32$, and $B=16$.
 
{\bf Comparison of RAPSA and ARAPSA} 
We now compare the performance of RAPSA and its accelerated variant on the MNIST digit recognition problem for a binary subset of the training data consisting of $N=10^5$ samples. We run both algorithms on an $I=16$ processor simulated architecture with $B=64$ blocks, such that $r=1/4$ of the elements of $\bbx$ are operated upon at each step. We consider the constant algorithm step-size scheme $\gamma=10^{-2}$ with mini-batch size $L=10$. 

\begin{figure}\centering
\subfigure[$F(\bbx^t) - F(\bbx^*)$ vs. iteration $t$]{
\includegraphics[width=0.48\linewidth,,height=0.28\linewidth]
{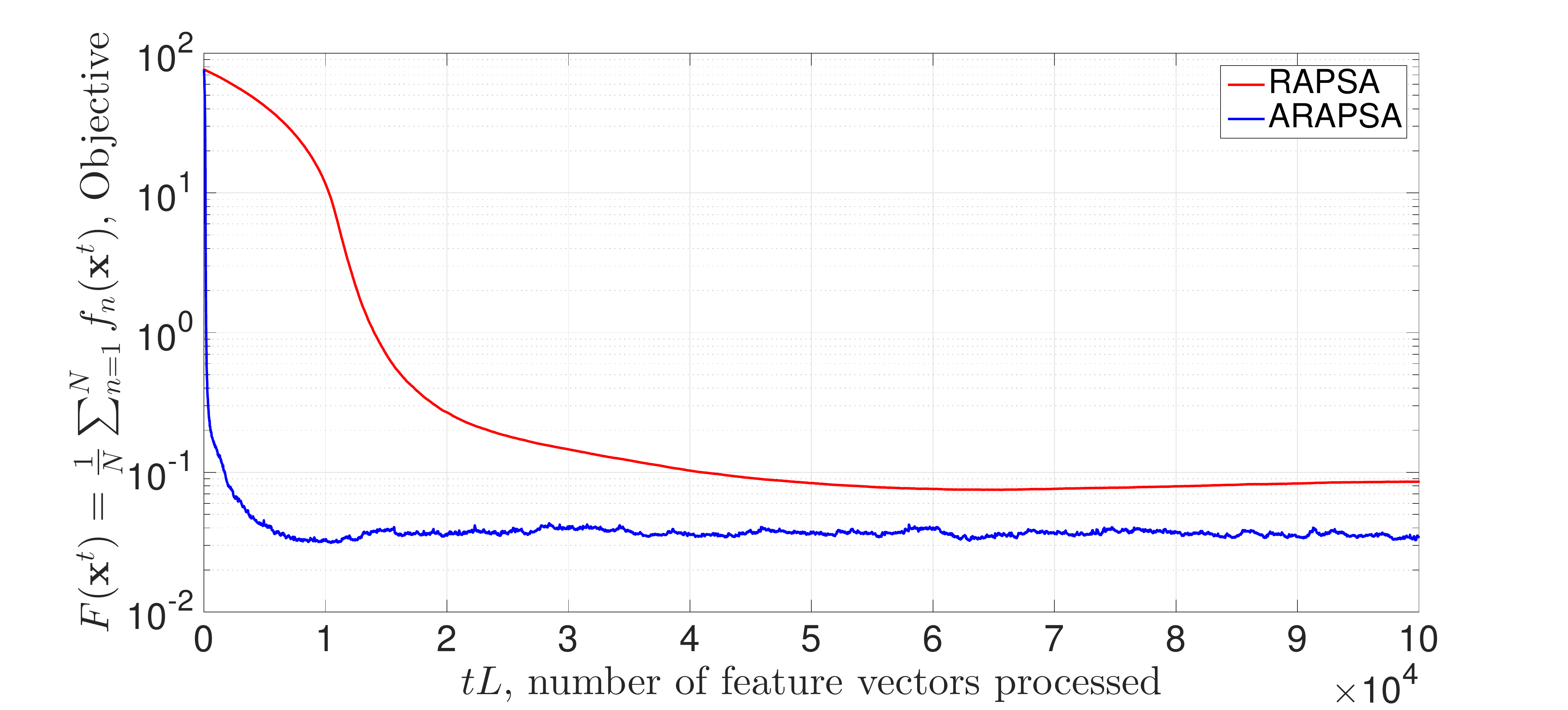}\label{subfig:rapsa_vs_arapsa_logistic_a}}
\subfigure[$F(\bbx^t) - F(\bbx^*)$ vs. clock time (s)]{
\includegraphics[width=0.48\linewidth,,height=0.28\linewidth]
{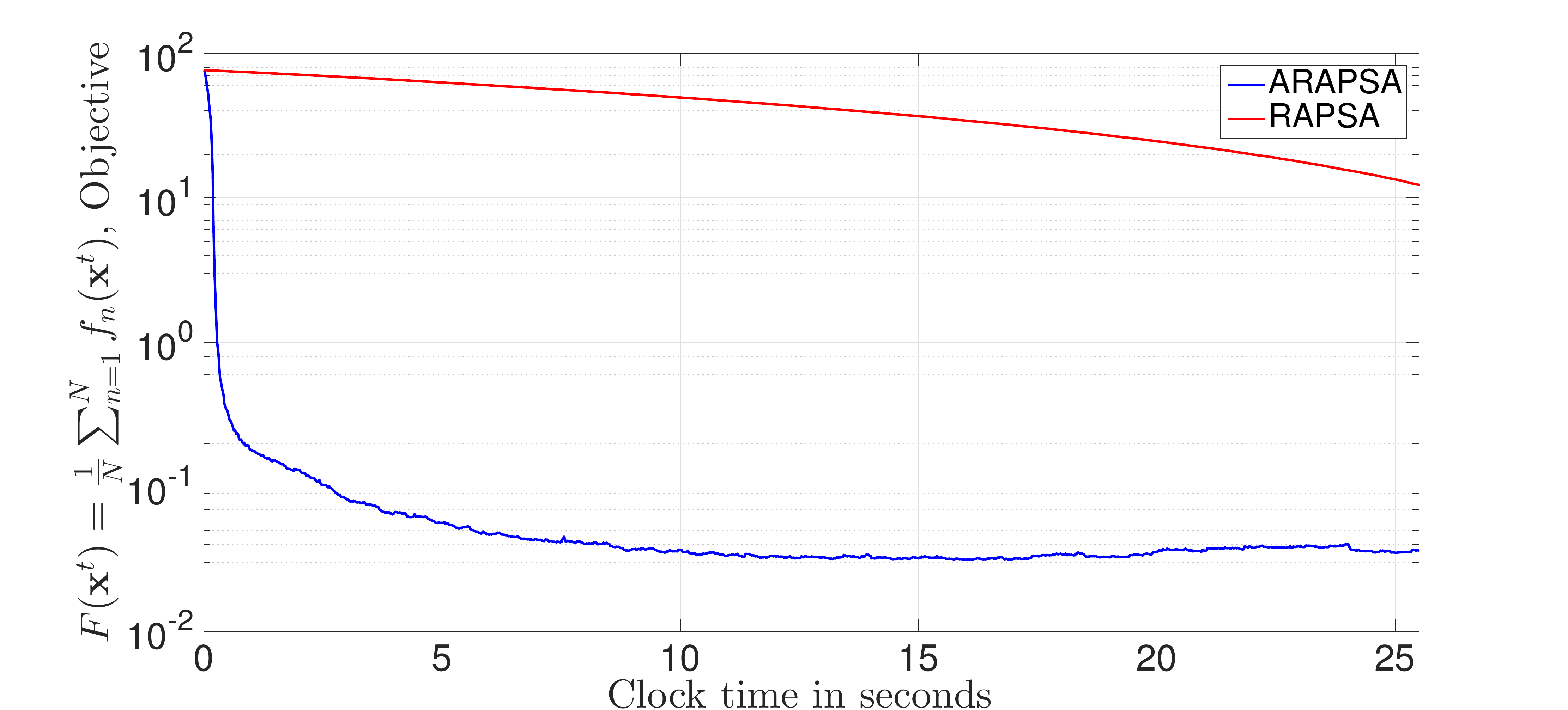}\label{subfig:rapsa_vs_arapsa_logistic_b}}
\caption{A comparison of RAPSA and ARAPSA on the MNIST digit recognition problem for a binary training subset of size $N=10^3$ with mini-batch size $L=10$ in the constant step-size scheme $\gamma=10^{-2}$. The objective optimality gap $F(\bbx^t) - F(\bbx^*)$ is shown with respect to the number of feature vectors processed $tL$ (left) and actual elapsed time (right). While the performance difference between RAPSA and ARAPSA is not as large as in the linear estimation problem, we still observe that ARAPSA substantially accelerates the convergence of RAPSA for a standard machine learning problem.} \label{fig:rapsa_vs_arapsa_logistic}
\end{figure}

The results of this online training procedure are given in \eqref{fig:rapsa_vs_arapsa_logistic}, where we plot the objective optimality gap $F(\bbx^t) - F(\bbx^*)$ versus the number of feature vectors processed $tL$ (Figure \ref{subfig:rapsa_vs_arapsa_logistic_a}) and actual elapsed time (Figure \ref{subfig:rapsa_vs_arapsa_logistic_b}). We see ARAPSA achieves superior convergence behavior with respect to RAPSA in terms of number of feature vectors processed: to achieve the benchmark $F(\bbx^t) - F(\bbx^*)\leq 10^{-1}$, ARAPSA requires fewer than $tL=200$ feature vectors, whereas RAPSA requires $tL=4\times10^4 $ feature vectors. This relationship is corroborated in Figure \ref{subfig:rapsa_vs_arapsa_logistic_b}, where we see that within a couple seconds ARAPSA converges to within $10^{-1}$, whereas after $\emph{five}$ times as long, RAPSA does not achieve this benchmark.


{\bf Results for Asynchronous RAPSA} 
We now evaluate the empirical performance of the asynchronous variant of RAPSA (Algorithm \ref{algo_asyn}) proposed in Section \ref{subsec:asynchronous} on the logistic regression formulation of the MNIST digit recognition problem. The model we use for asynchronicity is the one outlined in Section \ref{subsec:lmmse}, that is, each local processor has a distinct local clock which is not required coincide with others, begins at time $t_0^i=t_0$ for all processors $i=1,\dots,I$, and then selects subsequent times as $t_{k} = t_{k-1} + w_k^i$. Here $w_k^i \sim \ccalN(\mu,\sigma^2)$ is a normal random variable with mean $\mu$ and variance $\sigma^2$ which controls the amount of variability between the clocks of distinct processors. We run the algorithm with no regularization $\lambda=0$ or mini-batching $L=1$ and initialization  $\bbx_0 =\bbone $.

\begin{figure}[t]\centering 
\subfigure[Objective $F(\bbx^t)$ vs. iteration $t$.]{
\includegraphics[width=0.45\linewidth,height=0.25\linewidth]
		{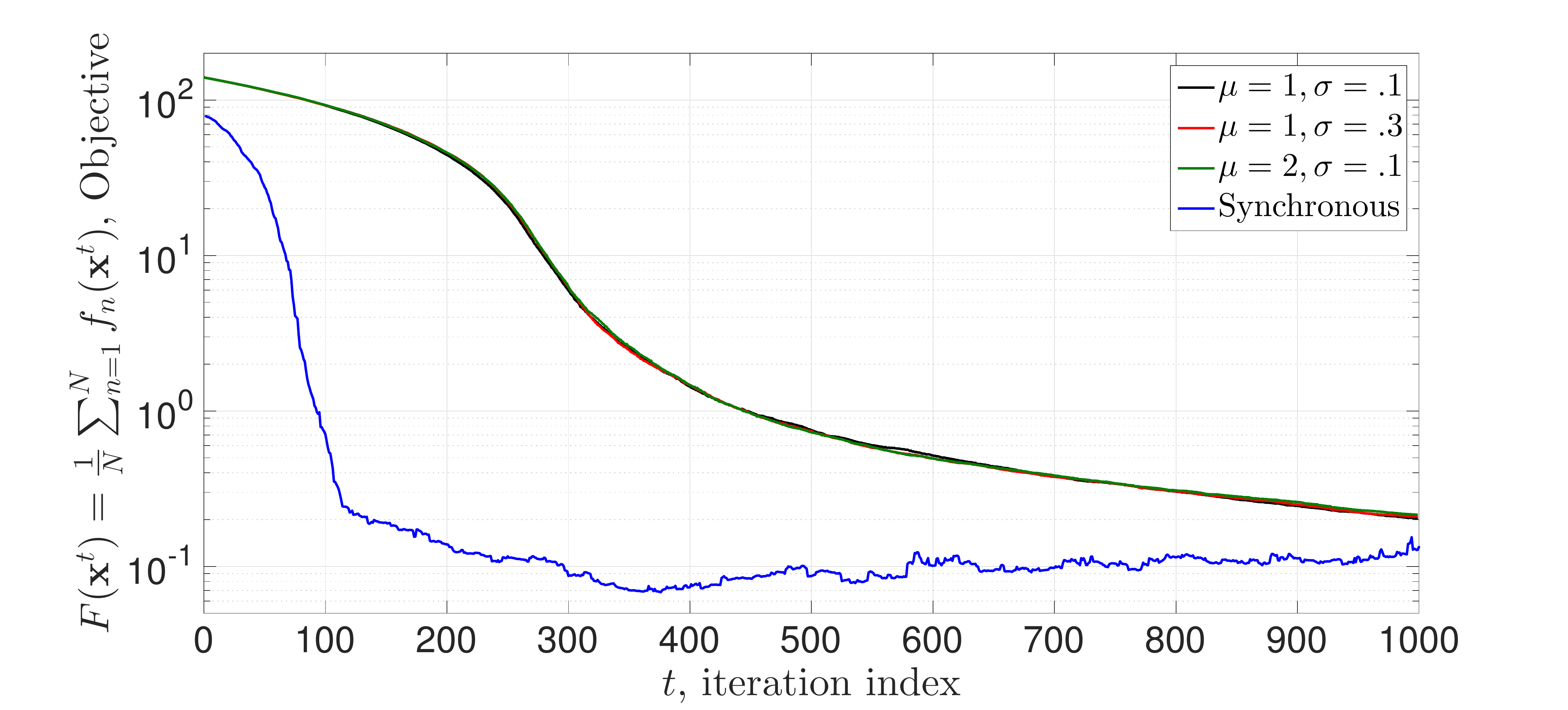}\label{subfig:asyn_rapsa_mnist_a}}
\subfigure[Objective $F(\bbx^t)$ vs. iteration $t$.]{
\includegraphics[width=0.45\linewidth,height=0.25\linewidth]
                {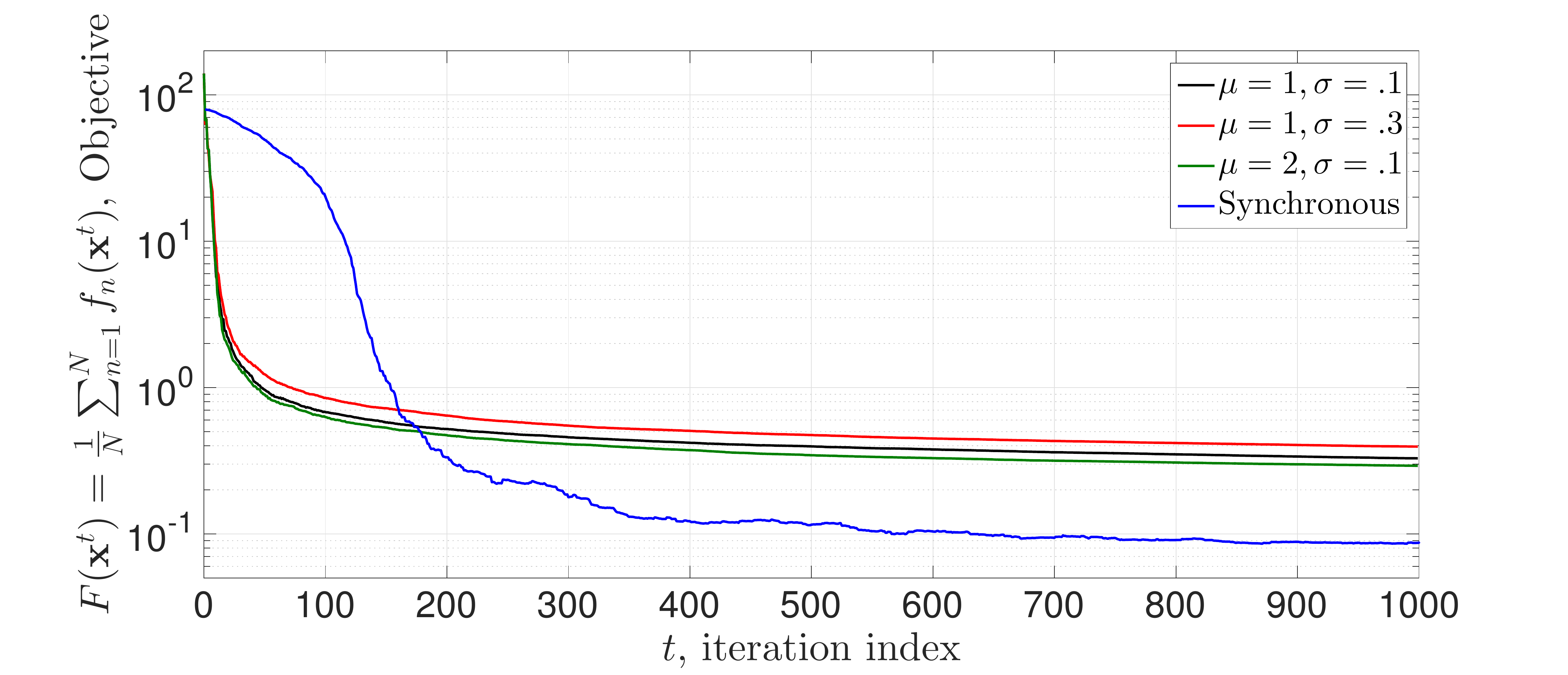}\label{subfig:asyn_rapsa_mnist_b}}       
\caption{Asynchronous RAPSA on MNIST data in the constant ($\gamma=10^{-2}$, left) and diminishing ($\gamma^t=1/t$, right) step-size schemes with no mini-batching $L=1$ for a binary training subset of size $N=10^3$ with no regularization $\lambda=0$ when the algorithm is initialized as $\bbx_0 =\bbone $. The variability in local processor clocks does not significantly impact performance in both the diminishing and constant step-size settings; however, the synchronous algorithm converges at a faster rate.}\label{fig:asyn_rapsa_mnist}
\end{figure}

The results of this numerical setup are given in Figure \ref{fig:asyn_rapsa_mnist}. We consider the expected risk $F(\bbx^t)$ in both both the constant ($\gamma=10^{-2}$, Figure \ref{subfig:asyn_rapsa_mnist_a}) and diminishing ($\gamma^t=1/t$, Figure \ref{subfig:asyn_rapsa_mnist_b}) algorithm step-size schemes. We see that the level of asynchronicity does not significantly impact the performance in either scheme, and that the convergence guarantees established in Theorem \ref{RAPSA_convg_thm_asyn} hold true in practice. We again observe that the version of RAPSA with synchronized computations converges at a faster rate than Asynchronous RAPSA.

%
 

\section{Conclusions}\label{sec_conclusions}

We proposed the random parallel stochastic algorithm (RAPSA) proposed as a doubly stochastic approximation algorithm capable of optimization problems associated with learning problems in which both the number of predictive parameters and sample size are huge-scale. RAPSA is doubly stochastic since each processors utilizes a random set of functions to compute the stochastic gradient associated with a randomly chosen sets of variable coordinates. We showed the proposed algorithm converges to the optimal solution sublinearly when the step-size is diminishing. Moreover, linear convergence to a neighborhood of the optimal solution can be achieved using a constant step-size. We further introduced accelerated and asynchronous variants of RAPSA, and presented convergence guarantees for asynchronous RAPSA.

A detailed numerical comparison between RAPSA and parallel SGD for learning a linear estimator and a logistic regressor is provided. The numerical results showcase the advantage of RAPSA with respect to parallel SGD. Further empirical results illustrate the advantages of ARAPSA with respect to parallel oL-BFGS, and that implementing the algorithm on a lock-free parallel computing cluster does not substantially degrade empirical performance.

{

\appendix

\section{Proof of Results Leading to Theorems \ref{RAPSA_convg_thm} and \ref{RAPSA_convg_thm_finite}}\label{apx_syn}
\subsection{Proof of Lemma \ref{exp_wrt_blocks}}\label{apx_lemma_exp_wrt_blocks}

Recall that the components of vector $\bbx^{t+1}$ are equal to the components of $\bbx^t$ for the coordinates that are not updated at step $t$, i.e., $i\notin \ccalI^t$. For the updated coordinates $i\in \ccalI^t$ we know that $\bbx^{t+1}_{i}=\bbx^{t}_{i}-\gamma^t  \nabla_{\bbx_i^t} f( \bbx^{t}, \bbtheta^t)$. Therefore, $B-I$ blocks of the vector $\bbx^{t+1}-\bbx^t$ are 0 and the remaining $I$ randomly chosen blocks are given by $-\gamma^t  \nabla_{\bbx_i^t} f( \bbx^{t}, \bbtheta^t)$. Notice that there are ${B}\choose{I}$ different ways for picking $I$ blocks out of the whole $B$ blocks. Therefore, the probability of each combination of blocks is $1/ {{B}\choose{I}}$. Further, each block appears in ${B-1}\choose{I-1}$ of the combinations. Therefore, the expected value can be written as 
\begin{equation}\label{lemma_RAPS_dec_20}
\mathbb{E}_{\ccalI^t}\!\left[   \bbx^{t+1}-\bbx^t \mid \ccalF^t   \right]
	=\frac{{{B-1}\choose{I-1}}}{{{m}\choose{I}}} \left(  -\gamma^t  \nabla f( \bbx^{t}, \bbTheta^t)  \right).
\end{equation}
Observe that simplifying the ratio in the right hand sides of \eqref{lemma_RAPS_dec_20} leads to 
\begin{equation}\label{lemma_RAPS_dec_30}
\frac{{{B-1}\choose{I-1}}}{{{B}\choose{I}}} =\frac{\frac{(B-1)!}{(I-1)!\times (B-I)!}}{\frac{p!}{I!\times (B-I)!}}=\frac{I}{B}=r.
\end{equation}
Substituting the simplification in \eqref{lemma_RAPS_dec_30} into \eqref{lemma_RAPS_dec_20} follows the claim in \eqref{lemma_RAPS_dec_claim_1}. To prove the claim in \eqref{lemma_RAPS_dec_claim_2} we can use the same argument that we used in proving \eqref{lemma_RAPS_dec_claim_1} to show that 
\begin{equation}\label{lemma_RAPS_dec_40}
\mathbb{E}_{\ccalI^t}\!\left[ \|   \bbx_{t+1}\!-\!\bbx^t\|^2\! \mid \ccalF^t   \right]
	\!=\!\frac{{{B-1}\choose{I-1}}}{{{B}\choose{I}}} (\gamma^t)^2\!  \left\|\nabla f( \bbx^{t}, \bbTheta^t)\right\|^2\!\!\!.
\end{equation}
By substituting the simplification in \eqref{lemma_RAPS_dec_30} into \eqref{lemma_RAPS_dec_40} the claim in \eqref{lemma_RAPS_dec_claim_2} follows.

\subsection{Proof of Proposition \ref{martingale_prop}}\label{apx_martingale_prop}


By considering the Taylor's expansion of $F(\bbx^{t+1})$ near the point $\bbx^t$ and observing the Lipschitz continuity of gradients $\nabla F$ with constant $M$ we obtain that the average objective function $F(\bbx^{t+1}) $  is bounded above by
\begin{equation}\label{martingale_10}
F(\bbx^{t+1})\leq F(\bbx^{t}) +\nabla F(\bbx^{t})^T (\bbx^{t+1}-\bbx^{t})+\frac{M}{2}\|\bbx^{t+1}-\bbx^{t}\|^2.
\end{equation}
Compute the expectation of the both sides of \eqref{martingale_10} with respect to the random set $\ccalI^t$ given the observed set of information $\ccalF^t$. Substitute $\mathbb{E}_{\ccalI^t}\!\left[{\bbx^{t+1}-\bbx^{t}\mid \ccalF^t}\right] $ and $\mathbb{E}_{\ccalI^t}\!\left[\|{\bbx^{t+1}-\bbx^{t}\|^2\mid \ccalF^t}\right] $ with their simplifications in \eqref{lemma_RAPS_dec_claim_1} and \eqref{lemma_RAPS_dec_claim_2}, respectively, to write
\begin{align}\label{martingale_20}
\mathbb{E}_{\ccalI^t}\left[F(\bbx^{t+1})\mid \ccalF^t\right]
	\leq 
	 F(\bbx^{t}) - {r\gamma^t }{}\ \nabla F(\bbx^{t})^T\nabla f( \bbx^{t}, \bbTheta^t)
	+ \frac{rM(\gamma^t)^2}{2 }\ \left\|\nabla f( \bbx^{t}, \bbTheta^t)\right\|^2.
\end{align}
Notice that the stochastic gradient $\nabla f( \bbx^{t}, \bbTheta^t)$ is an unbiased estimate of the average function gradient $ \nabla F(\bbx^{t})$. Therefore, we obtain $\mathbb{E}_{\bbTheta^t} \left[ \nabla f( \bbx^{t}, \bbTheta^t) \mid \ccalF^t\right]= \nabla F(\bbx^{t})$. Observing this relation and considering the assumption in \eqref{ekhtelaf}, the expected value of \eqref{martingale_20} with respect to the set of realizations $\bbTheta^t$ can be written as
\begin{align}\label{martingale_30}
\mathbb{E}_{\ccalI^t,\bbTheta^t}\left[F(\bbx^{t+1})\mid \ccalF^t\right]
	&\leq 
	 F(\bbx^{t}) - {r\gamma^t}{}\ \left\|\nabla F(\bbx^{t})\right\|^2
	 + \frac{rM(\gamma^t)^2 K}{2 }.
\end{align}
Subtracting the optimal objective function value $F(\bbx^*)$ form the both sides of \eqref{martingale_30} implies that
\begin{align}\label{martingale_40}
&\mathbb{E}_{\ccalI^t,\bbTheta^t}\left[F(\bbx^{t+1})-F(\bbx^*)\mid \ccalF^t\right]
	 \leq 
	 F(\bbx^{t}) -F(\bbx^*)
	 - r\gamma^t \ \left\|\nabla F(\bbx^{t})\right\|^2
	 + \frac{rM(\gamma^t)^2 K}{2 }. 
\end{align}
We proceed to find a lower bound for the gradient norm $\| \nabla F(\bbx^{t})\|$ in terms of the objective value error $F(\bbx^{t}) -\ F(\bbx^*)$. Assumption \ref{convexity_assumption} states that the average objective function $F$ is strongly convex with constant $m>0$. Therefore, for any $\bby,\bbz\in \reals^p$ we can write
\begin{equation}\label{martingale_50}
   F(\bby) \geq\ F(\bbz) +\nabla F(\bbz)^{T}(\bby-\bbz)   
   + {{m}\over{2}}\|{\bby - \bbz}\|^{2}.
\end{equation}
For fixed $\bbz$, the right hand side of \eqref{martingale_50} is a quadratic function of $\bby$ whose minimum argument we can find by setting its gradient to zero. Doing this yields the minimizing argument $\hby = \bbz- (1/m) \nabla  F(\bbz)$ implying that for all $\bby$ we must have
\begin{alignat}{2}\label{martingale_60}
F(\bby) \geq\ 
    &\ F(\bbw) +\nabla F(\bbz)^{T}(\hby-\bbz)   
   + {{m}\over{2}}\|{\hby - \bbz}\|^{2} \nonumber \\
   \ =\ 
         &\ F(\bbz) - \frac{1}{2m} \| \nabla F(\bbz)\|^{2} .
\end{alignat}
Observe that the bound in \eqref{martingale_60} holds true for all $\bby$ and $\bbz$. Setting values $\bby=\bbx^{*}$ and $\bbz=\bbx^{t}$  in \eqref{martingale_60} and rearranging the terms yields a lower bound for the squared gradient norm $\|\nabla F(\bbx^t)\|^2$ as
\begin{equation}\label{martingale_70}
 \| \nabla F(\bbx^{t})\|^{2} \geq 2m( F(\bbx^{t}) -   F(\bbx^*) ).
\end{equation} 
Substituting the lower bound in \eqref{martingale_70} by the norm of gradient square $ \| \nabla F(\bbx^{t})\|^{2}$ in \eqref{martingale_40} follows the claim in \eqref{martingale_prop_claim}.

\subsection{Proof of Theorem \ref{RAPSA_convg_thm}}\label{apx_RAPSA_convg_thm}


We use the relationship in \eqref{martingale_prop_claim} to build a supermartingale sequence. To do so, define the stochastic process $\alpha^t$ as
\begin{equation}\label{martingale_41}
   \alpha^t := F(\bbx^{t})-F(\bbx^*) +\frac{rMK}{2} \sum_{u=t}^{\infty}  (\gamma^u)^2 .
\end{equation}
Note that $\alpha^t$ is well-defined because $\sum_{u=t}^{\infty}  (\gamma^u)^2 \leq\sum_{u=0}^{\infty}  (\gamma^u)^2 <\infty$ is summable. Further define the sequence $\beta_t$ with values
\begin{equation}\label{martingale_42}
   \beta^t :=\ {2m\gamma^t r} (F(\bbx^{t})-F(\bbx^*) ).
\end{equation}
The definitions of sequences $\alpha^t$ and $\beta^t$ in \eqref{martingale_41} and \eqref{martingale_42}, respectively, and the inequality in \eqref{martingale_prop_claim} imply that the expected value $\alpha^{t+1}$ given $\ccalF^t$ can be written as
\begin{equation}\label{martingale_43}
   \E{\alpha^{t+1} \given \ccalF^t} \ \leq\ \alpha^t - \beta^t.
\end{equation}
Since the sequences $\alpha^t$ and $\beta^t$ are nonnegative it follows from \eqref{martingale_43} that they satisfy the conditions of the supermartingale convergence theorem -- see e.g. Theorem E$7.4$ of \cite{Solo}.
Therefore, we obtain that: (i) The sequence $\alpha^t$ converges almost surely to a limit. (ii) The sum $\sum_{t=0}^{\infty}\beta^t < \infty$ is almost surely finite. The latter result yields
\begin{equation}\label{martingale_44}
   \sum_{t=0}^{\infty} {2m\gamma^t r} (F(\bbx^{t})-F(\bbx^*) ) < \infty.
       \qquad\text{a.s.}
\end{equation}
Since the sequence of step sizes is non-summable there exits a subsequence of sequence $F(\bbx^{t})-F(\bbx^*)$ which is converging to null. This observation is equivalent to almost sure convergence of $\liminf F(\bbx^{t})-F(\bbx^*)$ to null,
\begin{equation}\label{martingale_45}
   \liminf_{t\to  \infty} F(\bbx^{t})-F(\bbx^*)=0.
       \qquad\text{a.s.}
\end{equation}
Based on the martingale convergence theorem for the sequences $\alpha^t$ and $\beta^t$ in relation \eqref{martingale_43}, the sequence $\alpha^t$ almost surely converges to a limit. Consider the definition of $\alpha^t$ in \eqref{martingale_41}. Observe that the sum $\sum_{u=t}^{\infty}  (\gamma^u)^2$ is deterministic and its limit is null. Therefore, the sequence of the objective function value error $F(\bbx^t)-F(\bbx^*)$ almost surely converges to a limit. This observation in association with the result in \eqref{martingale_45} implies that the whole sequence of $F(\bbx^t)-F(\bbx^*)$ converges almost surely to null,  
\begin{equation}\label{martingale_71}
   \lim_{t\to  \infty}\ F(\bbx^{t}) -   F(\bbx^*) =0.
       \qquad\text{a.s.}
\end{equation}
The last step is to prove almost sure convergence of the sequence $\|\bbx^t-\bbx^*\|^2$ to null, as a result of the limit in \eqref{martingale_71}. To do so, we follow by proving a lower bound for the objective function value error $F(\bbx^{t}) -   F(\bbx^*)$ in terms of the squared norm error $\|\bbx^t-\bbx^*\|^2$.
According to the strong convexity assumption, we can write the following inequality 
\begin{equation}\label{martingale_72}
F(\bbx^t)\geq F(\bbx^*)+\nabla F(\bbx^*)^T(\bbx^t-\bbx^*)+\frac{m}{2}\|\bbx^t-\bbx^*\|^2.
\end{equation}
Observe that the gradient of the optimal point is the null vector, i.e., $\nabla F(\bbx^*)=\bb0$. This observation and rearranging the terms in \eqref{martingale_72} imply that 
\begin{equation}\label{martingale_73}
F(\bbx^t) - F(\bbx^*)\geq \frac{m}{2}\|\bbx^t-\bbx^*\|^2.
\end{equation}
The upper bound in \eqref{martingale_73} for the squared norm $\|\bbx^t-\bbx^*\|^2$ in association with the fact that the sequence $F(\bbx^t) - F(\bbx^*)$ almost surely converges to null, leads to the conclusion that the sequence $\|\bbx^t-\bbx^*\|^2$ almost surely converges to zero. Hence, the claim in \eqref{rapsa_as_convg} is valid.

The next step is to study the convergence rate of RAPSA in expectation. In this step we assume that the diminishing stepsize is defined as $\gamma^t=\gamma^0 T^0/(t+T^0)$. Recall the inequality in \eqref{martingale_prop_claim}. Substitute $\gamma^t$ by $\gamma^0 T^0/(t+T^0)$ and compute the expected value of \eqref{martingale_prop_claim} given $\ccalF^0$ to obtain
\begin{align}\label{martingale_90}
&\E{F(\bbx^{t+1})-F(\bbx^*)}
\leq 
	\left( 1- \frac{2mr \gamma^0 T^0}{(t+T^0)} \right)\E{F(\bbx^{t}) -F(\bbx^*)}+ \frac{rMK(\gamma^0 T^0)^2}{2 (t+T^0)^2}.
\end{align}
We use the following lemma to show that the result in \eqref{martingale_90} implies  sublinear convergence of the sequence of expected objective value error $\E{F(\bbx^{t})-F(\bbx^*)}$.

%
\begin{lemma}\label{lecce22}
Let $c>1$, $b>0$ and $t^0 > 0$ be given constants and $u_{t}\geq 0$ be a nonnegative sequence that satisfies 
\begin{equation}\label{claim23}
   u^{t+1} \leq \left( 1- \frac{c}{t+t^0} \right) u^{t} 
                 + \frac{b}{{(t+t^0)}^{2}}\ ,
\end{equation}
{for all times $t\geq0$}. The sequence $u^t$ is then bounded as
\begin{equation}\label{lemma3_claim}
u^{t} \leq\  \frac{Q}{t+t^{0}},
\end{equation}
for all times $t\geq0$, where the constant $Q$ is defined as $ Q:=\max \{{b}/({c-1}),\ t^{0} u^{0} \}$ .
\end{lemma}
\begin{proof}
See Section 2 in (\cite{Nemirovski}).
\end{proof}

Lemma \ref{lecce22} shows that if a sequence  $u^t$ satisfies the condition in \eqref{claim23} then the sequence $u^t$ converges to null at least with the rate of $\ccalO(1/t)$. By assigning values $t^0=T^0$, $u^t=\E{F(\bbx^t)-F(\bbx^*)}$, $c=2m r\gamma^0 T^0$, and $b=rM  K(\gamma^0 T^0)^2/2$, the relation in \eqref{martingale_90} implies that the inequality in \eqref{claim23} is satisfied for the case that $2mr\gamma^0 T^0>1$. Therefore, the result in \eqref{lemma3_claim} holds and we can conclude that 
\begin{equation}\label{martingale_100}
\E{F(\bbx^t)-F(\bbx^*)}\leq \frac{ C}{t+T^0},
\end{equation}
where the constant $C$ is defined as 
\begin{equation}\label{martingale_110}
C= \max\left\{\frac{rMK (\gamma^0 T^0)^2}{4rm\gamma^0 T^0-2},\ T^0(F(\bbx^0)-F(\bbx^*))\right\}.
\end{equation}

\subsection{Proof of Theorem \ref{RAPSA_convg_thm_finite}}\label{apx_RAPSA_convg_thm_finite}


To prove the claim in \eqref{rapsa_as_convg_finite} we use the relationship in \eqref{martingale_prop_claim} (Proposition \ref{martingale_prop}) to construct a supermartingale. Define the stochastic process $\alpha^t$ with values 
\begin{equation}\label{finite_10}
\alpha^t\!:=\!\left( F(\bbx^t)-F(\bbx^*) \right) \times
	 \mathbf{1}\!\left\{\min_{u\leq t}  F(\bbx^u)-F(\bbx^*) \!>\! \frac{\gamma M K}{4m}\right\}
\end{equation}
The process $\alpha^t$ tracks the optimality gap $F(\bbx^t)-F(\bbx^*)$ until the gap becomes smaller than $\gamma M K/{2m}$ for the first time at which point it becomes $\alpha^t=0$. Notice that the stochastic process $\alpha^t$ is always non-negative, i.e., $\alpha^t\geq0$. Likewise, we define the stochastic process $\beta^t$ as
\begin{align}\label{finite_20}
\beta^t:={2\gamma m r}{}\left( F(\bbx^t)-F(\bbx^*)-\frac{\gamma M K}{4m} \right) 
\times
	 \mathbf{1}\left\{\min_{u\leq t}\  F(\bbx^u)-F(\bbx^*) > \frac{\gamma M K}{4m}\right\},
\end{align}
which follows $2\gamma m r\left( F(\bbx^t)-F(\bbx^*)-{\gamma M K}/{4m} \right) $ until the time that the optimality gap $F(\bbx^t)-F(\bbx^*)$ becomes smaller than $\gamma M K/{2m}$ for the first time. After this moment the stochastic process $\beta^t$ becomes null. According to the definition of $\beta^t$ in \eqref{finite_20}, the stochastic process satisfies $\beta^t\geq0$ for all $t\geq0$. Based on the relationship \eqref{martingale_prop_claim} and the definitions of stochastic processes $\alpha^t$ and $\beta^t$ in \eqref{finite_10} and \eqref{finite_20} we obtain that for all times $t\geq0$
\begin{equation}\label{finite_30}
\E{\alpha^{t+1} \mid \ccalF^t} \leq \alpha^t-\beta^t.
\end{equation}
To check the validity of \eqref{finite_30} we first consider the case that $\min_{u\leq t}\  F(\bbx^u)-F(\bbx^*) > {\gamma M K}{/4m}$ holds. In this scenario we can simply the stochastic processes in \eqref{finite_10} and \eqref{finite_20} as $\alpha^t=F(\bbx^t)-F(\bbx^*)$ and $\beta^t=2\gamma m r\left( F(\bbx^t)-F(\bbx^*)-{\gamma M K}/{4m} \right)$. Therefore, according to the inequality in \eqref{martingale_prop_claim} the result in \eqref{finite_30} is valid. The second scenario that we check is $\min_{u\leq t}\  F(\bbx^u)-F(\bbx^*) \leq {\gamma M K}{/4m}$. Based on the definitions of stochastic processes $\alpha^t$ and $\beta^t$, both of these two sequences are equal to 0. Further, notice that when $\alpha^t=0$, it follows that $\alpha^{t+1}=0$. Hence, the relationship in \eqref{finite_30} is true.

Given the relation in \eqref{finite_30} and non-negativity of stochastic processes $\alpha^t$ and $\beta^t$ we obtain that $\alpha^t$ is a supermartingale. The supermartingale convergence theorem yields: i) The sequence $\alpha^t$ converges to a limit almost surely. ii) The sum $\sum_{t=1}^\infty \beta^t$ is finite almost surely. The latter result implies that the sequence $\beta^t$ is converging to null almost surely, i.e., 
\begin{equation}\label{finite_40}
\lim_{t \to \infty} \beta^t \ =\ 0 \quad  \text{a.s. }
\end{equation}
Based on the definition of $\beta^t$ in \eqref{finite_20}, the limit in \eqref{finite_40} is true if one of the following events holds: i) The indicator function is null after for large $t$. ii) The limit $\lim_{t\to \infty} \left( F(\bbx^t)-F(\bbx^*)-{\gamma M K}/{4m} \right) =0$ holds true. From any of these two events we it is implied that  
\begin{equation}\label{finite_50}
\liminf_{t\to \infty}\ F(\bbx^t)-F(\bbx^*)\ \leq \ \frac{\gamma M K}{4m}\quad \text{a.s.}
\end{equation}
Therefore, the claim in \eqref{rapsa_as_convg_finite} is valid. The result in \eqref{finite_50} shows the objective function value sequence $F(\bbx^t)$ almost sure converges to a neighborhood of the optimal objective function value $F(\bbx^*)$. 

We proceed to prove the result in \eqref{rapsa_rate_finite}. Compute the expected value of \eqref{martingale_prop_claim}  given $\ccalF^0$ and set $\gamma^t=\gamma$ to obtain
\begin{align}\label{finite_constant_10}
\E{F(\bbx^{t+1})-F(\bbx^*)}
	&\leq 
	\left( 1- 2m  \gamma r \right)\E{ F(\bbx^{t}) -F(\bbx^*)}
	+ \frac{rM K\gamma^2}{2 }.
\end{align}
Notice that the expression in \eqref{finite_constant_10} provides an upper bound for the expected value of objective function error $\E{F(\bbx^{t+1})-F(\bbx^*)}$ in terms of its previous value $\E{F(\bbx^{t})-F(\bbx^*)}$ and an error term. Rewriting the relation in \eqref{finite_constant_10} for step $t-1$ leads to 
\begin{align}\label{finite_constant_20}
\E{F(\bbx^{t})-F(\bbx^*)}
	    \leq 
	\left( 1- 2m  \gamma r \right)\E{ F(\bbx^{t-1}) -F(\bbx^*)}
	+ \frac{rM K\gamma^2}{2 }.
\end{align}
Substituting the upper bound in \eqref{finite_constant_20} for the expectation $\E{F(\bbx^{t})-F(\bbx^*)}$ in \eqref{finite_constant_10} follows an upper bound for the expected error $\E{F(\bbx^{t+1})-F(\bbx^*)}$ as
\begin{align}\label{finite_constant_30}
\E{F(\bbx^{t+1})\!-\!F(\bbx^*)}
	\leq
	\left( 1- 2m  \gamma r \right)^2\E{ F(\bbx^{t-1}) \!-\!F(\bbx^*)}
	+ \frac{rM K\gamma^2}{2 }\left(1+\left( 1\!- \!2mr \gamma\right)\right)\!.
\end{align}
By recursively applying the steps in \eqref{finite_constant_20}-\eqref{finite_constant_30} we can bound the expected objective function error $\E{F(\bbx^{t+1})-F(\bbx^*)}$ in terms of the initial objective function error $F(\bbx^{0}) -F(\bbx^*)$ and the accumulation of the errors as
\begin{align}\label{finite_constant_40}
\E{F(\bbx^{t+1})\!-\!F(\bbx^*)}
	\leq 
	\left( 1- 2m  \gamma r \right)^{t+1}( F(\bbx^{0}) -F(\bbx^*))
	+ \frac{rM K\gamma^2}{2 }
	\sum_{u=0}^t
	\left( 1- {2mr\gamma}{} \right)^u\!\!\!.
\end{align}
Substituting $t$ by $t-1$ and simplifying the sum in the right hand side of \eqref{finite_constant_40} yields 
\begin{align}\label{finite_constant_50}
\E{F(\bbx^{t})-F(\bbx^*)}
	\leq 
	\left( 1- 2m  \gamma r \right)^{t}( F(\bbx^{0}) -F(\bbx^*))
	+ \frac{M  K\gamma}{4 m}
	\left[ 1- \left( 1- {2mr \gamma}{} \right)^t\right].
\end{align}
Observing that the term $1- \left( 1- {2mr \gamma}{} \right)^t$ in the right hand side of \eqref{finite_constant_50} is strictly smaller than $1$ for the stepsize $\gamma<1/(2mr)$, the claim in \eqref{rapsa_rate_finite} follows.

\section{Proofs Leading up to Theorem \ref{RAPSA_convg_thm_asyn}}\label{apx_asyn}

\subsection{Proof of Lemma \ref{exp_wrt_blocks_asyn}}\label{apx_exp_wrt_blocks_asyn}
\begin{myproof}
Recall that the components of vector $\bbx^{t+1}$ are equal to the components of $\bbx^t$ for the coordinates that are not updated at step $t$, i.e., $i\notin \ccalI^t$. For the updated coordinates $i\in \ccalI^t$ we know that $\bbx^{t+1}_{i}=\bbx^{t}_{i}-\gamma^t  \nabla_{\bbx_i^t} f( \bbx^{t-\tau}, \bbtheta^{t-\tau})$. Therefore, $B-1$ blocks of the vector $\bbx^{t+1}-\bbx^t$ are 0 and only one block is given by $-\gamma^t  \nabla_{\bbx_i} f( \bbx^{t-\tau}, \bbtheta^{t-\tau})$. Since the corresponding processor picks its block uniformly at random from the $B$ sets of blocks we obtain that the expected value of the difference $\bbx^{t+1}-\bbx^t$ with respect to the index of the block at time $t$ is given by
\begin{equation}\label{lemma_RAPS_dec_20_asyn}
\mathbb{E}_{\ccalI^t}\!\left[   \bbx^{t+1}-\bbx^t \mid \ccalF^{t}   \right]
	= \frac{1}{B} \left(  -\gamma^t  \nabla f( \bbx^{t-\tau}, \bbTheta^{t-\tau})  \right).
\end{equation}
Substituting the simplification in \eqref{lemma_RAPS_dec_20_asyn} in place of \eqref{lemma_RAPS_dec_20} in the proof of Lemma \ref{exp_wrt_blocks} and simplifying the resulting expression yields the claim in \eqref{lemma_RAPS_dec_claim_1_asyn}. To prove the claim in \eqref{lemma_RAPS_dec_claim_2_asyn} we can use the same argument that we used in proving \eqref{lemma_RAPS_dec_claim_1_asyn} to show that 
\begin{equation}\label{lemma_RAPS_dec_40_asyn}
\mathbb{E}_{\ccalI^t}\left[ \|   \bbx_{t+1}-\bbx^t\|^2 \mid \ccalF^{t}   \right]
	=\frac{(\gamma^t)^2}{B}   \left\|\nabla f( \bbx^{t-\tau}, \bbTheta^{t-\tau})\right\|^2,
\end{equation}
which completes the proof.
\end{myproof}

\subsection{Proof of Proposition \ref{martingale_prop_asyn}}\label{apx_RAPSA_convg_thm_finite_asyn}


By considering the Taylor's expansion of $F(\bbx^{t+1})$ near the point $\bbx^t$ and observing the Lipschitz continuity of gradients $\nabla F$ with constant $M$ we obtain that the average objective function $F(\bbx^{t+1}) $  is bounded above by
\begin{equation}\label{martingale_10_asyn}
F(\bbx^{t+1})\leq F(\bbx^{t}) +\nabla F(\bbx^{t})^T (\bbx^{t+1}-\bbx^{t})+\frac{M}{2}\|\bbx^{t+1}-\bbx^{t}\|^2.
\end{equation}
Compute the expectation of the both sides of \eqref{martingale_10_asyn} with respect to the random indexing set $\ccalI^{t}\subset \{1,\dots,B\}$ associated with chosen blocks given the observed set of information $\ccalF^{t}$. Substitute $\mathbb{E}_{\ccalI^t}\!\left[{\bbx^{t+1}-\bbx^{t}\mid \ccalF^t}\right] $ and $\mathbb{E}_{\ccalI^t}\!\left[\|{\bbx^{t+1}-\bbx^{t}\|^2\mid \ccalF^t}\right] $ with their simplifications in \eqref{lemma_RAPS_dec_claim_1_asyn} and \eqref{lemma_RAPS_dec_claim_2_asyn}, respectively, to write
\begin{align}\label{martingale_20_asyn}
\mathbb{E}_{\ccalI^t}\left[F(\bbx^{t+1})\mid \ccalF^t\right]
	\leq 
	 F(\bbx^{t}) - \frac{\gamma^t }{B}\ \nabla F(\bbx^{t})^T\nabla f( \bbx^{t-\tau}, \bbTheta^{t-\tau})+ \frac{M(\gamma^t)^2}{2B }\ \left\|\nabla f( \bbx^{t-\tau}, \bbTheta^{t-\tau})\right\|^2.
\end{align}
Notice that the stochastic gradient $\nabla f( \bbx^{t-\tau}, \bbTheta^{t-\tau})$ is an unbiased estimate of the average function gradient $ \nabla F(\bbx^{t-\tau})$. Therefore, we obtain $\mathbb{E} \left[ \nabla f( \bbx^{t-\tau}, \bbTheta^{t-\tau}) \mid \ccalF^t\right]= \nabla F(\bbx^{t-\tau})$. Observing this relation and considering the assumption in \eqref{ekhtelaf}, the expected value of \eqref{martingale_20_asyn} given the sigma algebra $\ccalF^t$ can be written as
\begin{align}\label{martingale_30_asyn}
\mathbb{E}\left[F(\bbx^{t+1})\mid \ccalF^t\right]
	\leq 
	 F(\bbx^{t}) - \frac{\gamma^t}{B}\ \nabla F(\bbx^{t})^T\nabla F( \bbx^{t-\tau})
 + \frac{M(\gamma^t)^2 K}{2B }.
\end{align}
By adding and subtracting the term $({\gamma^t}/{B}) \|\nabla F(\bbx^{t})\|^2$ to the right hand side of \eqref{martingale_30_asyn} we obtain 
\begin{align}\label{martingale_31_asyn}
\mathbb{E}\left[F(\bbx^{t+1})\mid \ccalF^t\right]
	\leq 
	 F(\bbx^{t}) - \frac{\gamma^t}{B}\ \|\nabla F(\bbx^{t})\|^2
 + \frac{\gamma^t}{B} \left(\|\nabla F(\bbx^{t})\|^2-\nabla F(\bbx^{t})^T\nabla F( \bbx^{t-\tau})\right)
	 + \frac{M(\gamma^t)^2 K}{2B}.
\end{align}
Observe that the third term on the right-hand side of \eqref{martingale_31_asyn} is the directional error due to the presence of delays from asynchronicity. We  proceed to find an upper bound for the expression $\|\nabla F(\bbx^{t})\|^2-\nabla F(\bbx^{t})^T\nabla F( \bbx^{t-\tau})$, which means that the error due to delay may be mitigated. To do so, notice that we can write 
\begin{align}\label{martingale_32_asyn}
\|\nabla F(\bbx^{t})\|^2-\nabla F(\bbx^{t})^T\nabla F( \bbx^{t-\tau})
&
=\nabla F(\bbx^{t})^T(\nabla F(\bbx^{t})-\nabla F( \bbx^{t-\tau}))
\nonumber\\
& 
\leq \|\nabla F(\bbx^{t})\|\|\nabla F(\bbx^{t})-\nabla F( \bbx^{t-\tau})\|,
\end{align}
where for the inequality we have used the Cauchy--Schwarz inequality. Apply the fact that the gradient of the objective function is $M$-Lipschitz continuous, which implies that $\|\nabla F(\bbx^{t})-\nabla F( \bbx^{t-\tau})\|\leq M\|\bbx^t-\bbx^{t-\tau}\|$. Substituting the upper bound $ M\|\bbx^t-\bbx^{t-\tau}\|$ for $\|\nabla F(\bbx^{t})-\nabla F( \bbx^{t-\tau})\|$ into \eqref{martingale_32_asyn} we obtain 
\begin{align}\label{martingale_33_asyn}
&\|\nabla F(\bbx^{t})\|^2-\nabla F(\bbx^{t})^T\nabla F( \bbx^{t-\tau})
\leq M \|\nabla F(\bbx^{t})\|\|\bbx^t-\bbx^{t-\tau}\|.
\end{align}
The difference norm $\|\bbx^t-\bbx^{t-\tau}\|$ is equivalent to $\|\sum_{s=t-\tau}^{t-1} (\bbx^{s+1}-\bbx^{s})\|$ which can be bounded above by $\sum_{s=t-\tau}^{t-1} \|\bbx^{s+1}-\bbx^{s}\|$ by the triangle inequality. Therefore,
\begin{align}\label{martingale_34_asyn}
\|\nabla F(\bbx^{t})\|^2-\nabla F(\bbx^{t})^T\nabla F( \bbx^{t-\tau})
\leq M \|\nabla F(\bbx^{t})\|\sum_{s=t-\tau}^{t-1} \|\bbx^{s+1}-\bbx^{s}\|.
\end{align}
Substitute the upper bound in \eqref{martingale_34_asyn} for $\|\nabla F(\bbx^{t})\|^2-\nabla F(\bbx^{t})^T\nabla F( \bbx^{t-\tau})$ into \eqref{martingale_31_asyn} to obtain 
\begin{align}\label{martingale_35_asyn}
\mathbb{E}\left[F(\bbx^{t+1})\mid \ccalF^t\right]
	\leq 
	 F(\bbx^{t}) - \frac{\gamma^t}{B}\ \|\nabla F(\bbx^{t})\|^2
 + \frac{M\gamma^t}{B}   \|\nabla F(\bbx^{t})\|\sum_{s=t-\tau}^{t-1} \|\bbx^{s+1}-\bbx^{s}\|
	 + \frac{M(\gamma^t)^2 K}{2B}.
\end{align}
Note that for any positive scalars $a$, $b$, and $\rho$ the inequality $ab \leq (\rho/2)a^2+(1/2\rho)b^2$ holds. If we set $a:=\|\nabla F(\bbx^t)\|$ and $b:=\sum_{s=t-\tau}^{t-1} \|\bbx^{s+1}-\bbx^{s}\|$ we obtain that 
\begin{align}\label{martingale_36_asyn}
 \|\nabla F(\bbx^{t})\|  \sum_{s=t-\tau}^{t-1} \|\bbx^{s+1}-\bbx^{s}\|
 &
 \leq 
 \frac{\rho}{2}    \|\nabla F(\bbx^{t})\|^2 
 +  \frac{1}{2\rho}\left[ \sum_{s=t-\tau}^{t-1} \|\bbx^{s+1}-\bbx^{s}\|\right]^2 \nonumber
 \\
 &
 \leq 
  \frac{\rho}{2}    \|\nabla F(\bbx^{t})\|^2 
 +  \frac{\tau}{2\rho} \sum_{s=t-\tau}^{t-1} \|\bbx^{s+1}-\bbx^{s}\|^2,
\end{align}
where the last inequality is an application of the triangle inequality to the second term on the right-hand side of the first line in \eqref{martingale_36_asyn}. Now substituting the upper bound in \eqref{martingale_36_asyn} into \eqref{martingale_35_asyn} yields 
\begin{align}\label{martingale_37_asyn}
\mathbb{E}\left[F(\bbx^{t+1})\mid \ccalF^t\right]
	\leq 
	 F(\bbx^{t}) - \left(\frac{\gamma^t}{B} -\frac{\rho M \gamma^t}{2B}\right) \|\nabla F(\bbx^{t})\|^2
	+  \frac{\tau M\gamma^t}{2\rho B} \sum_{s=t-\tau}^{t-1} \|\bbx^{s+1}-\bbx^{s}\|^2
	 + \frac{M(\gamma^t)^2 K}{2B}.
\end{align}
Compute the expected value of the both sides of \eqref{martingale_37_asyn} given the sigma-algebra $\ccalF^{t-1}$ to obtain
\begin{align}\label{martingale_38_asyn}
\mathbb{E}\left[F(\bbx^{t+1})\mid \ccalF^{t-1}\right]
	 &	\leq 
	 \mathbb{E}\left[F(\bbx^{t})\mid \ccalF^{t-1}\right] - \left(\frac{\gamma^t}{B} -\frac{\rho M \gamma^t}{2B}\right)  \mathbb{E}\left[\|\nabla F(\bbx^{t})\|^2\mid \ccalF^{t-1}\right]
	  \nonumber\\
	 & \qquad +  \frac{\tau M \gamma^t}{2\rho B}  \mathbb{E}\left[\sum_{s=t-\tau}^{t-1} \|\bbx^{s+1}-\bbx^{s}\|^2\mid \ccalF^{t-1}\right]
	 + \frac{M(\gamma^t)^2 K}{2B}\; , 
\end{align}
which can be simplified as 
\begin{align}\label{martingale_39_asyn}
\mathbb{E}\left[F(\bbx^{t+1})\mid \ccalF^{t-1}\right]
	 &	\leq 
	 \mathbb{E}\left[F(\bbx^{t})\mid \ccalF^{t-1}\right] - \left(\frac{\gamma^t}{B} -\frac{\rho M \gamma^t}{2B}\right)  \mathbb{E}\left[\|\nabla F(\bbx^{t})\|^2\mid \ccalF^{t-1}\right]
	  \nonumber\\
	 & \quad +  \frac{\tau M \gamma^t}{2\rho B}  \mathbb{E}\left[\sum_{s=t-\tau}^{t-2} \|\bbx^{s+1}-\bbx^{s}\|^2\mid \ccalF^{t-1}\right]
	 +  \frac{\tau M \gamma^t (\gamma^{t-1})^2 K}{2\rho B^2}  + \frac{M(\gamma^t)^2 K}{2B}.
\end{align}
Do the same up to $t-\tau$ to get
\begin{align}\label{martingale_391_asyn}
\mathbb{E}\left[F(\bbx^{t+1})\mid \ccalF^{t-\tau}\right]
	 &	\leq 
	 \mathbb{E}\left[F(\bbx^{t})\mid \ccalF^{t-\tau}\right] - \left(\frac{\gamma^t}{B} -\frac{\rho M \gamma^t}{2B}\right)  \mathbb{E}\left[\|\nabla F(\bbx^{t})\|^2\mid \ccalF^{t-\tau}\right]
	  \nonumber\\
	 &
	 \quad +  \frac{\tau M \gamma^t K}{2\rho B^2} \sum_{s=t-\tau}^{t-1} (\gamma^s)^2
	 + \frac{M(\gamma^t)^2 K}{2B}.
\end{align}
Notice that the sequence of stepsizes $\gamma^t$ is decreasing, thus the sum $ \sum_{s=t-\tau}^{t-1} (\gamma^s)^2$ in \eqref{martingale_391_asyn} can be bounded above by $\tau (\gamma^{t-\tau})^2$. Applying this substutition and subtracting the optimal objective function value $F(\bbx^*)$ from both sides of the implied expression lead to
\begin{align}\label{martingale_392_asyn}
\mathbb{E}\left[F(\bbx^{t+1})-F(\bbx^*)\mid \ccalF^{t-\tau}\right]
	 &	\leq 
	 \mathbb{E}\left[F(\bbx^{t})-F(\bbx^*)\mid \ccalF^{t-\tau}\right] 
	- \left(\frac{\gamma^t}{B} -\frac{\rho M \gamma^t}{2B}\right)  \mathbb{E}\left[\|\nabla F(\bbx^{t})\|^2\mid \ccalF^{t-\tau}\right]
	  \nonumber\\
	 &
	 \qquad +  \frac{\tau^2 M \gamma^t K(\gamma^{t-\tau})^2}{2\rho B^2}
	 + \frac{M(\gamma^t)^2 K}{2B}.
\end{align}
We make use of the fact that the average function $F(\bbx)$ is $m$-strongly convex in applying the relation $ \| \nabla F(\bbx^{t})\|^{2} \geq 2m( F(\bbx^{t}) -   F(\bbx^*) )$ to the expression \eqref{martingale_392_asyn}. Therefore,
\begin{align}\label{martingale_392_asyn}
\mathbb{E}\left[F(\bbx^{t+1})-F(\bbx^*)\mid \ccalF^{t-\tau}\right]
	 &	\leq 
	 \mathbb{E}\left[F(\bbx^{t})-F(\bbx^*)\mid \ccalF^{t-\tau}\right] 
 -2m \left(\frac{\gamma^t}{B} -\frac{\rho M \gamma^t}{2B}\right)  \mathbb{E}\left[ F(\bbx^{t}) -   F(\bbx^*) \mid \ccalF^{t-\tau}\right]
	  \nonumber\\
	 &
	 \qquad +  \frac{\tau^2 M \gamma^t K(\gamma^{t-\tau})^2}{2\rho B^2}
	 + \frac{M(\gamma^t)^2 K}{2B} \; ,
\end{align}
as stated in Proposition \ref{martingale_prop_asyn}.

\subsection{Proof of Theorem \ref{RAPSA_convg_thm_asyn}}\label{apx_RAPSA_convg_thm_asyn} 

\begin{myproof}
We use the result in Proposition \ref{martingale_prop_asyn} to define a martingale difference sequence with delay. Begin by defining the non-negative stochastic processes $\alpha^t$, $\beta^t$, and $\zeta^t$ for $t\geq 0$ as 
\begin{align}\label{proof_asyn_10}
&\alpha^t:= F(\bbx^t)-F(\bbx^*), \qquad 
\beta^t :=\frac{2m \gamma^t}{B}  \left[1-\frac{\rho M}{2}\right] (F(\bbx^{t}) -F(\bbx^*)), \nonumber\\
& \zeta^t:=\frac{M  K(\gamma^t)^2}{2B }+  \frac{\tau^2 MK \gamma^t (\gamma^{t-\tau})^2}{2\rho B^2}.
\end{align}
According to the definitions in \eqref{proof_asyn_10} and the inequality in \eqref{martingale_prop_claim_asyn} we can write 
\begin{equation}\label{proof_asyn_20}
\mathbb{E}\left[\alpha^{t+1}\mid \ccalF^{t-\tau}\right]\leq \mathbb{E}\left[\alpha^t\mid \ccalF^{t-\tau}\right] 
-\mathbb{E}\left[\beta^t\mid \ccalF^{t-\tau}\right] 
	+ \zeta^t.
\end{equation}
Computing the expected value of both sides of \eqref{proof_asyn_20} with respect to the initial sigma algebra $\mathbb{E}\left[\cdot\mid\ccalF^{0}\right]=\mathbb{E}\left[\cdot\right]$ yields
\begin{equation}\label{proof_asyn_30}
\mathbb{E}\left[\alpha^{t+1}\right]\leq \mathbb{E}\left[\alpha^t\right] 
-\mathbb{E}\left[\beta^t\right] 
	+ \zeta^t.
\end{equation}
Sum both sides of \eqref{proof_asyn_30} from $t=0$ to $t=\infty$ and consider the fact that $\zeta^t$ is summable and the sequence $\alpha^t$ is non-negative. Thus, we obtain that the series $\sum_{t=0}^{\infty} \mathbb{E}\left[\beta^t \right] < \infty$ is finite. By using Monotone Convergence Theorem, we pull the expectation outside the summand to obtain that $\mathbb{E}\left[\sum_{t=0}^{\infty} \beta^t \right] < \infty$. If we define $Y_n:=\sum_{t=0}^{n} \beta^t$, we obtain that $Y_n\geq 0 $ and $Y_n\leq Y_{n+1}$. Thus, from the result $\mathbb{E}\left[\sum_{t=0}^{\infty} \beta^t \right] < \infty$ we can conclude that $\sum_{t=0}^{\infty} \beta^t < \infty$ with probability 1. Now considering the definition of $\beta^t$ in \eqref{proof_asyn_10} and the non-summability of the stepsizes $\sum_{t=0}^{\infty} \gamma^t=\infty$, we obtain that a subsequence of the sequence $F(\bbx^{t}) -F(\bbx^*)$ almost surely converges to zero, i.e., the liminf of the sequence $F(\bbx^{t}) -F(\bbx^*)$ is zero with probability 1, 
\begin{align}\label{proof_asyn_40}
\liminf_{t\to \infty}\ F(\bbx^{t}) -F(\bbx^*)=0,\quad \text{a.s.}
\end{align}

The next step is to study the convergence rate of asynchronous RAPSA in expectation. By setting $\gamma^t=\gamma^0 T^0/(t+T^0)$ in  \eqref{martingale_prop_claim_asyn} and computing the expected value given the initial sigma algebra $\ccalF^0$ we obtain
\begin{align}\label{martingale_90_asyn}
&\mathbb{E}\left[F(\bbx^{t+1})-F(\bbx^*)\right]\\
&\quad	\leq 
	\left( 1-\frac{2m\gamma^0 T^0}{B(t+T^0)}  \left[1-\frac{\rho M}{2}\right] \right)\mathbb{E}\left[F(\bbx^{t}) -F(\bbx^*)\right]
	+ \frac{M  K(\gamma^0 T^0)^2}{2B(t+T^0)^2 }+  \frac{\tau^2 MK (\gamma^0 T^0)^3}{2\rho B^2(t+T^0)(t-\tau+T^0)^2}.\nonumber
\end{align}
Observe that it is not hard to check that if $t\geq2\tau+1$, then the inequality $(t-\tau+T^0)^2>t+T^0$ holds and we can substitute $1/((t-\tau+T^0)^2)$ in \eqref{martingale_90_asyn} by the upper bound $1/(t+T^0)$. Applying this substitution yields
\begin{align}\label{martingale_91_asyn}
&\mathbb{E}\left[F(\bbx^{t+1})-F(\bbx^*)\right]\nonumber\\
&\quad	\leq 
	\left( 1-\frac{2m\gamma^0 T^0}{B(t+T^0)}  \left[1-\frac{\rho M}{2}\right] \right)\mathbb{E}\left[F(\bbx^{t}) -F(\bbx^*)\right]
	+ \frac{M  K(\gamma^0 T^0)^2}{2B(t+T^0)^2 }+  \frac{\tau^2 MK (\gamma^0 T^0)^3}{2\rho B^2(t+T^0)^2}.
\end{align}
We use the result in Lemma \ref{lecce22} to show sublinear convergence of the sequence of expected objective value error $\E{F(\bbx^{t})-F(\bbx^*)}$.

Lemma \ref{lecce22} shows that if a sequence  $u^t$ satisfies the condition in \eqref{claim23} then the sequence $u^t$ converges to null at least with the rate of $\ccalO(1/t)$. By assigning values $t^0=T^0$, $u^t=\E{F(\bbx^t)-F(\bbx^*)}$, $c=(2m \gamma^0 T^0/B)(1-\rho M/2)$, and $b=M  K(\gamma^0 T^0)^2/2B+(\tau^2 MK (\gamma^0 T^0)^3)(2\rho B^2)$, the relation in \eqref{martingale_90} implies that the inequality in \eqref{claim23} is satisfied for the case that $c=(2m \gamma^0 T^0/B)(1-\rho M/2)>1$. Therefore, the result in \eqref{lemma3_claim} holds and we can conclude that 
\begin{equation}\label{martingale_100_asyn}
\E{F(\bbx^t)-F(\bbx^*)}\leq \frac{ C}{t+T^0},
\end{equation}
where the constant $C$ is defined as 
\begin{align}\label{martingale_110_asyn}
C= \max \left\{\frac{M  K(\gamma^0 T^0)^2/2B+(\tau^2 MK (\gamma^0 T^0)^3)(2\rho B^2)}{(2m \gamma^0 T^0/B)(1-\rho M/2)-1},\ T^0(F(\bbx^0)-F(\bbx^*))\right\}.
\end{align}
\end{myproof}

 \bibliographystyle{icml2014}
  \bibliography{bmc_article}
}

\end{document}